\documentclass{article}

\PassOptionsToPackage{numbers, compress}{natbib}

\usepackage[main, final]{neurips_2026}

\usepackage[utf8]{inputenc} 
\usepackage[T1]{fontenc}    
\usepackage{hyperref}       
\usepackage{url}            
\usepackage{booktabs}       
\usepackage{amsfonts}       
\usepackage{amsmath,amssymb,amsthm,algorithm,mathtools,bm}
\usepackage{nicefrac}       
\usepackage{microtype}      
\usepackage{xcolor}         
\usepackage{graphicx}       
\usepackage{enumitem}
\usepackage{tcolorbox}
\usepackage{algpseudocodex}
\usepackage{float}
\usepackage{amsmath,amsfonts,bm,amssymb,amsthm}
\usepackage{mathtools}
\usepackage{thmtools}
\usepackage{thm-restate}
\theoremstyle{plain}

\usepackage{cleveref}
\crefname{thm}{Theorem}{Theorems}
\crefname{lem}{Lemma}{Lemmas}

\def\1{\bm{1}}

\DeclareMathAlphabet{\mathsfit}{\encodingdefault}{\sfdefault}{m}{sl}
\SetMathAlphabet{\mathsfit}{bold}{\encodingdefault}{\sfdefault}{bx}{n}

\def\gN{{\mathcal{N}}}

\newcommand{\E}{\mathbb{E}}

\newcommand{\R}{\mathbb{R}}

\newcommand{\Var}{\mathrm{Var}}

\newcommand{\Cov}{\mathrm{Cov}}

\DeclareMathOperator*{\argmax}{arg\,max}

\DeclareMathOperator{\diag}{diag}

\DeclareMathOperator{\tr}{tr}

\usepackage{tikz}

\tikzset{
    carc/.style args={#1:#2:#3:#4:#5}{
        insert path={
            ({#1 + #5*cos(#3)}, {#2 + #5*sin(#3)}) arc (#3:#4:#5)
        }
    }
}       

\theoremstyle{plain}
\newtheorem{theorem}{Theorem}[section]
\newtheorem{proposition}[theorem]{Proposition}
\newtheorem{lemma}[theorem]{Lemma}

\theoremstyle{definition}

\newtheorem{assumption}{Assumption}
\theoremstyle{remark}
\newtheorem{remark}{Remark}

\graphicspath{{figs/}}

\AtBeginDocument{%
  \setlength{\abovedisplayskip}{3pt plus 2pt minus 2pt}
  \setlength{\belowdisplayskip}{3pt plus 2pt minus 2pt}
  \setlength{\abovedisplayshortskip}{1pt plus 2pt minus 2pt}
  \setlength{\belowdisplayshortskip}{1pt plus 2pt minus 2pt}
}

\title{Geometry-Aware Score-Repellent Monte Carlo}

\author{%
  Jie Hu$^{1}$\quad Lingyun Chen$^{2}$\quad Do Young Eun$^{2}$ \\
  $^{1}$Oakland University\quad
  $^{2}$NC State University \\
  \texttt{jiehu@oakland.edu}, \texttt{lchen47@ncsu.edu}, \texttt{dyeun@ncsu.edu}
}

\begin{document}

\maketitle

\begin{abstract}
The recently proposed Score-Repellent Monte Carlo (SRMC) method improves MCMC sampling by tilting the target density $\pi$ in $\R^d$ with the factor $\exp(-\alpha\,\theta_n^\top s(x))$, where the score function $s(x) = \nabla_x \log \pi(x)$, and $\theta_n$ is the score average from past samples. However, a single scalar repellence strength $\alpha\ge0$ cannot handle both the steep and the flat directions in anisotropic energy landscapes at once. We introduce \emph{Geometry-Aware SRMC} (GA-SRMC), which generalizes the Euclidean alignment to $\theta^\top M s(x)$ for any symmetric positive-definite matrix $M$. A stochastic-approximation central limit theorem under a trace constraint on $M$ identifies the trace-normalized inverse Fisher $M_{\rm opt}\propto S^{-1}$ as the unique minimizer of the worst-direction covariance bound, elevating the Fisher information $S=\Cov_\pi(s,s)$ from a proposal-side preconditioner (e.g., FisherMALA, natural gradient) to the bound-optimal repulsion-side matrix, with a single online estimate $\widehat S_n$ serving both roles at no additional computational cost. The full-Fisher online metric $M_n$ converges almost surely to $M_{\rm opt}$ and satisfies a CLT, providing a theoretical guarantee for the online implementation of the GA-SRMC framework. On an $8$-target simulation spanning the condition number $\kappa\!\in\![1.3,100]$ and dimension $d\!\in\![2,785]$, GA-SRMC dominates SRMC in ESS by $\boldsymbol{4.3\times}$ on anisotropic Gaussian ($\kappa\!=\!100$) and $\boldsymbol{6.8\times}$ on MNIST under the MALA baseline, by $\boldsymbol{16\text{--}32\%}$ on three well-conditioned logistic-regression posteriors, and by up to $\boldsymbol{+48\%}$ under FisherMALA; gains peak on the most anisotropic targets, and posterior-mean MSE is lower than with SRMC in all $16$ full-Fisher implementations.
\end{abstract}

\section{Introduction}
\label{sec:intro}

Efficiently sampling from complex high-dimensional continuous distributions remains one of the central algorithmic challenges for modern Markov chain Monte Carlo (MCMC) in Bayesian inference and statistical machine learning. The challenge goes beyond designing an unbiased Markov chain with the correct stationary distribution $\pi$: the chain must also explore efficiently under anisotropy, curved geometry, and metastable basins within a practical computational budget.

Methods such as the Metropolis-Adjusted Langevin Algorithm (MALA) \citep{roberts1996exponential} and Hamiltonian Monte Carlo (HMC) \citep{duane1987hybrid, neal2011mcmc} marked a major advance by using first-order information to steer proposals toward informative regions of the target. Subsequent developments, including Riemannian-manifold variants \citep{girolami2011riemann}, adaptive covariance schemes \citep{haario2001adaptive,andrieu2008tutorial}, and Fisher-based proposal metrics \citep{titsias2023optimal,hird2025Quantifying, seyboldt2026Preconditioning}, further sharpened the role of geometry in accelerating exploration. These methods act on the proposal mechanism of a sampler whose target distribution remains fixed.

A complementary philosophy is to use trajectory history to reshape the effective landscape itself. Metadynamics \citep{laio2002escaping,barducci2008well,henin2022enhanced}, adaptive biasing potential/force methods \citep{darve2001calculating,benaim2016convergence,benaim2018abp,benaim2020analysis}, the Wang-Landau algorithm \citep{wang2001efficient}, contour SGLD \citep{deng2020contour}, and Stein self-repulsive dynamics \citep{ye2020stein} all accumulate `summaries' of past exploration, e.g., bias histograms, collective-variable grids, or past samples used to evaluate repulsive corrections. While these summaries are effective, they either lead to increasing memory usage over time or demand explicit reweighting to recover unbiased Monte Carlo estimates. Combining self-repellence, asymptotic unbiasedness, and fixed memory usage remains difficult.

On finite state spaces, Self-Repellent Random Walk (SRRW) \citep{doshi2023self} and History-Driven Target (HDT)-MCMC \citep{hu2025beyond} employ the empirical occupation measure to build a history-driven approach that still converges to $\pi$ while penalizing states that have been visited too frequently. This mechanism can drive the asymptotic variance close to zero, while the full occupation measure does not scale to continuous state spaces $\R^d$. Score-Repellent Monte Carlo (SRMC) \citep{hu2026srmc} resolves this scalability bottleneck by compressing history into a fixed-dimensional running average of the score function $s(x)\triangleq\nabla_x\log\pi(x)$ in $\R^d$, yielding an auxiliary iterate $\theta_n\in\R^d$ such that
\begin{equation}
  \label{eq:sa-recursion}
  X_{n+1} \sim K_{\theta_n}(X_n,\cdot),
  \qquad
  \theta_{n+1}
  =
  \theta_n + \gamma_{n+1}\bigl(s(X_{n+1})-\theta_n\bigr),
\end{equation}
where $\gamma_n$ is the step size and \(K_\theta\) is any MCMC kernel with its invariant distribution
\begin{equation}
\label{eq:srmc_surrogate}
\pi_\theta(x)\propto \pi(x)\exp \bigl(-\alpha\,\theta^\top s(x)\bigr).
\end{equation}
Here, \(\alpha \ge 0\) quantifies the strength of self-repellency, and \(\pi_\theta\) is a history-dependent surrogate target, parameterized by the score average \(\theta\). Choosing either \(\alpha = 0\) or \(\theta = 0\) in \eqref{eq:srmc_surrogate} recovers the original target \(\pi\). The SRMC framework can be wrapped around essentially any base MCMC scheme. \citet{hu2026srmc} proved that the stochastic approximation (SA) recursion \eqref{eq:sa-recursion} converges almost surely to \(\theta^\star=0\), where the surrogate target $\pi_{\theta^\star} = \pi$ makes SRMC asymptotically unbiased and improves the sampling efficiency over the base MCMC algorithm. SRMC therefore occupies a distinctive position in this landscape: it is neither a purely kinetic adaptive sampler nor a biased history-flattening method; instead, it is a target-adaptive framework that discourages revisitation while converging back to~$\pi$, with $O(d)$ memory and normalization-free implementation.

Despite its attractive properties, however, SRMC remains \textit{geometry-blind} in a way that
complicates tuning. A practitioner must choose the repellence strength $\alpha$, but in \eqref{eq:srmc_surrogate}, it multiplies the raw Euclidean alignment $\theta^\top s(x)$, whose magnitude depends on the units, curvature, and anisotropy of the score field. The original SRMC work treats $\alpha$ as a target-dependent tuning parameter, using heuristic diagnostics such as keeping the exponent scale $\alpha|\theta_n^\top s(X_n)|$ moderate or increasing $\alpha$ adaptively, rather than providing a systematic treatment~\citep{hu2026srmc}. For an anisotropic Gaussian target $\pi = \mathcal N(0,\Sigma)$ with $\Sigma=\diag(\sigma_1^2,\ldots,\sigma_d^2)$, the coordinate scores satisfy $s_i(x) \!=\! -x_i/\sigma_i^2$, so $\theta^\top s(x) \!=\! -\sum_{i=1}^d \theta_i x_i/\sigma_i^2$. Because $\theta$ is a running average of past scores, the Euclidean inner product $\theta^\top s(x)$ amplifies curvature scale in the surrogate target: low-variance/high-curvature directions dominate the repulsion exponent simply because their scores are numerically large. A single scalar $\alpha$ must therefore handle directions with different variances: if it is large enough to move the chain along broad, flat directions, it overshoots in steep directions where scores are already large; if it is made small enough to avoid this overreaction, the repulsion is too weak to help in broad directions where mixing is slow. As a result, geometry-blind SRMC applies repulsion where the score is numerically large, rather than where the chain actually has difficulty mixing.

This mismatch suggests a more refined question: \textit{instead of tuning $\alpha$ around raw score scales $\theta^\top s(x)$, should the repulsion term compare the history so far with the current score using a different metric?} Common scale-free normalizations such as cosine similarity and $\theta^\top s(x)/\|s(x)\|$, while seemingly effective at first sight, can in fact prevent the surrogate target from recovering $\pi$ at equilibrium, thereby inducing bias; see Appendix~\ref{apx:alt-repulsion} for a detailed account. On the other hand, in the aforementioned Gaussian example, taking the weighted inner product with the weight matrix $M = \Sigma$ gives
$
\theta^\top M s(x) = \theta^\top \Sigma s(x) = -\sum_{i=1}^d \theta_i x_i,
$
which removes the variance-dependent denominators from the repulsion term. The issue is therefore not the score-based history itself, but the metric used to measure the alignment between the history variable $\theta$ and the current score $s(x)$. This naturally leads to the weighted inner product $\theta^\top M s(x)$, where $M$ should encode the geometry of the target. 

Thus motivated, in this paper, we introduce \textbf{Geometry-Aware} (GA)-SRMC, defined by the surrogate target family with a weight matrix $M \in \R^{d \times d}$:
\begin{equation}
  \label{eq:ga-surrogate}
  \pi_\theta^{(M)}(x) \propto \pi(x)\exp\bigl(-\alpha\,\theta^\top M s(x)\bigr),
  \qquad M = M^\top \succ 0,
\end{equation}
where $M = I_d$ recovers the original SRMC surrogate target. The SA update for $\theta_n$ is unchanged; only the metric through which the score average acts back on the surrogate target is modified. This identifies a new degree of freedom for the \emph{repulsion-side} geometry within the surrogate target $\pi_\theta^{(M)}$, which is distinct from proposal-side preconditioning. In preconditioned MALA \citep{titsias2023optimal,hird2025Quantifying}, Riemannian or preconditioned HMC \citep{girolami2011riemann,seyboldt2026Preconditioning}, and adaptive Metropolis \citep{haario2001adaptive}, a symmetric matrix $A$ determines the proposal distribution for a chosen target. For example, if a preconditioned Langevin proposal is applied to a generic target $\nu$ with score $s_\nu(x)=\nabla_x\log\nu(x)$, then $q_A(y \mid x)=\gN\!\left(y\mid x+\frac{1}{2}A s_\nu(x),A\right)$. Classically, one takes $\nu=\pi$. In GA-SRMC, by contrast, the matrix $M$ specifies the history-dependent surrogate target $\pi_\theta^{(M)}$ in \eqref{eq:ga-surrogate}. Consequently, $M$ deforms the energy landscape passed to the base kernel, whereas $A$ shapes how the base kernel proposes moves on that landscape by enforcing this surrogate target. 
For a Metropolis–Hastings kernel with target $\nu$ and proposal $q_A$, imposing $\nu = \pi_\theta^{(M)}$ in \eqref{eq:ga-surrogate} yields the following normalization-free acceptance ratio:
\begin{equation}
\label{eq:accept-ratio}
\frac{\pi_{\theta}^{(M)}(y)\,q_A(x\mid y)}{\pi_{\theta}^{(M)}(x)\,q_A(y \mid x)} 
=\underbrace{\frac{\pi(y)\,q_A(x \mid y)}{\pi(x)\,q_A(y \mid x)}}_{\text{proposal-side correction}}
\underbrace{\exp\{-\alpha\theta^\top M(s(y)-s(x))\}}_{\text{Ours: geometry-aware score repellence}},
\end{equation}
which highlights that the two mechanisms are complementary, with $M$ acting as a repulsion-side matrix and $A$ as a proposal-side preconditioner.
In the smooth continuous setting, the matrix $S \!\triangleq\! \Cov_\pi(s,s) \!=\! \E_\pi[s(X)s(X)^\top]$ is the Fisher information matrix by Stein's identity $\E_{\pi}[s(X)] = 0$.\footnote{As in \citet{titsias2023optimal}, Fisher information matrix $S$ is the score covariance under $\pi$, not to be confused with the classical position-dependent Fisher information $\E_{Y \sim p(\cdot | x)}[\nabla_x \log p(Y | x) \nabla_x \log p(Y | x)^\top]$ of a model $p (y | x)$ with parameter $x$.} Our Theorem~\ref{thm:optimal} identifies the inverse-Fisher choice $M_{\rm opt}\propto S^{-1}$ as the minimizer of a worst-direction bound derived from the central limit theorem (CLT). This finding strengthens the role of $S^{-1}$ as a preconditioner for gradient-based proposals \citep{titsias2023optimal,hird2025Quantifying,seyboldt2026Preconditioning} and natural-gradient optimization \citep{amari1998natural,martens2020new,wu2024improved}. We show that the same Fisher geometry governs not only proposal-side preconditioning but also the repulsion term of a score-repellent surrogate target.
{%
\setlength{\textfloatsep}{0pt plus 2pt minus 2pt}
\setlength{\intextsep}{0pt plus 2pt minus 2pt}
\begin{figure}[t]
  \centering
  \includegraphics[width=\linewidth]{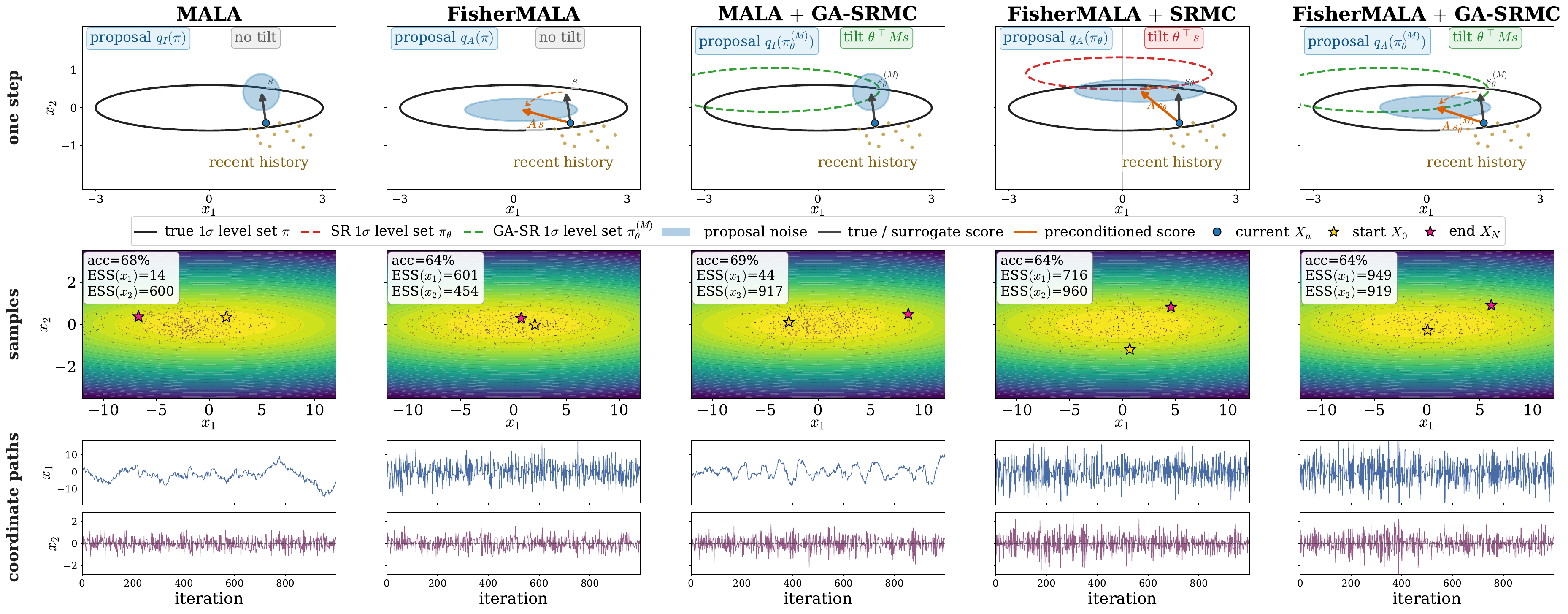}
\caption{
\textbf{Proposal-side and repulsion-side geometry are complementary.}
Top: one-step mechanism. The preconditioner $A$ shapes proposal movements, while the GA-SRMC matrix $M$ shapes the history-dependent surrogate target. Middle/bottom: samples and coordinate paths on an anisotropic Gaussian target. Combining FisherMALA with GA-SRMC yields the most balanced effective sample sizes (ESS) across coordinates, computed from $N = 1000$ post-burn-in samples with ESS estimation in \citet{geyer1992practical}. Full setup and details are deferred to Appendix \ref{apx:fig1-setup}.}
\label{fig:geometry-blind}
\end{figure}
}%

Figure~\ref{fig:geometry-blind} illustrates the geometric issue motivating GA-SRMC. Standard MALA \citep{roberts1996exponential} uses isotropic proposal noise and mixes poorly along the elongated coordinate $x_1$, whose trace is highly autocorrelated, resulting in the imbalanced $(\mathrm{ESS}(x_1),\mathrm{ESS}(x_2))=(14,600)$. FisherMALA \citep{titsias2023optimal} preconditions the local move on the proposal side, so its $x_1$ trace is less `sticky' than under MALA, and $\mathrm{ESS}(x_1)$ rises from $14$ to $601$. MALA with GA-SRMC adds history-dependent repulsion and improves over MALA along the slow direction, but without preconditioning its $x_1$ trajectory remains clearly correlated and still underperforms FisherMALA, showing that the base sampler’s quality is crucial. Conversely, FisherMALA with SRMC \citep{hu2026srmc} shows that score repellence can further improve a preconditioned sampler, $(\mathrm{ESS}(x_1),\mathrm{ESS}(x_2))=(716,960)$, yet its Euclidean tilt is not aligned with the target geometry. FisherMALA with GA-SRMC aligns both the preconditioner $A$ and the Fisher-weighted matrix $M$ with this geometry and attains the most balanced pair, $(\mathrm{ESS}(x_1),\mathrm{ESS}(x_2))=(949,919)$, at essentially the same acceptance rate (`acc' in the legend).

Our contributions are as follows:

1. 
We generalize SRMC by replacing the Euclidean alignment $\theta^\top s(x)$ in the surrogate target $\pi_{\theta}$ in \eqref{eq:srmc_surrogate} with a geometry-aware alignment $\theta^\top M s(x)$. For any fixed symmetric positive-definite $M$, we show that the surrogate family is well-defined, GA-SRMC remains asymptotically unbiased, and its limiting covariance is determined by $SM$, with the choice of $M$ playing a role.

2. 
We show that the inverse-Fisher matrix $M_{\rm opt}\propto S^{-1}$ equalizes the spectrum of $SM$ under the trace constraint $\operatorname{tr}(SM)=\tau$, uniquely minimizing the universal worst-direction bound on the score-history covariance. We then estimate $S$ online from the score that GA-SRMC already computes, with diagonal and full-Fisher versions that trade computational costs against geometric fidelity. Under additional stability assumption, the learned full-Fisher metric $M_n$ converges almost surely to $M_{\rm opt}$, and the score history and the estimator obey the same CLT as under the fixed metric $M_{\rm opt}$ in hindsight.

3.
We evaluate GA-SRMC on $8$ targets with two sampler families (MALA and FisherMALA) and three variants $\{$base, $+$\,SRMC, $+$\,GA-SRMC$\}$. In the worst-coordinate ESS under MALA, GA-SRMC dominates SRMC by $\boldsymbol{4.3\times}$ on anisotropic Gaussian ($\kappa\!=\!100$) and $\boldsymbol{6.8\times}$ on MNIST at each method's best stability-capped $\alpha$; under FisherMALA, it adds $+\,48\%$ on the strongly anisotropic Gaussian and ties with SRMC on near-isotropic targets. Its posterior-mean MSE is lower than SRMC's in all $16$ full-Fisher comparisons.

The message is straightforward: \emph{historical trajectory should be used in combination with the appropriate geometry.} SRMC introduces a constant-memory score-repellent mechanism for target adaptation \citep{hu2026srmc}; GA-SRMC identifies the metric in this mechanism as an object with its own asymptotic geometry. For smooth continuous targets, this geometry is governed by the Fisher information matrix $S=\Cov_\pi(s,s)$, which shapes not only proposal-side preconditioning in gradient-based MCMC, but also the repellent surrogate target itself.
\section{GA-SRMC with a Fixed Weight Matrix \texorpdfstring{$M$}{M}}
\label{sec:ga-srmc}

\textbf{Notation.}
All vectors are column vectors. For symmetric $A,B$, Loewner ordering $A\!\succeq\! B$ (\textit{resp}. $A\!\succ\! B$) means $A-B$ is positive semidefinite (\textit{resp}. definite). We write $I_d$ for the $d\!\times\! d$ identity matrix, $\tr(A)$ for the trace, and $A^{1/2}$ for the principal square root when $A\!\succeq\!0$. Let $\lambda_{\min}(A)$ and $\lambda_{\max}(A)$ be the smallest and largest eigenvalues of a square $A$. The operator $\operatorname{diag}(A)$ extracts the diagonal of $A$, and $\operatorname{diag}(a)$ is the diagonal matrix with entries from the vector $a$. For vector-valued functions $g,h$ and a probability distribution $\pi$, denote their marginal covariance by $\Cov_\pi(g,h)\!\triangleq\! \E_\pi[(g(X)\!-\!\E_\pi[g(X)])(h(X)\!-\!\E_\pi [h(X)])^\top]$. We use $\xrightarrow[]{a.s.}$ and $\xrightarrow[]{dist.}$ for almost sure and weak convergence, respectively. Let $\gN(\mu, V)$ be a multivariate Gaussian with mean $\mu$ and covariance $V$.

\textbf{Basic structure of GA-SRMC.}
We analyze the GA-SRMC surrogate target $\pi_\theta^{(M)}$ in \eqref{eq:ga-surrogate} with a fixed symmetric matrix $M\succ0$. For each $\theta\in\R^d$, define
\begin{equation}\textstyle
\label{eq:ZM-SM-FM}
    Z_\theta^{(M)} \triangleq \int_{\R^d} \pi(x)\exp\{-\alpha\,\theta^\top M s(x)\}\,dx,~~
    S_M(\theta)\triangleq \E_{\pi_\theta^{(M)}}[s(X)],~~
    F_M(\theta)\triangleq \E_{\pi_\theta^{(M)}}[f(X)],
\end{equation}
where the score function $s=\nabla_x \log\pi$, and $f:\R^d\to\R^m$ is a test function of interest, with $\mu\triangleq\E_\pi[f(X)]$ for the quantity to be estimated. We write the joint iterate and its update function as:
\begin{equation}
\label{eq:mean-field-def}
\vartheta_n\triangleq(\theta_n,\mu_n)\in\R^{d+m}, ~~~~H(\vartheta_n,x)\triangleq(s(x)-\theta_n,~f(x)-\mu_n).
\end{equation}
Let $K_{\nu}$ denote an arbitrary base MCMC kernel with invariant distribution $\nu$, e.g., random walk Metropolis, MALA, or HMC. At each iteration, GA-SRMC runs a base kernel $K_{\theta, M}$ targeting the current surrogate $\pi_{\theta}^{(M)}$ in \eqref{eq:ga-surrogate} and updates the joint iterate online via the SA recursion
\begin{equation}
\label{eq:ga-srmc-recursion}
    X_{n+1}\!\sim\! K_{\theta_n, M}(X_n,\cdot),~    \vartheta_{n+1}\!=\!\vartheta_n+\gamma_{n+1}\,H(\vartheta_n,X_{n+1}), ~ \text{where} ~ \gamma_n\!=\!(n+1)^{-\rho},~ \rho\!\in\!(0.5,1].
\end{equation}
Note that the weight matrix $M$ enters \eqref{eq:ga-srmc-recursion} only through the surrogate target used by the transition kernel; the SA recursion itself is unchanged from SRMC.

\textbf{Assumptions.}
We adopt the regularity assumptions of the SA framework for SRMC \citep{hu2026srmc}, specialized to the surrogate family $\{\pi_\theta^{(M)}\}_{\theta\in\R^d}$ with fixed $M$: Lipschitz continuity and appropriate tail behavior of the score $s$ \citep{durmus2017nonasymptotic,chak2023optimal} (see also \citet[Appendix C.1]{hu2026srmc}), along with a drift condition and Lipschitz continuity of the Markov kernel \citep{andrieu2006ergodicity,roberts2007coupling,fort2011convergence}. They make the surrogates $\pi_{\theta}^{(M)}$ well-defined, and the base MCMC algorithm uniformly geometrically ergodic while $\theta$ stays bounded. Formal statements and proofs are in Appendix~\ref{app:section2-proofs}.

\begin{proposition}[Well-posedness \& equilibrium]
\label{prop:fixedM-structure}
Under Assumption \ref{ass:target}, for every symmetric $M\succ 0$:
\begin{enumerate}[leftmargin=1.25em,itemsep= -0.25em]
    \item (\textbf{Well-posedness}) For every $\theta\in\R^d$ and every $\alpha\ge 0$, the normalizing constant $Z_\theta^{(M)}<\infty$, so the surrogate $\pi_\theta^{(M)}$ in \eqref{eq:ga-surrogate} is a well-defined probability distribution on $\R^d$.
    \item (\textbf{Equilibrium}) The point $\vartheta^\star\triangleq (0, \mu)\in\R^{d+m}$ is the unique globally asymptotically stable equilibrium of the mean-field ODE $\dot\vartheta_t=h_M(\vartheta_t)$, where $h_M(\vartheta_t)\triangleq (S_M(\theta_t)-\theta_t, ~ F_M(\theta_t)-\mu_t)$.
\end{enumerate}
\end{proposition}
Proposition~\ref{prop:fixedM-structure} shows that the geometry-aware tilt is a valid target-adaptation mechanism. While $M$ changes how the accumulated score history contributes to the surrogate exponent, it does not change the desired equilibrium: for every $\theta$ and $M$, the surrogate remains a proper probability distribution and coincides with $\pi_0^{(M)}=\pi$ at the unique equilibrium $\theta^\star=0$. Moreover, the joint SA recursion \eqref{eq:ga-srmc-recursion} retains the correct equilibrium with the mean field $h_M(\vartheta^\star) \!=\! 0$ and is globally stable. The proof verifies that the weighted score tilt does not introduce any alternative stable equilibria, so $M$ can reshape the transient dynamics and asymptotic covariance without altering the limiting target $\pi$.

The SA viewpoint shows how this `reshaping' occurs. The recursion \eqref{eq:ga-srmc-recursion} is a noisy, decreasing-step Euler discretization of the mean-field ODE $\dot\vartheta_t=h_M(\vartheta_t)$ with controlled Markovian noise from the current surrogate kernel (see also \citep{kushner2003stochastic,borkar2009stochastic,borkar2025ode} for ODE analysis of general SA). Around the stable equilibrium $\vartheta^\star=(0,\mu)$, the deterministic part is governed by the Jacobian
\begin{equation}
\label{eq:fixedM-jacobian}
    J_\star^{(M)}
    \triangleq
    \nabla_\vartheta h_M(\vartheta^\star)
    =
    \begin{bmatrix}
        -I_d-\alpha SM & 0\\
        -\alpha CM & -I_m
    \end{bmatrix},
    ~~
    S\triangleq \E_\pi[s(X)s(X)^\top],
    ~~
    C\triangleq \Cov_\pi(f,s),
\end{equation}
which captures the local deterministic damping of the mean-field ODE: the CLT covariance below balances this damping against the long-run covariance of the Markovian noise.

\textbf{Asymptotic analysis for GA-SRMC with fixed $M$.}
The following proposition extends the SA theory of SRMC \citep{hu2026srmc}: for fixed $M$, the surrogate $\pi_\theta^{(M)}$ still converges to $\pi$, and a CLT holds that Section~\ref{sec:ga-srmc-practical} uses to choose optimal $M_{\rm opt}$. Let $\{Z_k:k\in\mathbb Z\}$ be a trajectory of the base MCMC kernel $K_\pi$ targeting $\pi$. For vector-valued functions $a, b$ on $\R^d$, define the long-run covariance
\begin{equation}
\label{eq:long-run-cov-U}
U_{ab}\triangleq
\lim_{T\to\infty}\frac{1}{T}\,
\E\left[
\left(\sum_{k=0}^{T-1}(a(X_k) - \E_{\pi}[a(X)])\right)\,
\left(\sum_{k=0}^{T-1}(b(X_k) - \E_{\pi}[b(X)])\right)^\top
\right],
\end{equation}
and we write the long-run covariance associated with the function $H(\vartheta^\star, \cdot)$, as introduced in \eqref{eq:mean-field-def} and evaluated at the equilibrium point $\vartheta^\star$, as $U_{HH}=
\left[\begin{smallmatrix}
U_{ss} ~&~ U_{sf}\\
U_{fs} ~&~ U_{ff}
\end{smallmatrix}\right]$. Let $\mathbf 1_{(\cdot)}$ be the indicator function.
\begin{proposition}[Almost sure convergence and CLT for fixed $M$]
\label{prop:clt}
The iterate in~\eqref{eq:ga-srmc-recursion} satisfies $\vartheta_n\xrightarrow[n\to\infty]{a.s.}\vartheta^\star$ and
$\gamma_n^{-1/2}(\vartheta_n-\vartheta^\star)\xrightarrow[n\to\infty]{dist.}\gN(0,\Sigma_\vartheta^{(M)})$, where $\Sigma_\vartheta^{(M)}$ is the unique solution of
\begin{equation}
\label{eq:lyapunov-eq}
    \bigl(\tfrac{\mathbf{1}_{\{\rho=1\}}}{2}I+\!J_\star^{(M)}\bigr)\Sigma_\vartheta^{(M)}
    +\Sigma_\vartheta^{(M)}\bigl(\tfrac{\mathbf{1}_{\{\rho=1\}}}{2}I+J_\star^{(M)}\bigr)^\top +U_{HH}=0.
\end{equation}
\end{proposition}
Setting $M=I_d$ in Proposition~\ref{prop:clt} recovers \citet[Theorem 3.3]{hu2026srmc}. Proposition~\ref{prop:clt} has two consequences that will be used in Section~\ref{sec:ga-srmc-practical}. First, GA-SRMC remains asymptotically unbiased for every fixed symmetric $M\succ0$: the score-history variable satisfies $\theta_n\to0$ almost surely, the surrogate target returns to $\pi_0^{(M)}=\pi$, and the Monte Carlo estimator converges to $\mu=\E_\pi[f(X)]$.
Second, the proposition turns the mean-field picture into a covariance equation. The matrix $J_\star^{(M)}$ in \eqref{eq:fixedM-jacobian} represents the deterministic local linearization of the ODE that accounts for the influence of $M$, whereas $U_{HH}$ denotes the long-run covariance associated with the baseline MCMC sampler, in which the preconditioning matrix $A$ (e.g., as in preconditioning techniques) plays a role. The Lyapunov equation \eqref{eq:lyapunov-eq} balances these two effects. Let $\Sigma_{\theta\theta}^{(M)}$ and $\Sigma_{\mu\mu}^{(M)}$ denote the limiting covariance of $\theta$- and $\mu$-iterates, respectively, within $\Sigma_\vartheta^{(M)}$. Focusing on the sequence $\{\theta_n\}$ alone leads to the Lyapunov equation $D_M\Sigma_{\theta\theta}^{(M)} + \Sigma_{\theta\theta}^{(M)}D_M^\top = U_{ss}$, whose solution $\Sigma_{\theta\theta}^{(M)}$ admits the closed form:
\begin{equation}
\label{eq:theta-block-fixedM}
    \Sigma_{\theta\theta}^{(M)} = \int_{0}^{\infty} e^{-tD_M} U_{ss}e^{-tD_M^\top}dt,
    \quad \text{where} \quad
    D_M\triangleq \beta I_d+\alpha SM,
    \quad
    \beta\triangleq 1-\tfrac{\mathbf 1_{\{\rho=1\}}}{2}.
\end{equation}
We call $D_M \succ 0$ the \emph{CLT damping operator}. The Jacobian has the negative drift block $-I_d-\alpha SM$, while the Lyapunov equation uses its positive damping counterpart $D_M=\beta I_d+\alpha SM$. Directions associated with small eigenvalues of $D_M$ are weakly damped and therefore dominate the worst-direction covariance bounds. This is the analytical counterpart of the top row for Figure~\ref{fig:geometry-blind}: proposal-side preconditioning changes how the chain moves via $U_{ss}$, whereas repulsion-side preconditioning changes the geometry of the score-history damping through $SM$ and in turn $D_M$. Under the SRMC ($M\!=\!I_d$), anisotropy of $S$ produces uneven damping across score directions; under the inverse-Fisher choice $M\!\propto\! S^{-1}$, the damping becomes balanced across directions. This is the bridge to Section~\ref{sec:ga-srmc-practical}.

\section{Choosing and Learning the Optimal Weight Matrix \texorpdfstring{$M_{\rm opt}$}{Mopt}}
\label{sec:ga-srmc-practical}

\textbf{Optimal choice of $M_{\rm opt}$.} By Proposition \ref{prop:clt}, any fixed symmetric $M \succ 0$ retains the limiting target $\pi$ and the equilibrium $\theta^\star=0$ unchanged, while only redistributing the damping in $D_M$ across score directions. The worst-direction bound on $\Sigma_{\theta\theta}^{(M)}$ is governed by the smallest eigenvalue of $D_M$, i.e., $\lambda_{\min}(D_M)\!=\!\beta\!+\!\alpha\lambda_{\min}(SM)$, so a poor choice of $M$ produces a slow eigendirection of $SM$ that no single scalar $\alpha$ can rescue uniformly. Multiplying $M$ by scalar $c$ has the same effect on the surrogate exponent~\eqref{eq:ga-surrogate} as multiplying $\alpha$ by $c$, so an unconstrained optimization over $M$ would only re-parametrize the repellence strength. We therefore impose the trace constraint
\begin{equation}
\label{eq:trace-constraint}
 \tr(SM)\;=\;\E_\pi\!\bigl[s(X)^\top M s(X)\bigr]\;=\;\tau\;>\;0,
\end{equation}
which fixes the mean squared score magnitude under $\pi$ and the trace of $\alpha SM$ in $D_M$. Our GA-SRMC framework embeds an optimization question: \emph{which weight matrix $M$ should be selected to best enhance the worst-case performance?} The relationship between $\Sigma_{\theta\theta}^{(M)}$ and $\lambda_{\min}(SM)$, along with the resulting optimal choice $M_{\rm opt}$, is given below.

\begin{theorem}[Trace-normalized inverse-Fisher matrix]
\label{thm:optimal}
Assume the fixed-$M$ CLT in Proposition~\ref{prop:clt} and suppose $S\succ0$. Then, the limiting covariance $\Sigma_{\theta\theta}^{(M)}$ of the $\theta_n$-iterate satisfies
\begin{equation}
\label{eq:theta-universal-bound}
 \Sigma_{\theta\theta}^{(M)}
 \preceq
 \frac{\lambda_{\max}(U_{ss}S^{-1})}{2\{\beta+\alpha\lambda_{\min}(SM)\}}\,S.
\end{equation}
Among all $M \!\!=\!\! M^\top \!\!\succ\! 0$ with \eqref{eq:trace-constraint}, the upper bound in \eqref{eq:theta-universal-bound} is uniquely minimized in Loewner order by
\begin{equation}
\label{eq:m-opt}
 M_{\rm opt}=\frac{\tau}{d}S^{-1}.
\end{equation}
At this choice, limiting covariances associated with score history and Monte Carlo estimator become
\begin{equation}
\label{eq:theta-exact-opt}
 \Sigma_{\theta\theta}^{(M_{\rm opt})}
 =\frac{U_{ss}}{2(\beta+\alpha\tau/d)}, \quad \Sigma_{\mu\mu}^{(M_{\rm opt})} = \frac{U_{gg}}{2\beta} + O(\alpha^{-1}),
\end{equation}
where $U_{gg}$ is defined by \eqref{eq:long-run-cov-U} with the function $g(x) \triangleq f(x) - CS^{-1}s(x)$. 
\end{theorem}
Proofs are in Appendices~\ref{app:metric-proof} and \ref{app:residual-decomp}. Intuitively, $SM$ is similar to $S^{1/2}MS^{1/2} \!\succ\! 0$, so its eigenvalues are positive, and the trace constraint fixes their average at $\tau/d$. A larger $\lambda_{\min}(SM)$ leads to a tighter bound in \eqref{eq:theta-universal-bound}. Because $\lambda_{\min}(SM) \le \tau/d$, and equality occurs only when $S^{1/2}MS^{1/2} = (\tau/d) I_d$ (equivalently, $M = (\tau/d)I_d$), the inverse Fisher choice minimizes the bound: it maximizes the weakest guaranteed score-history contraction under the trace constraint. Moreover, $D_{M_{\rm opt}} = (\beta + \alpha \tau / d) I_d$, which turns \eqref{eq:theta-block-fixedM} into the closed form \eqref{eq:theta-exact-opt}. This limiting covariance is strictly Loewner-monotone in $\alpha$, recovers the base sampler's long-run covariance $U_{ss}/(2\beta)$ at $\alpha\!=\!0$, and vanishes as $\alpha\!\to\!\infty$. Hence, it gives a strictly stronger result than the Frobenius-norm ordering obtained for the suboptimal $M = I_d$ in \citep[Proposition~3.4]{hu2026srmc}.

The limiting covariance $\Sigma_{\mu\mu}^{(M_{\rm opt})}$ in~\eqref{eq:theta-exact-opt} reflects a variance-reduction principle: the score history damps only the score-aligned component $CS^{-1}s$ of the test function $f$, while the score-orthogonal residual $g \!\triangleq\! f \!-\! CS^{-1} s$, with $\Cov_{\pi}(g, s) = 0$, sets the floor $U_{gg}/(2\beta)$, which is the long-run covariance of the residual noise $g(X)\!-\!\mu$ under the base MCMC sampler and is independent of $M$ for a fixed kernel. As $\alpha \to \infty$, GA-SRMC reduces the score-aligned covariance at rate $O(\alpha^{-1})$, but leaves a residual $U_{gg}/2\beta$ that score-based repulsion alone cannot remove. This decomposition aligns with established Monte Carlo variance-reduction methods, namely zero-variance MCMC \citep{assaraf1999zero,mira2013zero} and the control-functional approach \citep{oates2017control}, in which the score-aligned term $CS^{-1}s$ corresponds exactly to the portion of $f$ that is eliminated by score-based control variates, with a detailed discussion in Appendix \ref{app:position-vs-srmc}. In Appendix~\ref{app:gaussian-proof}, we discuss two special cases of the independent surrogate sampler and Gaussian target to specify the role of $M_{\rm opt}$. 

\begin{remark}[why optimize the score-history covariance]
The score history $\theta_n$ is the memory of GA-SRMC and the control signal that reshapes the next surrogate target. Proposition~\ref{prop:clt} implies that, at large time,
$
 \E[\theta_n\theta_n^\top]
 =
 \gamma_n \Sigma_{\theta\theta}^{(M)} + o(\gamma_n),
$
so $\Sigma_{\theta\theta}^{(M)}$ is the leading covariance constant of the running score imbalance. Reducing this block means that the trajectory cancels score directions more efficiently and that the surrogate feedback learns the local target geometry with less residual fluctuation. The inverse-Fisher choice $M_{\rm opt}$ therefore optimizes the asymptotic covariance of the very statistic that drives score repellence. The residual decomposition above explains when this improvement transfers directly to Monte Carlo efficiency.
\end{remark}

The inverse Fisher $S^{-1}$ has an established \emph{proposal-side} role in FisherMALA~\citep{titsias2023optimal,hird2025Quantifying}, FisherHMC~\citep{seyboldt2026Preconditioning}, and in natural-gradient methods~\citep{amari1998natural,girolami2011riemann,martens2020new}; Theorem~\ref{thm:optimal} identifies it as the bound-optimal \emph{repulsion-side} choice. Although these two criteria act on different objects, i.e., expected squared jumped distance or the condition number of the target distribution versus CLT damping operator $D_M$ in the score-history feedback, they both select $S^{-1}$ because both whiten a score-driven object against the same anisotropy of $\pi$. Practically, an online estimate of $S^{-1}$ can simultaneously serve both purposes without any additional overhead, placing GA-SRMC at the unique intersection of two closely related design choices: preconditioned-only samplers leverage Fisher geometry on the proposal side but lack score repellence, whereas SRMC modifies the target on the repulsion side while using the geometry-blind metric $M\!=\!I_d$. GA-SRMC uniquely employs Fisher geometry on \emph{both} sides whenever a preconditioning sampler is used, which can further improve sampling efficiency.

\textbf{Online Fisher implementation.}
We replace the unknown Fisher $S$ by an online estimate $\widehat S_n$ from the score evaluations $\{s(X_k)\}_{k\le n}$ already computed along the chain and then normalize the trace via
\begin{equation}
\label{eq:online-fisher-trace}
 M_n
 \;=\;
 \frac{\tr(\widehat S_{n})}{d}\,\widehat S_{n}^{-1},
\end{equation}
which decouples $\alpha$ from the geometry: $\tr(\widehat S_{n} M_n)\!=\!\tr(\widehat S_{n})$ matches the Fisher-weighted trace of the SRMC choice $M\!=\!I_d$, so $\alpha$ solely sets the repellence strength and $M_n$ only its geometry. The \emph{diagonal} version stores coordinate-wise score second moments and inverts a vector at $O(d)$ cost per iteration; it captures axis-aligned anisotropy but ignores cross-coordinate score correlations, as in TensorFlow/Stan~\citep{abadi2016tensorflow,carpenter2017stan}, and can be counterproductive on highly correlated targets~\citep[Section~3.5]{hird2025Quantifying}. The \emph{full} version, in the spirit of FisherMALA~\citep{titsias2023optimal}, maintains $\widehat S_{n+1}\!=\!(1\!-\!\gamma_{n+1})\widehat S_n+\gamma_{n+1}s_ns_n^\top$, making it suitable when score correlations are dense and the dimension is moderate. This recursion costs $O(d^2)$ from the Sherman-Morrison update (Appendix \ref{app:fisher-full}). In
Algorithm~\ref{alg:online-fisher-ga-srmc}, \textsc{FisherUpdate} updates the inverse Fisher estimate on the fly. Complete implementations of the two Fisher estimates and Algorithm~\ref{alg:online-fisher-ga-srmc} are provided in Appendix~\ref{app:online-fisher-details}.

{%
\setlength{\textfloatsep}{0pt plus 2pt minus 2pt}
\setlength{\intextsep}{2pt plus 2pt minus 2pt}
\begin{algorithm}[t]
\caption{GA-SRMC with online trace-normalized Fisher estimation}
\label{alg:online-fisher-ga-srmc}
\begin{algorithmic}[1]
\Require score $s$; kernel family $K_{\theta,M}$; repellence $\alpha$; small
$\delta$; stepsizes $\gamma_n,\omega_n$; type $\in\{\mathrm{diag},\mathrm{full}\}$.
\State Initialize $X_0$, $\theta_0=0$, $\mu_0$, and $\hat S_n \succ 0$.
\For{$n=0,\ldots,N-1$}
 \State Compute $M_{n}$ via \eqref{eq:online-fisher-trace}; draw $X_{n+1}\sim K_{\theta_n,M_{n}}(X_n,\cdot)$ targeting~\eqref{eq:ga-surrogate}.
 \State $s_{n+1}\gets s(X_{n+1})$; update $\vartheta_{n+1} \gets \vartheta_n + \gamma_{n+1} H(\vartheta_n, X_{n+1})$ as in \eqref{eq:ga-srmc-recursion}.
 \State \textsc{FisherUpdate}: $(\hat S_{n+1} \gets \hat S_n + \gamma_{n+1} (s_{n+1}s_{n+1}^\top - \hat S_n)$ (diag: diagonal only).
\EndFor
\State \Return samples $\{X_n\}$, Monte Carlo estimator $\mu_N$.
\end{algorithmic}
\end{algorithm}
}%

For Langevin-type base kernels, GA-SRMC adds \emph{one Hessian-vector product (HVP) per surrogate-score evaluation}, two per MALA step, over its base sampler. By~\eqref{eq:ga-surrogate}, the surrogate score is
\begin{equation}
\label{eq:surrogate-score}
\nabla_x\log\pi_\theta^{(M)}(x)
=
s(x)-\alpha\nabla_x^2\log\pi(x)M\theta,
\end{equation}
so that a preconditioned Langevin proposal with proposal matrix $A_n$ and step size $\epsilon_n$ takes the form
\(
Y\!=\!X_n\!+\!\epsilon_n A_n\nabla_x\log\pi_\theta^{(M_n)}(x)\!+\!\sqrt{2\epsilon_n}A_n^{1/2}\xi_n,
\)
followed by the Metropolis accept–reject step in \eqref{eq:accept-ratio}. The HVP $\nabla_x^2\log\pi(X_n)\,M_n\theta_n$ can be computed via automatic differentiation \citep{pearlmutter1994fast,griewank2008evaluating} or finite differences of the gradient \citep{dagreou2024howtocompute}, without forming the Hessian; for samplers that already maintain an online inverse-Fisher preconditioner~\citep{titsias2023optimal,seyboldt2026Preconditioning}, one trace-normalized inverse-Fisher estimate serves both purposes. As in \citep{titsias2023optimal}, the step size $\epsilon_n$ is adapted on the log scale, i.e., $\log\epsilon_{n+1}=\log\epsilon_n+c_\epsilon(a_n-0.574)$, where $c_\epsilon\in(0,1)$ is a positive adaptation gain, $a_n$ is the realized Metropolis acceptance probability, and $0.574$ is the standard MALA optimal-acceptance target \citep{roberts1998optimal,roberts2001optimal}. This is the adaptive scheme used in Figure~\ref{fig:geometry-blind}.

Propositions~\ref{prop:fixedM-structure}--\ref{prop:clt} and Theorem~\ref{thm:optimal} hold $M$ fixed. In Algorithm~\ref{alg:online-fisher-ga-srmc}, $M_n$ is updated online at every step, so the augmented iterate $z_n \!\triangleq\! (\theta_n, \mu_n, \widehat S_n)$ follows one SA recursion of the form~\eqref{eq:ga-srmc-recursion}, with the extra coordinate $\widehat S_n$ and kernel $K_{\theta_n, M_n}$.

\begin{theorem}[Online inverse-Fisher metric]
\label{thm:online}
Suppose additionally that $\widehat S_n$ remains bounded in the positive definite cone (Assumption~\ref{ass:online-stability}). Then, with $\tau\!=\!\tr S$ in $M_{\rm opt}$, (i) $\theta_n \!\to\! 0$, $\mu_n\!\to\!\mu$, $\widehat S_n\!\to\! S$, and $M_n\!\to\! M_{\rm opt}$ almost surely; and (ii) $z_n$ satisfies a joint CLT whose $(\theta,\mu)$ marginal is the CLT of Proposition~\ref{prop:clt} at $M\!=\!M_{\rm opt}$: $\gamma_n^{-1/2}(\vartheta_n-\vartheta^\star)\xrightarrow{dist.}\gN(0,\Sigma_\vartheta^{(M_{\rm opt})})$.
\end{theorem}
The proof in Appendix~\ref{app:online-proof} follows the steps of Proposition~\ref{prop:clt} for the augmented iterate $z_n$ with two modifications: the moving-metric Lyapunov function $\theta^\top\widehat S^{-1}\theta$ and the identity $\pi_0^{(M)}=\pi$ for every $M\succ 0$. Because of the latter, the Fisher state does not contribute to the linearization of the $(\theta,\mu)$ mean field at equilibrium, which implies that, in the asymptotic regime, the online estimate $M_n$ does not compromise the optimality of the hindsight selection $M_{\rm opt}$ in Theorem~\ref{thm:optimal}.

\begin{remark}[Scope of the fixed-$M$ and online results]
\label{rem:online-scope}
Proposition~\ref{prop:clt} and Theorem~\ref{thm:optimal} focus on a fixed weight matrix $M$, whereas Theorem~\ref{thm:online} analyzes Algorithm~\ref{alg:online-fisher-ga-srmc} in a fully online regime using the full Fisher information estimator. In this setting, an inaccurate Fisher estimate during the early iterations may induce a mis-scaling of the proposal through $A_n$, and through $M_n$, it can reduce $\lambda_{\min}(SM_n)$ and thereby weaken the bound in~\eqref{eq:theta-universal-bound}. Theorem~\ref{thm:online} establishes only that this degradation is asymptotically negligible. Practical stabilization is provided by a regularization parameter $\delta$, trace normalization, and initialization based on the MAP Hessian, as stated in Section \ref{sec:experiments}.
\end{remark}
\section{Experiments}
\label{sec:experiments}

We evaluate GA-SRMC on an anisotropic Gaussian target with condition number $\kappa\!=\!100$ (\textsc{Gauss-2D}), chosen to isolate geometric effects, and on seven Bayesian logistic-regression (BLR) posteriors~\eqref{eq:blr-posterior} with a Gaussian prior of precision $\tau_0\!=\!0.1$ on standard real-world benchmarks: \textsc{Ripley}~\citep{ripley2007pattern}, the UCI~\citep{dua2019uci} datasets \textsc{Pima}, \textsc{Heart}, \textsc{Australian}, and \textsc{German}, the CoIL Challenge \textsc{Caravan} dataset~\citep{vanderputten2000caravan}, and \textsc{MNIST}~\citep{lecun1998gradient} (full target specification in Appendix~\ref{apx:blr-target}). Collectively, these problems span dimensionalities $d\!\in\![2,785]$ and posterior condition numbers $\kappa\!\in\![1.3,100]$. Following the implementation similar to~\citet{titsias2023optimal}, we run $40$ independent Markov chains for each algorithm, collecting $20{,}000$ post-burn-in samples after a burn-in of $2{,}000$ iterations, and evaluate each chain on two complementary metrics (mean $\pm$ standard deviation across the $40$ chains): (i)~the \emph{worst-coordinate ESS} (Table~\ref{tab:blr}), defined as the minimum across coordinates of the single-chain bulk effective sample size computed via ArviZ~\citep{arviz2019}, which quantifies mixing along the slowest coordinate; and (ii)~the \emph{per-dimension posterior-mean MSE} $\frac{1}{d}\|\bar X-\mu_{\rm ref}\|_2^2$ (Table~\ref{tab:mse} with details in Appendix~\ref{apx:mse}), which quantifies the estimation error of the posterior mean across all $d$ coordinates. All preconditioned variants in Tables~\ref{tab:blr} and~\ref{tab:mse} use the full online Fisher estimator from Algorithm~\ref{alg:online-fisher-ga-srmc} with $\widehat S_{n,\delta}\!\triangleq\!\widehat S_n+\delta I_d$ and $\delta\!=\!0.01$ to ensure invertibility; the sensitivity to $\delta$ is analyzed in Appendix~\ref{apx:lambda-tuning}.

\textbf{Stability cap on $\alpha$.}
For a fixed value of $\theta$, the GA-SRMC chain samples the surrogate $\pi_\theta^{(M)}$, which coincides with the true target $\pi$ only in the asymptotic regime $\theta_n\!\to\!\theta^\star\!=\!0$ established in Proposition~\ref{prop:clt} and Theorem~\ref{thm:online}. For finite~$n$ and sufficiently large~$\alpha$, the chain can temporarily explore a distorted distribution, so relying on ESS comparisons alone can be misleading. To guard against this, we impose a \emph{stability cap} defined separately for each target and each method: it is the largest $\alpha$ in our search grid for which the post-burn-in marginal standard deviations remain within $30\%$ of those under a Gaussian Laplace approximation at the maximum a posteriori (MAP), following the validation diagnostic of~\citet{titsias2023optimal} (complete details in Appendix~\ref{apx:validity}). Because the unscaled tilt in SRMC destabilizes the chain at smaller $\alpha$ than the geometry-aware tilt in GA-SRMC, the two methods typically admit different stability caps. Accordingly, all results in Tables~\ref{tab:blr} and~\ref{tab:mse} are reported at each method's best-ESS $\alpha$ under its stability cap.

\begin{table}[t]
\centering
\caption{Worst-coordinate ESS (mean$\pm$std over $40$ chains, $20{,}000$ post-burn-in samples) at each method's best stability-capped $\alpha$; $\delta\!=\!0.01$. \textbf{Bold}: better of \{SRMC, GA-SRMC\} per family; \underline{\textbf{underlined}}: row best; $\kappa$: condition number of the posterior precision at the MAP.}
\footnotesize
\begin{tabular}{lrrrrrr}
\toprule
 & \multicolumn{3}{c}{\textsc{MALA} family} & \multicolumn{3}{c}{\textsc{FisherMALA} family} \\
\cmidrule(lr){2-4}\cmidrule(lr){5-7}
Target $(n,d,\kappa)$ & base & + SRMC & + GA-SRMC & base & + SRMC & + GA-SRMC \\
\midrule
Gauss\,2D (2,2,$10^{2}$) & 94$\pm$21 & 104$\pm$21 & \textbf{452$\pm$17} & 9894$\pm$371 & 12459$\pm$224 & \underline{\textbf{18425$\pm$204}} \\
\midrule
Australian (690,15,2.8) & 2245$\pm$170 & 4885$\pm$122 & \textbf{5937$\pm$149} & 4087$\pm$215 & 11170$\pm$228 & \underline{\textbf{11638$\pm$260}} \\
Heart (270,14,2.7) & 2821$\pm$183 & 6310$\pm$150 & \textbf{7324$\pm$163} & 4330$\pm$133 & 9451$\pm$150 & \underline{\textbf{9549$\pm$159}} \\
Pima (768,9,2.7) & 3448$\pm$119 & 7638$\pm$137 & \textbf{10084$\pm$298} & 5422$\pm$177 & 10975$\pm$206 & \underline{\textbf{13358$\pm$279}} \\
Ripley (1250,3,1.3) & 8405$\pm$250 & 13372$\pm$180 & \textbf{13591$\pm$180} & 9514$\pm$277 & 15353$\pm$207 & \underline{\textbf{15477$\pm$167}} \\
German (1000,25,4.6) & 1437$\pm$83 & \textbf{4663$\pm$152} & 4173$\pm$110 & 3259$\pm$114 & 9338$\pm$205 & \underline{\textbf{9449$\pm$238}} \\
Caravan (5822,86,16) & 401$\pm$35 & 922$\pm$126 & \textbf{1353$\pm$32} & 1694$\pm$91 & \underline{\textbf{6381$\pm$202}} & 6022$\pm$172 \\
MNIST (4000,785,55) & 49$\pm$22 & 74$\pm$31 & \textbf{501$\pm$7} & 612$\pm$46 & 1389$\pm$38 & \underline{\textbf{1399$\pm$27}} \\
\bottomrule
\end{tabular}
\label{tab:blr}
\end{table}

\begin{table}[t]
\centering
\caption{Posterior-mean MSE $\frac{1}{d}\|\bar X-\mu_{\rm ref}\|_2^2$ (mean$\pm$std over $40$ runs, units of $10^{-4}$) for the samplers of Table~\ref{tab:blr}. \textbf{Bold}: best per family; \underline{\textbf{underlined}}: row best. Details in Appendix~\ref{apx:mse}.}
\label{tab:mse}
\scriptsize
\setlength{\tabcolsep}{3pt}
\begin{tabular}{lrrrrrr}
\toprule
 & \multicolumn{3}{c}{\textsc{MALA} family} & \multicolumn{3}{c}{\textsc{FisherMALA} family} \\
\cmidrule(lr){2-4}\cmidrule(lr){5-7}
Target & base & + SRMC & + GA-SRMC & base & + SRMC & + GA-SRMC \\
\midrule
Gauss\,2D & 1880.317$\pm$3070.938 & 1827.031$\pm$2775.423 & \textbf{1.236$\pm$1.775} & 10.647$\pm$12.809 & 6.563$\pm$7.206 & \underline{\textbf{0.227$\pm$0.252}} \\
\midrule
Australian & 17.702$\pm$9.018 & 1.273$\pm$0.832 & \textbf{0.908$\pm$0.474} & 13.722$\pm$5.160 & 0.977$\pm$0.436 & \underline{\textbf{0.830$\pm$0.333}} \\
Heart & 18.157$\pm$8.370 & 1.176$\pm$0.527 & \textbf{0.871$\pm$0.328} & 12.745$\pm$5.779 & 0.847$\pm$0.334 & \underline{\textbf{0.719$\pm$0.255}} \\
Pima & 14.418$\pm$10.815 & 0.627$\pm$0.422 & \textbf{0.370$\pm$0.167} & 7.884$\pm$4.325 & 0.346$\pm$0.143 & \underline{\textbf{0.289$\pm$0.105}} \\
Ripley & 5.786$\pm$5.515 & 0.205$\pm$0.188 & \underline{\textbf{0.198$\pm$0.178}} & 6.512$\pm$5.968 & 0.232$\pm$0.195 & \textbf{0.223$\pm$0.185} \\
German & 24.062$\pm$10.052 & 1.624$\pm$0.953 & \textbf{0.904$\pm$0.394} & 13.791$\pm$4.586 & 0.937$\pm$0.437 & \underline{\textbf{0.635$\pm$0.233}} \\
Caravan & 126.083$\pm$22.058 & 45.631$\pm$8.525 & \textbf{10.217$\pm$1.671} & 34.951$\pm$6.629 & 5.949$\pm$1.317 & \underline{\textbf{2.938$\pm$0.496}} \\
MNIST & 685.244$\pm$33.782 & 541.485$\pm$26.944 & \textbf{55.500$\pm$2.556} & 136.093$\pm$5.676 & 56.842$\pm$2.694 & \underline{\textbf{55.000$\pm$2.740}} \\
\bottomrule
\end{tabular}
\end{table}

\begin{figure}[t]
  \centering
  \includegraphics[width=0.9\linewidth]{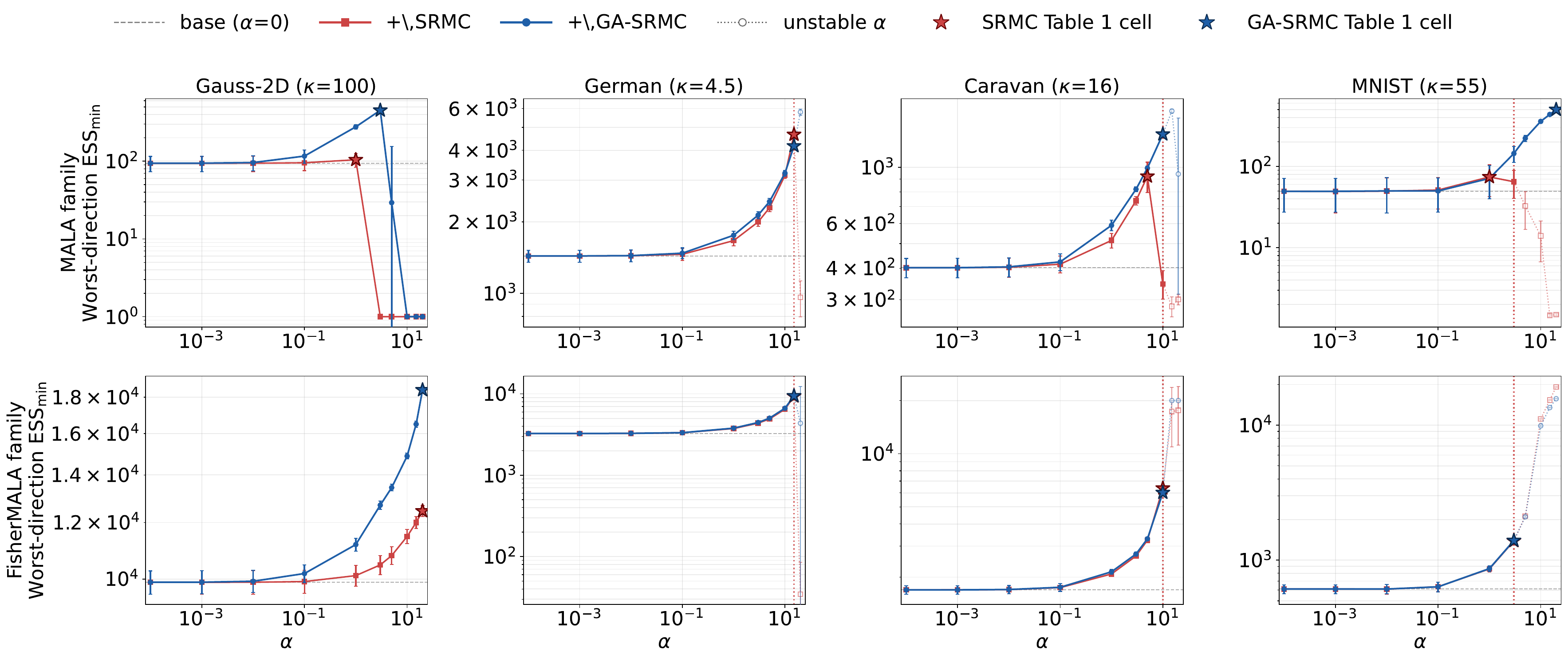}
  \caption{Worst-coordinate ESS versus $\alpha$ on four representative targets (log-log axes) for the MALA (top) and FisherMALA (bottom) proposals. Markers and bars are mean $\pm$ std across $40$ chains.}
  \label{fig:alpha-sweep}
\end{figure}

\textbf{Two sampler families.}
Table~\ref{tab:blr} reports the worst-coordinate ESS for the \textsc{MALA} family ($A\!=\!I_d$) and the \textsc{FisherMALA} family ($A\!\propto\!S^{-1}$), each evaluated in three versions: the base sampler, $+$\,SRMC, and $+$\,GA-SRMC.
The MALA family isolates the repulsion-side geometry most clearly, since $M\theta$ is the only channel through which Fisher geometry enters the sampler. On \textsc{Gauss-2D}, MALA$+$SRMC reaches $104$ at $\alpha\!=\!1$ compared with $452$ for MALA$+$GA-SRMC at $\alpha\!=\!3$, a $4.3\times$ gain; on \textsc{MNIST}, $74$ at $\alpha\!=\!1$ versus $501$ at $\alpha\!=\!20$, a $6.8\times$ gain; on \textsc{Caravan}, $922$ at $\alpha\!=\!5$ versus $1353$ at $\alpha\!=\!10$, a $47\%$ increase; and on \textsc{Australian}, \textsc{Heart}, and \textsc{Pima}, GA-SRMC yields $16$--$32\%$ higher ESS at $\alpha\!\in\!\{25,30\}$. In the FisherMALA family, where the proposal is already preconditioned by the inverse Fisher matrix, the geometry-aware correction is partly additive: the GA-SRMC$/$SRMC ratio is $1.48$ on \textsc{Gauss-2D} ($\alpha\!=\!20$), $1.01$--$1.22$ on \textsc{Australian}, \textsc{Pima}, \textsc{Heart}, \textsc{Ripley}, \textsc{German}, and \textsc{MNIST}, and $0.94$ on \textsc{Caravan} ($\alpha\!=\!10$). Across both families, GA-SRMC matches or improves upon SRMC in $14$ of $16$ pairs, with the only two ESS exceptions being \textsc{German} under MALA ($4173$ vs.\ $4663$) and \textsc{Caravan} under FisherMALA ($6022$ vs.\ $6381$).

\textbf{Stability cap of GA-SRMC vs.\ SRMC.}
Figure~\ref{fig:alpha-sweep} traces the worst-coordinate ESS as $\alpha$ is swept on four representative targets. Solid lines with filled markers denote $\alpha$ values within the stability cap; dotted lines with hollow markers denote unstable $\alpha$ values beyond the cap, where the chain mixes rapidly on the distorted surrogate $\pi_\theta^{(M)}$ but exhibits inflated marginal standard deviations relative to $\pi$; red and blue stars mark the Table~\ref{tab:blr} entries for SRMC and GA-SRMC, respectively. Within the MALA family, GA-SRMC remains stable for one (\textsc{Australian}, \textsc{Heart}) or two (\textsc{Pima}) additional $\alpha$-grid steps beyond SRMC, matches SRMC's cap on \textsc{Ripley}, \textsc{German}, and \textsc{Caravan} (where MALA$+$SRMC's ESS peaks earlier at $\alpha\!=\!5$), and on high-dimensional \textsc{MNIST} ($d\!=\!785$) remains stable through $\alpha\!=\!20$ whereas MALA$+$SRMC exits between $\alpha\!=\!3$ and $\alpha\!=\!10$ (Appendix~\ref{apx:validity}). Under FisherMALA on \textsc{MNIST}, both methods lose stability at $\alpha\!=\!10$ and perform comparably within noise for $\alpha\!\le\!3$, where the cap is imposed.

\textbf{Posterior-mean accuracy.}
Table~\ref{tab:mse} complements worst-coordinate ESS with the per-dimension MSE of the posterior mean $\bar X$ (details in Appendix~\ref{apx:mse}). Both repulsive samplers are substantially more accurate than their base samplers: on \textsc{Ripley}, \textsc{Pima}, \textsc{Heart}, \textsc{Australian}, and \textsc{German}, SRMC already reduces the MSE by $14$--$28\times$ in both families, and GA-SRMC lowers it further in all $16$ full-Fisher comparisons, so that $\text{GA-SRMC}<\text{SRMC}<\text{base}$ holds in every row. The reduction over SRMC is largest on the most anisotropic targets ($1.236$ vs.\ $1827.031$ on \textsc{Gauss-2D}, $55.500$ vs.\ $541.485$ on \textsc{MNIST}, and $10.217$ vs.\ $45.631$ on \textsc{Caravan} under MALA) and smallest on the near-isotropic \textsc{Ripley} ($\kappa\!=\!1.3$; $0.198$ vs.\ $0.205$), where $M_{\rm opt}$ is close to a multiple of the identity and GA-SRMC reduces to SRMC. Notably, even in the two pairs where worst-coordinate ESS favored SRMC (\textsc{German} under MALA and \textsc{Caravan} under FisherMALA), GA-SRMC yields a more accurate posterior mean on every one of the $40$ chains ($0.904$ vs.\ $1.624$ on \textsc{German} and $2.938$ vs.\ $5.949$ on \textsc{Caravan}): worst-coordinate ESS captures only the slowest-mixing coordinate, whereas the MSE aggregates estimation error across all $d$ coordinates. Moreover, adding GA-SRMC to MALA lowers the MSE more than replacing MALA with FisherMALA on all eight targets (by $2.45\times$ on \textsc{MNIST} to $32.9\times$ on \textsc{Ripley}).

\textbf{When SRMC is the safer choice.}
GA-SRMC's advantage on coupled posteriors relies on a faithful Fisher estimate; Appendix~\ref{apx:fisher-fidelity} compares the full $O(d^2)$ estimator against an $O(d)$ diagonal estimator that retains only per-coordinate squared scores. In worst-coordinate ESS (Table~\ref{tab:fisher-fidelity}), the diagonal estimator preserves GA-SRMC's advantage on axis-aligned targets such as \textsc{Gauss-2D} ($2.4\times$ under MALA) and in five of eight rows under diagonal FisherMALA, but flips to SRMC under MALA on the coupled BLR posteriors. In posterior-mean MSE (Table~\ref{tab:mse-diag}), using the diagonal estimate in the history feedback alone (MALA$+$diagonal GA-SRMC) still achieves a lower mean MSE than MALA$+$SRMC on all eight targets at $O(d)$ cost. Reusing the same imperfect diagonal estimate in both the proposal and the repulsion (diagonal FisherMALA$+$diagonal GA-SRMC), in contrast, yields a higher MSE than SRMC on six of the seven BLR targets, consistent with a double penalty when one coarse approximation enters both channels; in that dual-use configuration, SRMC is the safer choice.
\section{Conclusion and Future directions}
\label{sec:conclusion}

We introduced \emph{Geometry-Aware} SRMC, an improvement over SRMC \citep{hu2026srmc}, which substitutes the Euclidean alignment $\theta^\top s(x)$ in the surrogate target with a matrix-weighted expression $\theta^\top M s(x)$, where $M$ is optimally set to the trace-normalized inverse Fisher information. Because the same inverse Fisher already serves as the proposal-side preconditioner in modern Fisher-adaptive samplers~\citep{titsias2023optimal,hird2025Quantifying}, GA-SRMC adds essentially no bookkeeping over a Fisher-preconditioned base. Like SRMC, GA-SRMC is a \emph{wrapper}. In this paper, we apply MALA and FisherMALA to a geometry-aware surrogate target, so the main contribution is essentially a recipe that elevates the Fisher information from a mere proposal-side preconditioner to a geometric quantity that simultaneously shapes both the proposal and the score-history feedback. The same recipe should be applicable to any gradient-based sampler equipped with a Fisher-aware preconditioner.

\textbf{Future directions.} The wrapper view of GA-SRMC opens a broader design space than this paper can explore. A natural next step is the Hamiltonian family, e.g., HMC, Riemannian HMC, and Fisher-adaptive HMC~\citep{girolami2011riemann,seyboldt2026Preconditioning,hoffman2014nuts}, which already use a proposal-side metric. We expect the same dual-role Fisher recipe to extend to them with similar ease, making an empirical study on these Hamiltonian samplers the most immediate follow-up. In addition, the closed-form expression for the limiting covariance in our CLT suggests a self-adjusting repellence schedule that may eliminate the need for a stability sweep over individual targets. Discrete-state spaces and strongly multimodal settings pose deeper, more fundamental issues regarding the appropriate Fisher-like quantity and will require longer-term investigation. We interpret GA-SRMC as an initial move toward a more general geometry-aware framework for history-based MCMC.

\begin{ack}
We thank the anonymous reviewers for their constructive comments and suggestions. Jie Hu’s contribution to this work was supported in part by startup funding and Oakland University’s 2026 Research Committee (URC) Faculty Research Fellowship. Lingyun Chen and Do Young Eun’s contributions were supported in part by National Science Foundation grant IIS-2421484.

\end{ack}

\bibliographystyle{plainnat}
\bibliography{ref}

@article{roberts1996exponential,
  author    = {Roberts, Gareth O. and Tweedie, Richard L.},
  title     = {Exponential Convergence of {L}angevin Distributions
               and Their Discrete Approximations},
  journal   = {Bernoulli},
  volume    = {2},
  number    = {4},
  pages     = {341--363},
  year      = {1996}
}

@article{roberts1998optimal,
  author    = {Roberts, Gareth O. and Rosenthal, Jeffrey S.},
  title     = {Optimal Scaling of Discrete Approximations to
               {L}angevin Diffusions},
  journal   = {Journal of the Royal Statistical Society:
               Series B (Statistical Methodology)},
  volume    = {60},
  number    = {1},
  pages     = {255--268},
  year      = {1998}
}

@article{roberts2001optimal,
  title={Optimal scaling for various {M}etropolis--{H}astings algorithms},
  author={Roberts, Gareth O and Rosenthal, Jeffrey S},
  journal={Statistical Science},
  volume={16},
  number={4},
  pages={351--367},
  year={2001},
  publisher={Institute of Mathematical Statistics}
}

@article{duane1987hybrid,
  author    = {Duane, Simon and Kennedy, Anthony D. and Pendleton,
               Brian J. and Roweth, Duncan},
  title     = {Hybrid {M}onte {C}arlo},
  journal   = {Physics Letters B},
  volume    = {195},
  number    = {2},
  pages     = {216--222},
  year      = {1987}
}

@incollection{neal2011mcmc,
  title={{MCMC} using {H}amiltonian dynamics},
  author={Neal, Radford M},
  booktitle={Handbook of {M}arkov Chain {M}onte {C}arlo},
  pages={113--162},
  year={2011},
  publisher={Chapman and Hall/CRC}
}

@article{girolami2011riemann,
  author    = {Girolami, Mark and Calderhead, Ben},
  title     = {{R}iemann Manifold {L}angevin and {H}amiltonian
               {M}onte {C}arlo Methods},
  journal   = {Journal of the Royal Statistical Society:
               Series B (Statistical Methodology)},
  volume    = {73},
  number    = {2},
  pages     = {123--214},
  year      = {2011}
}

@article{hoffman2014nuts,
  author    = {Hoffman, Matthew D. and Gelman, Andrew},
  title     = {The {N}o-{U}-{T}urn Sampler: Adaptively Setting
               Path Lengths in {H}amiltonian {M}onte {C}arlo},
  journal   = {Journal of Machine Learning Research},
  volume    = {15},
  number    = {47},
  pages     = {1593--1623},
  year      = {2014}
}

@article{geyer1992practical,
  title={Practical {M}arkov Chain {M}onte {C}arlo},
  author={Geyer, Charles J},
  journal={Statistical Science},
  volume={7},
  number={4},
  pages={473--483},
  year={1992},
  publisher={Institute of Mathematical Statistics}
}

@article{haario2001adaptive,
  author    = {Haario, Heikki and Saksman, Eero and Tamminen,
               Johanna},
  title     = {An Adaptive {M}etropolis Algorithm},
  journal   = {Bernoulli},
  volume    = {7},
  number    = {2},
  pages     = {223--242},
  year      = {2001}
}

@article{andrieu2008tutorial,
  author    = {Andrieu, Christophe and Thoms, Johannes},
  title     = {A Tutorial on Adaptive {MCMC}},
  journal   = {Statistics and Computing},
  volume    = {18},
  number    = {4},
  pages     = {343--373},
  year      = {2008}
}

@inproceedings{titsias2023optimal,
  author    = {Titsias, Michalis K.},
  title     = {Optimal Preconditioning and {F}isher Adaptive
               {L}angevin Sampling},
  booktitle = {Advances in Neural Information Processing Systems},
  year      = {2023}
}

@article{laio2002escaping,
  author    = {Laio, Alessandro and Parrinello, Michele},
  title     = {Escaping Free-Energy Minima},
  journal   = {Proceedings of the National Academy of Sciences},
  volume    = {99},
  number    = {20},
  pages     = {12562--12566},
  year      = {2002}
}

@article{barducci2008well,
  author    = {Barducci, Andrea and Bussi, Giovanni and
               Parrinello, Michele},
  title     = {Well-Tempered Metadynamics: {A} Smoothly Converging
               and Tunable Free-Energy Method},
  journal   = {Physical Review Letters},
  volume    = {100},
  pages     = {020603},
  year      = {2008}
}

@article{henin2022enhanced,
  title={Enhanced Sampling Methods for Molecular Dynamics Simulations [Article v1. 0]},
  author={Henin, Jerome and Lelievre, Tony and Shirts, Michael R and Valsson, Omar and Delemotte, Lucie},
  journal={Living Journal of Computational Molecular Science},
  volume={4},
  number={1},
  year={2022}
}

@article{darve2001calculating,
  author    = {Darve, Eric and Pohorille, Andrew},
  title     = {Calculating Free Energies Using Average Force},
  journal   = {Journal of Chemical Physics},
  volume    = {115},
  number    = {20},
  pages     = {9169--9183},
  year      = {2001}
}

@article{wang2001efficient,
  author    = {Wang, Fugao and Landau, David P.},
  title     = {Efficient, Multiple-Range Random Walk Algorithm to
               Calculate the Density of States},
  journal   = {Physical Review Letters},
  volume    = {86},
  number    = {10},
  pages     = {2050--2053},
  year      = {2001}
}

@inproceedings{deng2020contour,
  author    = {Deng, Wei and Lin, Guang and Liang, Faming},
  title     = {A Contour Stochastic Gradient {L}angevin Dynamics
               Algorithm for Simulations of Multi-modal
               Distributions},
  booktitle = {Advances in Neural Information Processing Systems},
  year      = {2020}
}

@article{benaim2016convergence,
  title={Convergence of adaptive biasing potential methods for diffusions},
  author={Benaim, Michel and Brehier, Charles-Edouard},
  journal={Comptes Rendus. Mathematique},
  volume={354},
  number={8},
  pages={842--846},
  year={2016}
}

@article{benaim2018abp,
  title={Convergence analysis of adaptive biasing potential methods for diffusion processes},
  author={Bena{\"\i}m, Michel and Br{\'e}hier, Charles-Edouard},
  journal={Communications in Mathematical Sciences},
  volume={17},
  number={1},
  pages={81--130},
  year={2019},
  publisher={International Press}
}

@article{benaim2020analysis,
  title={Analysis of an Adaptive Biasing Force method based on self-interacting dynamics},
  author={Benaim, Michel and Brehier, Charles-Edouard and Monmarche, Pierre},
  journal={Electronic Journal of Probability},
  volume={25},
  year={2020}
}

@inproceedings{ye2020stein,
  author    = {Ye, Mao and Ren, Tongzheng and Liu, Qiang},
  title     = {Stein Self-Repulsive Dynamics: {B}enefits From
               Past Samples},
  booktitle = {Advances in Neural Information Processing Systems},
  year      = {2020}
}

@inproceedings{doshi2023self,
  author    = {Doshi, Vishwaraj and Hu, Jie and Eun, Do Young},
  title     = {Self-Repellent Random Walks on General Graphs
               --- {A}chieving Minimal Sampling Variance via
               Nonlinear {M}arkov Chains},
  booktitle = {International Conference
               on Machine Learning},
  year      = {2023}
}

@inproceedings{hu2025beyond,
  author    = {Hu, Jie and Ma, Yi-Ting and Eun, Do Young},
  title     = {Beyond Self-Repellent Kernels: History-Driven Target
               Towards Efficient Nonlinear {MCMC} on General
               Graphs},
  booktitle = {International Conference
               on Machine Learning},
  year      = {2025}
}

@article{hu2026srmc,
  title={Score-Repellent Monte Carlo: Toward Efficient Non-Markovian Sampler with Constant Memory in General State Spaces},
  author={Hu, Jie and Chen, Lingyun and Kim, Geeho and Choi, Jinyoung and Han, Bohyung and Eun, Do Young},
  journal={arXiv preprint arXiv:2604.22948 (To appear in ICML 2026)},
  year={2026}
}

@book{borkar2009stochastic,
  author    = {Borkar, Vivek S.},
  title     = {Stochastic Approximation: {A} Dynamical Systems
               Viewpoint},
  publisher = {Cambridge University Press},
  year      = {2009}
}

@article{seyboldt2026Preconditioning,
  title={Preconditioning {H}amiltonian {M}onte {C}arlo by minimizing {F}isher Divergence},
  author={Seyboldt, Adrian and Carlson, Eliot L and Carpenter, Bob},
  journal={arXiv preprint arXiv:2603.18845},
  year={2026}
}

@article{borkar2025ode,
  title={The {ODE} method for asymptotic statistics in stochastic approximation and reinforcement learning},
  author={Borkar, Vivek and Chen, Shuhang and Devraj, Adithya and Kontoyiannis, Ioannis and Meyn, Sean},
  journal={The Annals of Applied Probability},
  volume={35},
  number={2},
  pages={936--982},
  year={2025},
  publisher={Institute of Mathematical Statistics}
}

@article{amari1998natural,
  title={Natural gradient works efficiently in learning},
  author={Amari, Shun-Ichi},
  journal={Neural Computation},
  volume={10},
  number={2},
  pages={251--276},
  year={1998},
  publisher={MIT Press}
}

@article{hird2025Quantifying,
    title = {Quantifying the {Effectiveness} of {Linear} {Preconditioning} in {Markov} {Chain} {Monte} {Carlo}},
    volume = {26},
    number = {119},
    journal = {Journal of Machine Learning Research},
    author = {Hird, Max and Livingstone, Samuel},
    year = {2025},
    pages = {1--51},
}

@article{andrieu2006ergodicity,
  title={On the Ergodicity Properties of Some Adaptive {MCMC} Algorithms},
  author={Andrieu, Christophe and Moulines, {\'E}ric},
  journal={The Annals of Applied Probability},
  pages={1462--1505},
  year={2006},
  publisher={JSTOR}
}

@article{roberts2007coupling,
  title={Coupling and ergodicity of adaptive {M}arkov chain {M}onte {C}arlo algorithms},
  author={Roberts, Gareth O and Rosenthal, Jeffrey S},
  journal={Journal of Applied Probability},
  volume={44},
  number={2},
  pages={458--475},
  year={2007},
  publisher={Cambridge University Press}
}

@article{fort2011convergence,
  title={Convergence of adaptive and interacting {M}arkov chain {M}onte {C}arlo algorithms},
  author={Fort, G and Moulines, E and Priouret, P},
  journal={The Annals of Statistics},
  pages={3262--3289},
  year={2011},
  publisher={JSTOR}
}

@article{durmus2017nonasymptotic,
  title={Nonasymptotic Convergence Analysis For the Unadjusted {L}angevin Algorithm},
  author={Durmus, Alain and Moulines, {\'E}ric},
  journal={The Annals of Applied Probability},
  volume={27},
  number={3},
  pages={1551--1587},
  year={2017}
}

@article{chak2023optimal,
  title={Optimal friction matrix for underdamped {L}angevin sampling},
  author={Chak, Martin and Kantas, Nikolas and Leli{\`e}vre, Tony and Pavliotis, Grigorios A},
  journal={ESAIM: Mathematical Modelling and Numerical Analysis},
  volume={57},
  number={6},
  pages={3335--3371},
  year={2023},
  publisher={EDP Sciences}
}

@article{martens2020new,
  title={New insights and perspectives on the natural gradient method},
  author={Martens, James},
  journal={Journal of Machine Learning Research},
  volume={21},
  number={146},
  pages={1--76},
  year={2020}
}

@inproceedings{wu2024improved,
title={An Improved Empirical Fisher Approximation for Natural Gradient Descent},
author={Xiaodong Wu and Wenyi Yu and Chao Zhang and Phil Woodland},
booktitle={Advances in Neural Information Processing Systems},
pages={134151--134194},
year={2024}
}

@book{kushner2003stochastic,
  title={Stochastic approximation and recursive algorithms and applications},
  author={Kushner, Harold J and Yin, G George},
  year={2003},
  publisher={Springer}
}

@article{carpenter2017stan,
  title={{S}tan: A probabilistic programming language},
  author={Carpenter, Bob and Gelman, Andrew and Hoffman, Matthew D and Lee, Daniel and Goodrich, Ben and Betancourt, Michael and Brubaker, Marcus and Guo, Jiqiang and Li, Peter and Riddell, Allen},
  journal={Journal of Statistical Software},
  volume={76},
  pages={1--32},
  year={2017}
}

@article{abadi2016tensorflow,
  title = {{T}ensor{F}low: Large-Scale Machine Learning on Heterogeneous Distributed Systems},
  author = {Martín Abadi and Ashish Agarwal and Paul Barham and Eugene Brevdo and Zhifeng Chen and Craig Citro and Greg S. Corrado and Andy Davis and Jeffrey Dean and Matthieu Devin and Sanjay Ghemawat and Ian Goodfellow and Andrew Harp and Geoffrey Irving and Michael Isard and Yangqing Jia and Rafal Jozefowicz and Lukasz Kaiser and Manjunath Kudlur and Josh Levenberg and Dan Mane and Rajat Monga and Sherry Moore and Derek Murray and Chris Olah and Mike Schuster and Jonathon Shlens and Benoit Steiner and Ilya Sutskever and Kunal Talwar and Paul Tucker and Vincent Vanhoucke and Vijay Vasudevan and Fernanda Viegas and Oriol Vinyals and Pete Warden and Martin Wattenberg and Martin Wicke and Yuan Yu and Xiaoqiang Zheng},
  year = {2016},
  journal = {CoRR}
}

@article{pearlmutter1994fast,
  title={Fast exact multiplication by the {H}essian},
  author={Pearlmutter, Barak A},
  journal={Neural Computation},
  volume={6},
  number={1},
  pages={147--160},
  year={1994},
  publisher={MIT Press}
}

@book{griewank2008evaluating,
  title={Evaluating derivatives: principles and techniques of algorithmic differentiation},
  author={Griewank, Andreas and Walther, Andrea},
  year={2008},
  publisher={SIAM}
}

@article{mira2013zero,
  title={Zero variance {M}arkov chain {M}onte {C}arlo for {B}ayesian estimators},
  author={Mira, Antonietta and Solgi, Reza and Imparato, Daniele},
  journal={Statistics and Computing},
  volume={23},
  number={5},
  pages={653--662},
  year={2013},
  publisher={Springer}
}

@article{assaraf1999zero,
  title={Zero-variance principle for {M}onte {C}arlo algorithms},
  author={Assaraf, Roland and Caffarel, Michel},
  journal={Physical Review Letters},
  volume={83},
  number={23},
  pages={4682},
  year={1999},
  publisher={APS}
}

@article{oates2017control,
  title={Control functionals for {M}onte {C}arlo integration},
  author={Oates, Chris J and Girolami, Mark and Chopin, Nicolas},
  journal={Journal of the Royal Statistical Society Series B: Statistical Methodology},
  volume={79},
  number={3},
  pages={695--718},
  year={2017},
  publisher={Oxford University Press}
}

@inproceedings{dagreou2024howtocompute,
  author = {Dagréou, Mathieu and Ablin, Pierre and Vaiter, Samuel and Moreau, Thomas},
  title = {How to compute {H}essian-vector products?},
  booktitle = {ICLR Blogposts 2024},
  year = {2024},
  date = {May 7, 2024},
  url  = {https://iclr-blogposts.github.io/2024/blog/bench-hvp/}
}

@book{boyd2004convex,
  title={Convex Optimization},
  author={Boyd, Stephen and Vandenberghe, Lieven},
  year={2004},
  publisher={Cambridge University Press}
}

@article{arviz2019,
  title={{ArviZ} a unified library for exploratory analysis of {B}ayesian models in {P}ython},
  author={Kumar, Ravin and Carroll, Colin and Hartikainen, Ari and Martin, Osvaldo},
  journal={Journal of Open Source Software},
  volume={4},
  number={33},
  pages={1143},
  year={2019},
  publisher={The Open Journal}
}

@book{horn2012matrix,
  title={Matrix analysis},
  author={Horn, Roger A and Johnson, Charles R},
  year={2012},
  publisher={Cambridge university press}
}

@book{bhatia2013matrix,
  title={Matrix analysis},
  author={Bhatia, Rajendra},
  year={2013},
  publisher={Springer Science \& Business Media}
}

@book{ripley2007pattern,
  title={Pattern recognition and neural networks},
  author={Ripley, Brian D},
  year={2007},
  publisher={Cambridge university press}
}

@misc{dua2019uci,
  author       = {Markelle Kelly, Rachel Longjohn, Kolby Nottingham},
  title        = {The UCI Machine Learning Repository},
  howpublished = {\url{https://archive.ics.uci.edu}},
  institution  = {University of California, Irvine, School of Information and Computer Sciences},
  year = {2007}
}

@inproceedings{vanderputten2000caravan,
  title     = {{C}o{IL} {C}hallenge 2000: The Insurance Company Case},
  author    = {van der Putten, Peter and van Someren, Maarten},
  booktitle = {Technical Report 2000--09, Sentient Machine Research, Amsterdam},
  year      = {2000}
}

@article{lecun1998gradient,
  author={Lecun, Y. and Bottou, L. and Bengio, Y. and Haffner, P.},
  journal={Proceedings of the IEEE}, 
  title={Gradient-based learning applied to document recognition}, 
  year={1998},
  volume={86},
  number={11},
  pages={2278-2324}
}

@book{folland1999real,
  title={Real Analysis: Modern Techniques and Their Applications},
  author={Folland, Gerald B.},
  edition={2nd},
  year={1999},
  publisher={John Wiley \& Sons}
}

@book{rudin1976principles,
  title={Principles of Mathematical Analysis},
  author={Rudin, Walter},
  edition={3rd},
  year={1976},
  publisher={McGraw-Hill}
}

@book{lee2013smooth,
  title={Introduction to Smooth Manifolds},
  author={Lee, John M.},
  edition={2nd},
  year={2013},
  publisher={Springer}
}

@book{khalil2002nonlinear,
  title={Nonlinear Systems},
  author={Khalil, Hassan K.},
  edition={3rd},
  year={2002},
  publisher={Prentice Hall}
}

@book{billingsley1999convergence,
  title={Convergence of Probability Measures},
  author={Billingsley, Patrick},
  edition={2nd},
  year={1999},
  publisher={John Wiley \& Sons}
}

@book{van1998asymptotic,
  title={Asymptotic Statistics},
  author={van der Vaart, Aad W.},
  year={1998},
  publisher={Cambridge University Press}
}

\newpage
\appendix

\section{Additional Experimental Notes}
\label{app:additional-experiments}

\paragraph{Summary of Appendix \ref{app:additional-experiments}.}
This appendix collects the experimental details that did not fit into Section~\ref{sec:experiments}, organized into six subsections. Appendix~\ref{apx:fig1-setup} documents the complete setup for Figure~\ref{fig:geometry-blind} on the 2D anisotropic Gaussian: it defines the target, explains the five panels in the first row, and provides unified pseudocode that specializes Algorithm~\ref{alg:online-fisher-ga-srmc} to each of the five samplers. Appendix~\ref{apx:blr-target} writes out the explicit Bayesian logistic-regression (BLR) posterior \eqref{eq:blr-posterior} sampled from in all seven real-world experiments, together with the data preprocessing and the score and Hessian \eqref{eq:blr-score-hessian} that drive the proposal, the surrogate target, and the stability diagnostic. Appendix~\ref{apx:lambda-tuning} reports the sensitivity sweep over the regularizer $\delta$ in $\widehat S_{n,\delta}\!=\!\widehat S_n+\delta I_d$, and shows that the choice $\delta\!=\!0.01$ used in all main-text experiments outperforms the conventional $\delta\!=\!10$ of \citet{titsias2023optimal} on every tested target, with a gain that generally grows with the posterior condition number. Appendix~\ref{apx:validity} details the per-target stability cap on the repellence strength $\alpha$ used in Table~\ref{tab:blr}, which is computed via the standardized-marginal diagnostic $\max_j(\widehat\sigma_j/\sigma_{{\rm Laplace},j})\!\le\!1.30$ and is determined separately for each (target, method) pair. Appendix~\ref{apx:mse} specifies the posterior-mean MSE evaluation used in Tables~\ref{tab:mse} and~\ref{tab:mse-diag} and explains how worst-coordinate ESS and posterior-mean MSE complement each other. Finally, Appendix~\ref{apx:fisher-fidelity} compares diagonal and full Fisher estimates across all $8$ targets, two sampler families, and three repellence variants in both worst-coordinate ESS and posterior-mean MSE, identifying when a full Fisher estimate is required and when SRMC is the safer choice.

\subsection{Setup for Figure~\ref{fig:geometry-blind} (anisotropic Gaussian)}
\label{apx:fig1-setup}

\paragraph{Target.}
We use a 2D axis-aligned anisotropic Gaussian
\[
\pi(x_1,x_2)=\gN(0,\Sigma),\qquad \Sigma=\diag(\sigma_1^2,\sigma_2^2),\qquad (\sigma_1,\sigma_2)=(5,\,0.5),
\]
so the per-coordinate condition number is $\kappa=(\sigma_1/\sigma_2)^2=100$.  This clearly illustrates the geometry-blind effect: the $x_1$ direction has a small score, $|s_1(x)| = |x_1|/25$, whereas the $x_2$ direction has a large score, $|s_2(x)| = 4|x_2|$. Consequently, using a single repellence strength $\alpha$ on the raw inner product $\theta^\top s(x)$ pushes too hard along $x_2$ before it has any meaningful influence along $x_1$. All chains are initialized at $X_0 = (6,\,3)^\top$, which is off-center with respect to both axes, so the burn-in phase must traverse both the wide and the narrow directions.

Choosing $\kappa = 100$ makes the geometry-blind effect clearly observable: a single-$\alpha$ SR-MALA cannot properly handle both axes at once. For a milder setting $\kappa = 10$, the geometry-blind tilt from SRMC is still inferior to GA-SRMC, but the worst-coordinate ESS gap is not that significant. For a more extreme $\kappa = 10^4$, the standard MALA chain essentially never explores the slow axis, so the study reduces to comparing the underlying samplers alone. Thus, $\kappa = 100$ provides a sweet spot where it is simultaneously evident both that preconditioning helps and that GA-SRMC strictly outperforms SRMC for a hard target density.

\paragraph{Interpretation of row 1 (one-step schematic).}
The five panels in the top row of Figure~\ref{fig:geometry-blind} are constructed around a common reference configuration consisting of an interior state $X_n$, the recent-history samples $\{X_{k}\}_{k=1}^{n-1}$, and the local score field $s(X)=-\Sigma^{-1}X$. The panels differ exclusively with respect to: (a) the surrogate target $\nu$ implicitly targeted by the proposal kernel, (b) the presence or absence of Langevin preconditioning, and (c) whether the surrogate exponent uses the alignment $\theta^\top s$ or $\theta^\top Ms$. For each panel, define the empirical average of past scores
\[
\theta \triangleq \frac{1}{n-1}\sum_{k=1}^{n-1} s(X_k).
\]
The individual cases are as follows:
\begin{itemize}[leftmargin=1.6em,itemsep=-0.05em]
\item \textbf{(i) MALA.} The surrogate coincides with the target, $\nu=\pi$, and no tilting term is introduced. The proposal covariance is isotropic, with kernel
\[
q_I(\cdot\mid X)=\gN\!\bigl(X+\epsilon\,s(X),\,2\epsilon I_d\bigr).
\]
Accordingly, the schematic displays the unmodified score vector $s$ together with a circular noise contour, emphasizing that the drift direction is governed by $s$ and the stochastic scale by $\epsilon$.

\item \textbf{(ii) FisherMALA \citep{titsias2023optimal}.} Again $\nu=\pi$ and no tilt is applied, but the proposal is preconditioned:
\[
q_A(\cdot\mid X)=\gN\!\bigl(X+\epsilon\,A\,s(X),\,2\epsilon\,A\bigr),
\]
where the preconditioner $A\propto S^{-1}$ is the trace-normalized inverse Fisher information. The schematic depicts both $s$ and the preconditioned direction $As$ (with a dashed arc between them), as well as an anisotropic noise ellipse oriented according to $A$, indicating that preconditioning reorients the drift toward the slow principal axis.

\item \textbf{(iii) MALA + GA-SRMC.} The proposal noise remains isotropic, but the kernel targets the surrogate
\[
\nu=\pi_\theta^{(M)} \propto \pi(x)\exp\bigl(-\alpha\,\theta^\top M s(x)\bigr),
\qquad M=A,
\]
so that the drift is determined by the surrogate score
\[
s_\theta^{(M)}(x)\triangleq \nabla_x\log\pi_\theta^{(M)}(x)
= s(x)-\alpha\,\nabla_x^2\log\pi(x)\,M\,\theta.
\]
The schematic highlights the modified drift vector $s_\theta^{(M)}$ pointing away from the recent-history cloud, and a shifted (green dashed) level set of $\pi_\theta^{(M)}$, illustrating that score-history induces a repulsive displacement of the surrogate mass away from previously visited regions.

\item \textbf{(iv) FisherMALA + SRMC.} Here the surrogate uses a geometry-blind (Euclidean) alignment,
\[
\nu=\pi_\theta \propto \pi(x)\exp\bigl(-\alpha\,\theta^\top s(x)\bigr),
\qquad M=I_d,
\]
with corresponding surrogate score
\[
s_\theta(x)=s(x)-\alpha\,\nabla_x^2\log\pi(x)\,\theta,
\]
while the proposal remains preconditioned by $A$. The schematic shows both $s_\theta$ and $As_\theta$ (dashed arc), the shifted (red dashed) level set of $\pi_\theta$ displaced along the dominant-score direction (here $x_2$, associated with higher curvature), and an anisotropic noise ellipse. Because the tilt direction implied by $M=I_d$ need not coincide with the geometry imposed by $A$, part of the Metropolis correction is expended compensating for a misaligned surrogate-induced drift.

\item \textbf{(v) FisherMALA + GA-SRMC.} The surrogate is $\nu=\pi_\theta^{(M)}$ with matched geometry $M=A$ (the same trace-normalized inverse Fisher used for preconditioning). The schematic indicates that the surrogate displacement (green dashed level set) and the preconditioned drift align with the slow $x_1$ axis: the repulsion induced by score history and the proposal anisotropy act coherently, simultaneously biasing motion toward the slow direction and allocating proposal noise along that direction. This is the only configuration in which both the surrogate shift and the proposal covariance consistently promote exploration along the slow axis.
\end{itemize}
In particular, panels~(iv) and~(v) share the proposal-side preconditioner $A$ (the trace-normalized inverse Fisher) but differ in the repulsion-side matrix $M$: panel~(iv) uses $M\!=\!I_d$ (SRMC), while panel~(v) uses $M=A$ (GA-SRMC). The diagram makes the consequence visible: in panel~(iv) the red-dashed surrogate level set of $\pi_\theta$ shifts along the dominant-score axis (the narrow $x_2$-direction), \emph{against} the proposal preconditioner that stretches noise along the slow $x_1$-direction; the chain then spends acceptance budget undoing the wrong-axis tilt. In panel~(v), the green-dashed surrogate level set of $\pi_\theta^{(M)}$ shifts along the slow $x_1$-direction, so the score-history correction reinforces rather than opposes the proposal anisotropy. This is exactly the alignment that Theorem~\ref{thm:optimal} promotes, from Figure \ref{fig:geometry-blind} observation to a closed-form universal upper bound on $\Sigma_{\theta\theta}^{(M)}$.

\paragraph{Unified pseudocode for the five samplers.}
The five panels of Figure~\ref{fig:geometry-blind} differ only in two parts: whether the proposal is preconditioned ($\textsf{precond}\!\in\!\{\textsf{none},\textsf{Fisher}\}$, controlling $A_n$) and which surrogate target is sampled ($\textsf{tilt}\!\in\!\{\textsf{none},\textsf{SRMC},\textsf{GA-SRMC}\}$, controlling $M_n$).  Algorithm~\ref{alg:fig1-unified} below is a five-way specialization of the main-body Algorithm~\ref{alg:online-fisher-ga-srmc} (Section~\ref{sec:ga-srmc-practical}): it inherits the same online Fisher state, the same trace normalization, and the same SA recursion, but covers all five panels by branching on $(\textsf{precond},\textsf{tilt})$ via if-else.  Table~\ref{tab:fig1-config} lists the corresponding configurations; in particular, panel~(v) (\textsf{Fisher},\textsf{GA-SRMC}) is exactly Algorithm~\ref{alg:online-fisher-ga-srmc} with $A_n\!=\!M_n$, while the other four panels are obtained by switching off either the proposal-side preconditioning or the repulsion-side metric. All variants share the same constant-step-size Robbins–Monro update for $\log\epsilon$ and (where used) the same online Fisher state $\widehat S_n$. As in Algorithm \ref{alg:online-fisher-ga-srmc}, the metric used at iteration $n$ is computed from $\hat S_n$ and the Fisher state is updated with the score at the new sate $X_{n+1}$, using the same step size $\gamma_{n+1}$ as the score history. We write $\xi_n\!\sim\!\gN(0,I_d)$, $a_n$ for the realized Metropolis acceptance probability, and $\widehat M_{n}\!\triangleq\!(\tr(\widehat S_{n,\delta})/d)\widehat S_{n,\delta}^{-1}$ for the trace-normalized inverse Fisher. In this paper, the Hessian-vector product $\nabla_x^2\log\pi(X_n)\,M_{n}\theta_n$ never materializes the $d\!\times\!d$ Hessian: it is an elementwise multiplication for the diagonal Gaussian target here and is approximated via automatic differentiation \citep{pearlmutter1994fast,griewank2008evaluating} or via finite differences of gradients $\nabla_x^2\log\pi(x)\, v \approx (\nabla_x \log \pi(x + \zeta v) - \nabla_x \log \pi(x)) / \zeta$ as in \citep{dagreou2024howtocompute}.
\begin{algorithm}[ht]
\caption{Unified per-iteration update covering all five panels of Figure~\ref{fig:geometry-blind}.}
\label{alg:fig1-unified}
\begin{algorithmic}[1]
\Require state $X_n$, score history $\theta_n$ (used only when $\textsf{tilt}\!\ne\!\textsf{none}$), Fisher state $\widehat S_n$, switches $(\textsf{precond},\textsf{tilt})$ from Table~\ref{tab:fig1-config}.
\State Compute $\widehat M_n$ from $\widehat S_n$ (needed only if $\textsf{precond}=\textsf{Fisher}$ or $\textsf{tilt}=\textsf{GA-SRMC}$).
\State \textbf{Proposal preconditioner:} \quad $A_{n}\gets\widehat M_{n}$ if $\textsf{precond}=\textsf{Fisher}$, \;\, else $A_{n}\gets I_d$.
\State \textbf{Repulsion metric:} \quad $M_{n}\gets\widehat M_{n}$ if $\textsf{tilt}=\textsf{GA-SRMC}$, \;\, else $M_{n}\gets I_d$.
\If{$\textsf{tilt}=\textsf{none}$} \Comment{Langevin drift}
    \State $\widetilde s_n\gets s(X_n)$.
\Else
    \State $\widetilde s_n\gets s(X_n)-\alpha\,\nabla_x^2\log\pi(X_n)\,M_{n}\theta_n$ \Comment{Surrogate score computation}
\EndIf
\State \textbf{Propose} $Y\gets X_n+\epsilon\,A_{n}\widetilde s_n+\sqrt{2\epsilon}\,A_{n}^{1/2}\xi_n$.
\If{$\textsf{tilt}=\textsf{none}$} \Comment{Metropolis acceptance}
    \State $a_n\gets\min\{1,\,\pi(Y)q_{A_{n}}(X_n|Y)\,/\,[\pi(X_n)q_{A_{n}}(Y|X_n)]\}$.
\Else
    \State $a_n$ via Eq.~\eqref{eq:accept-ratio} with $A=A_{n}$ and $M=M_{n}$.
\EndIf
\State Update $X_{n+1}\!\gets\!Y$ with prob.\ $a_n$ else $X_{n+1}\!\gets\!X_n$; \;\, $\log\epsilon\!\gets\!\log\epsilon+c_\epsilon(a_n-0.574)$.
\If{$\textsf{tilt}\ne\textsf{none}$} \Comment{Score-history SA update}
    \State $\theta_{n+1}\gets\theta_n+\gamma_{n+1}(s(X_{n+1})-\theta_n)$.
\EndIf
\If{$\textsf{precond}=\textsf{Fisher}$ \textbf{or} $\textsf{tilt}=\textsf{GA-SRMC}$} \Comment{Online Fisher update}
    \State $\widehat S_{n+1}\gets(1\!-\!\gamma_{n+1})\widehat S_n+\gamma_{n+1}\,\diag(s(X_{n+1})s(X_{n+1})^\top)$.
\EndIf
\end{algorithmic}
\end{algorithm}

The bottom-row panels of Figure~\ref{fig:geometry-blind} are the empirical sample clouds and per-coordinate trace plots produced by Algorithm~\ref{alg:fig1-unified} under the five configurations of Table~\ref{tab:fig1-config}.  Effective sample size is computed coordinate-wise using Geyer's initial positive-sequence estimator \citep{geyer1992practical}; the reported pairs $(\mathrm{ESS}(x_1),\mathrm{ESS}(x_2))$ correspond to the post-burn-in portion of $N=1000$ samples.

\begin{table}[ht]
\caption{Configuration switches for Algorithm~\ref{alg:fig1-unified} corresponding to the five panels of Figure~\ref{fig:geometry-blind}; $\widehat M_n\propto\widehat S_{n,\delta}^{-1}$ is the trace-normalized inverse Fisher.}
\label{tab:fig1-config}
\centering
\small
\setlength{\tabcolsep}{6pt}
\begin{tabular}{clcccc}
\toprule
Panel & Method & $\textsf{precond}$ & $\textsf{tilt}$ & $A_{n}$ & $M_{n}$ \\
\midrule
(i)   & MALA               & none   & none   & $I_d$           & ---             \\
(ii)  & FisherMALA         & Fisher & none   & $\widehat M_n$ & ---             \\
(iii) & MALA + GA-SRMC       & none   & GA-SRMC  & $I_d$           & $\widehat M_n$ \\
(iv)  & FisherMALA + SRMC    & Fisher & SRMC     & $\widehat M_n$ & $I_d$           \\
(v)   & FisherMALA + GA-SRMC & Fisher & GA-SRMC  & $\widehat M_n$ & $\widehat M_n$ (shared) \\
\bottomrule
\end{tabular}
\end{table}

\subsection{Bayesian logistic regression: target distribution}
\label{apx:blr-target}
The seven real-world targets in Section~\ref{sec:experiments} (Ripley, Pima, Heart, Australian, German, Caravan, MNIST) are all Bayesian logistic-regression (BLR) posteriors over the parameter $\beta\!\in\!\mathbb R^d$, defined as follows. Given a dataset $\{(x_i,y_i)\}_{i=1}^N$ with feature vectors $x_i\!\in\!\mathbb R^d$ and binary labels $y_i\!\in\!\{0,1\}$, we collect the features into the design matrix $\Phi\!\in\!\mathbb R^{N\times d}$ whose $i$-th row is $x_i^\top$. Each $x_i$ consists of an intercept entry followed by the covariates; for all datasets except MNIST, every covariate is z-scored (centered and scaled to unit variance), and the whole design matrix is then divided by $\sqrt N$, which keeps the score magnitude comparable across datasets of very different sizes. For MNIST, which is binarized to a two-class problem in the spirit of \citet{titsias2023optimal}, we use the digits $3$ versus $8$, draw a random subsample of $N\!=\!4000$ images, and keep the raw pixel intensities scaled to $[0,1]$ without z-scoring; with the intercept this gives $d\!=\!785$, and the design is again divided by $\sqrt N$. Combining the standard logistic likelihood with the isotropic Gaussian prior $\pi_0(\beta)\!=\!\gN(\beta;0,\tau_0^{-1} I_d)$ of precision $\tau_0\!>\!0$ gives the (unnormalized) BLR posterior
\begin{equation}
\label{eq:blr-posterior}
\pi(\beta)\;\propto\;\exp\!\left\{-\frac{\tau_0}{2}\|\beta\|^2+\sum_{i=1}^N\!\left[y_i\,x_i^\top\beta-\log\!\left(1+e^{x_i^\top\beta}\right)\right]\right\},
\end{equation}
which is the target sampled from in all BLR experiments of Section~\ref{sec:experiments}, with the prior precision fixed to $\tau_0\!=\!0.1$ (prior standard deviation $\sqrt{10}$) throughout. The corresponding score and Hessian are
\begin{equation}
\label{eq:blr-score-hessian}
s(\beta)\;=\;-\tau_0\beta+\Phi^\top\!\bigl(y-\hat p(\beta)\bigr),\qquad -\nabla^2\!\log\pi(\beta)\;=\;\tau_0I_d+\Phi^\top D(\beta)\Phi,
\end{equation}
where $\hat p_i(\beta)\!=\!\sigma(x_i^\top\beta)\!=\!1/(1+e^{-x_i^\top\beta})$ and $D(\beta)\!=\!\diag(\hat p_i(\beta)(1-\hat p_i(\beta)))$. Evaluating \eqref{eq:blr-score-hessian} at the MAP $x^\star\!=\!\arg\max_\beta\pi(\beta)$ yields the observed Fisher $Z\!=\!-\nabla^2\!\log\pi(x^\star)$, which initializes the Fisher state in all BLR experiments (Section~\ref{sec:experiments}) and is used in the standardized-marginal stability diagnostic of Appendix~\ref{apx:validity} (Eq.~\eqref{eq:sigma-laplace}); the off-diagonal coupling through $\Phi^\top D(x^\star)\Phi$ is what governs the diagonal-vs-full-Fisher gap discussed in Appendices~\ref{apx:fisher-fidelity} and~\ref{app:fisher-diag}.

\subsection{Choice of the regularizer \texorpdfstring{$\delta$}{delta}}
\label{apx:lambda-tuning}
The online inverse-Fisher state employed by GA-SRMC and FisherMALA is based on the regularized matrix $\widehat S_{n,\delta}\!=\!\widehat S_n+\delta I_d$. The scalar $\delta$ therefore bounds the eigenvalues of $\widehat S_{n,\delta}^{-1}$ above by $1/\delta$ and, equivalently, governs the effective proposal step size in directions where $\widehat S_n$ has small eigenvalues. The original FisherMALA work~\citep{titsias2023optimal} fixes $\delta=10$. We performed a sweep over $\delta\in\{0.01,0.1,1.0,10.0\}$ on seven targets, running each target at the value $\alpha^\star$ listed in Table~\ref{tab:lambda-sweep} (taken from an earlier tuning round restricted to $\alpha\le20$) and holding all other hyperparameters constant; Table~\ref{tab:lambda-sweep} shows the resulting worst-coordinate ESS for FisherMALA+GA-SRMC. Its $\delta=0.01$ column therefore coincides with Table~\ref{tab:blr} on Heart, German, and Caravan, where $\alpha^\star$ equals the Table~\ref{tab:blr} value, but not on Ripley, Pima, Australian, and Gauss-2D, where Table~\ref{tab:blr} uses $\alpha=30$, $25$, $25$, and $20$, respectively. Across all tested targets, the smallest $\delta$ in this grid yields the highest ESS, and the improvement relative to the conventional $\delta=10$ generally grows with the posterior's condition number. On Ripley, whose posterior is nearly isotropic, the gain is modest ($16\%$); on Gauss-2D, it exceeds three orders of magnitude. Mechanistically, this is because a smaller $\delta$ allows the preconditioner $A_n=(d/\tr \widehat S_{n,\delta}^{-1})\, \widehat S_{n,\delta}^{-1}$ to more faithfully approximate the true inverse Fisher, so the sampler can take larger steps along low-curvature directions and mix much faster in those components. We did not explore $\delta<0.01$, the smallest value in the sweep, and we adopt $\delta=0.01$ in all main-text experiments.

\begin{table}[t]
\caption{Worst-direction ESS for FisherMALA+GA-SRMC across $\delta$ values (mean$\pm$std over $40$ replicate chains, $20000$ post-burn-in samples) at each target's stability-capped best $\alpha$. The smallest $\delta$ tested is preferred on every row; bold marks the row maximum.}
\label{tab:lambda-sweep}
\centering
\begin{tabular}{lcrrrrr}
\toprule
Target & $\kappa$ & $\alpha^\star$ & $\delta=0.01$ & $\delta=0.1$ & $\delta=1$ & $\delta=10$ \\
\midrule
Ripley     & $1.3$  & $20$ & \textbf{13958}$\pm$138 & 13574$\pm$197 & 12500$\pm$262 & 12027$\pm$262 \\
Heart      & $2.7$  & $20$ & \textbf{9549}$\pm$159  &  8716$\pm$162 &  6982$\pm$168 &  6402$\pm$148 \\
Pima       & $2.7$  & $20$ & \textbf{11164}$\pm$193 & 10319$\pm$217 &  8409$\pm$143 &  7719$\pm$128 \\
Australian & $2.8$  & $20$ & \textbf{9148}$\pm$124  &  7699$\pm$150 &  5480$\pm$127 &  4950$\pm$118 \\
German     & $4.5$  & $15$ & \textbf{9449}$\pm$238  &  7798$\pm$184 &  5354$\pm$155 &  4763$\pm$210 \\
Caravan    & $16.3$ & $10$ & \textbf{6022}$\pm$172  &  4693$\pm$116 &  1096$\pm$1177 &   317$\pm$89 \\
Gauss-2D   & $100$  & $\;3$ & \textbf{12695}$\pm$173 &  4995$\pm$106 &   671$\pm$37 &   $\;$10$\pm$40 \\
\bottomrule
\end{tabular}
\end{table}

\subsection{Per-target stability caps from \texorpdfstring{$\widehat\sigma/\sigma_{\rm Laplace}$}{widehat-sigma/sigma-Laplace}}
\label{apx:validity}
A score-repellent sampler at fixed $\theta$ in fact samples from the surrogate distribution $\pi_\theta^{(M)}(x)\propto\pi(x)\exp(-\alpha\theta^\top M s(x))$, not from $\pi$ itself. These two coincide only in the limiting regime $\theta_n\!\to\!\theta^\star\!=\!0$ guaranteed by Proposition~\ref{prop:fixedM-structure}. For a finite number of iterations $n$ and sufficiently large $\alpha$, the chain may temporarily explore a distorted target, in which case comparing effective sample sizes across samplers would not be meaningful. To guard against this, we use the standardized-marginal diagnostic of~\citet{titsias2023optimal} originally proposed for FisherMALA validation. After burn-in, we compute the per-coordinate empirical standard deviation $\widehat\sigma_j$ of the $j$-th component of the chain (based on the pooled sample $\{X_n^{(j)}\}_{n=N_{\rm burn}+1}^{N_{\rm burn}+N_{\rm collect}}$ across the 40 replicate chains) and compare it to the Laplace prediction
\begin{equation}
\label{eq:sigma-laplace}
\sigma_{{\rm Laplace},j}\;=\;\sqrt{(Z^{-1})_{jj}},\qquad Z\;\triangleq\;-\nabla^2\!\log\pi(x^\star)\;\succ\;0,
\end{equation}
where $x^\star\!=\!\arg\max_x\pi(x)$ is the MAP, $Z$ is the Hessian of the negative log-density at $x^\star$ (also known as the observed Fisher information at $x^\star$, and equal to $\Cov_\pi(s,s)\!=\!S$ when $\pi$ is Gaussian), and $Z^{-1}$ denotes its inverse. The $j$-th diagonal entry $(Z^{-1})_{jj}$ approximates the $j$-th marginal posterior variance under the Gaussian Laplace approximation $\pi(x)\!\approx\!\gN(x;x^\star,Z^{-1})$. We define the stability cap on $\alpha$ as the largest value in our grid for which $\max_j(\widehat\sigma_j/\sigma_{{\rm Laplace},j})\!\le\!1.30$, meaning that it flags the chain as unstable as soon as any marginal standard deviation inflates by more than $30\%$ relative to the Laplace benchmark. The ratio is rounded to two decimals before this comparison; three cells of Table~\ref{tab:blr} pass only after this rounding (GA-SRMC on Australian and Pima under FisherMALA at $\alpha\!=\!25$, and on Pima under MALA at $\alpha\!=\!30$).

The caps in Table~\ref{tab:blr} are determined separately for each method. We swept over $\alpha\!\in\!\{10^{-4},10^{-3},10^{-2},0.1,1,3,5,10,15,20\}$ on all targets, extended by $\alpha\!\in\!\{25,30,50\}$ on the six BLR targets other than MNIST, and recorded the std-ratio for every configuration. For each (target, method) pair, the cap is the largest $\alpha$ in this set that still satisfies the std-ratio condition, and Table~\ref{tab:blr} reports each method at its best-ESS $\alpha$ at or below its cap; for example, MALA$+$SRMC is reported at $\alpha\!=\!5$ on Caravan and at $\alpha\!=\!1$ on MNIST, below its caps of $10$ and $3$. For SRMC, this procedure yields caps of $\alpha\!=\!30$ on Ripley, $\alpha\!=\!25$ on Australian for FisherMALA alone, $\alpha\!=\!20$ for the remaining Australian chains as well as Heart and Pima, $\alpha\!=\!15$ on German, $\alpha\!=\!10$ on Caravan, and $\alpha\!=\!3$ on MNIST. For GA-SRMC, the caps are $\alpha\!=\!30$ on Ripley and on Pima under MALA, $\alpha\!=\!25$ on Australian, on Heart under MALA, and on Pima under FisherMALA, $\alpha\!=\!20$ on Heart under FisherMALA, $\alpha\!=\!15$ on German, and $\alpha\!=\!10$ on Caravan; within the MALA family, they are thus higher than SRMC's on Australian, Heart, and Pima and equal on Ripley, German, and Caravan. On MNIST MALA the GA-SRMC cap is $\alpha\!=\!20$, with std-ratios at $\alpha\in\{3,5,10,15,20\}$ equal to $\{0.96,0.93,0.90,0.88,0.87\}$, decreasing monotonically as $\alpha$ grows. In contrast, SRMC on MNIST leaves the stability regime between $\alpha\!=\!3$ (std-ratio $1.14$) and $\alpha\!=\!10$ (std-ratio $1.83$). This discrepancy in caps matches the mechanism described in Section~\ref{sec:experiments}: using MALA as the baseline, GA-SRMC's weighted tilt distributes repulsion according to the Fisher geometry, allowing the chain to remain well-behaved as $\alpha$ increases, whereas SRMC's unscaled tilt overemphasizes the coordinate with the largest score.

\subsection{Posterior-mean MSE metric}
\label{apx:mse}
Tables~\ref{tab:mse} and~\ref{tab:mse-diag} measure Monte Carlo accuracy by the error of the posterior-mean estimate. Each configuration is run as $40$ independent chains with $2{,}000$ burn-in iterations and $20{,}000$ retained samples, matching the ESS metric. For chain $b$, let $\bar X^{(b)}$ be the plain, equally weighted average of its $20{,}000$ retained samples, i.e., the estimator of $\E_\pi[X]$ obtained with $f(x)\!=\!x$; this is the estimator we evaluate, and it differs from the SA estimate $\mu_N$ returned by Algorithm~\ref{alg:online-fisher-ga-srmc}. Each table entry reports the mean $\pm$ standard deviation over the $40$ chains of $\frac{1}{d}\|\bar X^{(b)}-\mu_{\rm ref}\|_2^2$, in units of $10^{-4}$. For the BLR targets, the reference mean $\mu_{\rm ref}$ is computed from four NUTS chains~\citep{hoffman2014nuts}, each with $2{,}000$ warm-up iterations and $200{,}000$ retained draws ($800{,}000$ draws in total); for Gauss-2D, $\mu_{\rm ref}$ is the exact mean $0$. Because the MSE is non-negative and right-skewed (most visibly for MALA and MALA$+$SRMC on Gauss-2D), $\pm$ one standard deviation describes the spread across runs rather than a symmetric confidence interval. The MALA and MALA$+$SRMC columns of Table~\ref{tab:mse-diag}, which do not use a Fisher estimate, are carried over from Table~\ref{tab:mse}; its four diagonal-Fisher configurations come from a separate set of $40$ runs. Comparisons between these two sets of runs, including those between Tables~\ref{tab:mse} and~\ref{tab:mse-diag}, are therefore descriptive. On Caravan, where the two MALA-family samplers are evaluated at their respective best-ESS values ($\alpha\!=\!5$ for SRMC and $\alpha\!=\!10$ for GA-SRMC) in Tables~\ref{tab:blr} and~\ref{tab:mse}, GA-SRMC also achieves a lower MSE at the common value $\alpha\!=\!5$ ($24.468$ vs.\ $45.631$, lower on all $40$ chains).

\paragraph{ESS and MSE are complementary.}
The per-chain worst-coordinate ESS of Tables~\ref{tab:blr} and~\ref{tab:fisher-fidelity} quantifies the autocorrelation of the slowest-mixing coordinate of a chain, whereas the MSE quantifies the finite-run error of the full $d$-dimensional posterior-mean estimate, aggregating all coordinates. The two metrics can therefore rank methods differently, as for German under MALA and Caravan under FisherMALA (Section~\ref{sec:experiments}) and for the diagonal estimator under MALA (Appendix~\ref{apx:fisher-fidelity}); we report both.

\subsection{Geometric fidelity: diagonal vs full Fisher state}
\label{apx:fisher-fidelity}

The Fisher state $\widehat S_n$ in Algorithm~\ref{alg:online-fisher-ga-srmc} can be instantiated at two levels of geometric fidelity. Writing $s_{n+1}\!\triangleq\!s(X_{n+1})$, the \emph{diagonal} estimator stores only the per-coordinate squared-score recursion $\widehat r_{n+1}\!=\!(1\!-\!\gamma_{n+1})\widehat r_n+\gamma_{n+1}\,s_{n+1}\!\odot\!s_{n+1}$ at memory and per-step cost $O(d)$; it captures axis-aligned anisotropy but discards cross-coordinate score correlations. The \emph{full} estimator maintains the rank-one recursion $\widehat S_{n+1}\!=\!(1\!-\!\gamma_{n+1})\widehat S_n+\gamma_{n+1}s_{n+1}s_{n+1}^\top$ at $O(d^2)$ memory and per-step cost (plus an amortized $O(d^3/K)$ Cholesky-of-inverse refresh). The same regularized matrix $\widehat S_{n,\delta}\!=\!\widehat S_n+\delta I_d$ feeds the proposal preconditioner $A_n$ and (for GA-SRMC) the repulsion-side metric $M_n$ in both cases. As in Section~\ref{sec:experiments}, the ESS ablation of Table~\ref{tab:fisher-fidelity} initializes the Fisher state at the MAP Hessian (at its diagonal for the diagonal estimator) and freezes it after one update (Remark~\ref{rem:online-scope}); it therefore compares the metrics induced by the MAP Hessian and by its diagonal, not two online estimation trajectories. Detailed implementations are in Appendix \ref{app:online-fisher-details}.

\begin{table}[t]
\caption{Diagonal vs full Fisher estimators across all $8$ targets, MALA and FisherMALA families, and three repellence variants, at the same per-cell $\alpha$ as Table~\ref{tab:blr} (the stability caps were not re-evaluated for the diagonal estimator). Mean worst-coordinate ESS across $40$ replicate chains.}
\label{tab:fisher-fidelity}
\centering
\small
\setlength{\tabcolsep}{2.6pt}
\renewcommand{\arraystretch}{1.05}
\begin{tabular}{lc|rrr|rrr|rrr|rrr}
\toprule
& & \multicolumn{6}{c|}{\textsc{MALA} family} & \multicolumn{6}{c}{\textsc{FisherMALA} family} \\
& & \multicolumn{3}{c|}{diagonal} & \multicolumn{3}{c|}{full} & \multicolumn{3}{c|}{diagonal} & \multicolumn{3}{c}{full} \\
Target & $\kappa$ & base & $+$\,SR & $+$\,GA & base & $+$\,SR & $+$\,GA & base & $+$\,SR & $+$\,GA & base & $+$\,SR & $+$\,GA \\
\midrule
Gauss-2D   & $100$ & 94 & 104 & \textbf{250} & 94 & 104 & \textbf{452} & 9884 & 10717 & \textbf{13220} & 9894 & 12459 & \textbf{18425} \\
Australian & $2.8$ & 2245 & \textbf{4885} & 3077 & 2245 & 4885 & \textbf{5937} & 2912 & 3983 & \textbf{4030} & 4087 & 11170 & \textbf{11638} \\
Heart      & $2.7$ & 2821 & \textbf{6310} & 3870 & 2821 & 6310 & \textbf{7324} & 2963 & 3930 & \textbf{3933} & 4330 & 9451 & \textbf{9549} \\
Pima       & $2.7$ & 3448 & \textbf{7638} & 4921 & 3448 & 7638 & \textbf{10084} & 3313 & 4365 & \textbf{4539} & 5422 & 10975 & \textbf{13358} \\
Ripley     & $1.3$ & 8405 & \textbf{13372} & 9917 & 8405 & 13372 & \textbf{13591} & 9478 & 11266 & \textbf{11305} & 9514 & 15353 & \textbf{15477} \\
German     & $4.6$ & 1437 & \textbf{4663} & 2034 & 1437 & \textbf{4663} & 4173 & 1394 & \textbf{1992} & 1984 & 3259 & 9338 & \textbf{9449} \\
Caravan    & $16$  & 401 & \textbf{922} & 695 & 401 & 922 & \textbf{1353} & 375 & \textbf{632} & 624 & 1694 & \textbf{6381} & 6022 \\
MNIST      & $55$  & 49 & \textbf{74} & 52 & 49 & 74 & \textbf{501} & 67 & \textbf{154} & 143 & 612 & 1389 & \textbf{1399} \\
\bottomrule
\end{tabular}
\end{table}

Table~\ref{tab:fisher-fidelity} reports the worst-coordinate ESS for both estimators on the full $8$-target suite, MALA and FisherMALA families, three repellence variants $\{$base, $+$\,SRMC, $+$\,GA-SRMC$\}$, all at the same per-target $\alpha$ used in Table~\ref{tab:blr}. The diagonal estimator costs $O(d)$ memory and per-step time, while the full estimator costs $O(d^2)$ memory plus an amortized $O(d^3/K)$ Cholesky refresh. By construction, the MALA-family base and $+$\,SRMC cells do not exercise the Fisher state ($A_n\!=\!I_d$ for MALA; $M_n\!=\!I_d$ for SRMC), so the diagonal and full columns coincide there. The informative diagonal-vs-full comparison is therefore the four cells where the Fisher state enters: MALA $+$\,GA-SRMC (through $M_n$) and the entire FisherMALA family (through $A_n$, and additionally through $M_n$ for $+$\,GA-SRMC).

We observe three patterns in ESS. First, on Gauss-2D, whose covariance is exactly diagonal in the standard basis, GA-SRMC keeps its advantage under the diagonal estimator: the diagonal MALA GA-SRMC reaches $250$ versus SRMC's $104$ ($2.4\times$) and the diagonal FisherMALA GA-SRMC reaches $13{,}220$ versus SRMC's $10{,}717$ ($1.23\times$), smaller than, but in the same direction as, the corresponding full-Fisher ratios ($4.3\times$ and $1.48\times$). Second, under diagonal FisherMALA, GA-SRMC remains the within-family winner in five of the eight rows (Gauss-2D, Pima, Australian, and, by margins below $0.5\%$, Ripley and Heart), while German, Caravan, and MNIST flip to SRMC by margins of $0.4$--$7.7\%$. Third, on the BLR posteriors with a MALA proposal and a diagonal $M_n$, SRMC attains a higher worst-coordinate ESS than GA-SRMC: by $1.3$--$1.6\times$ on Australian, Heart, Pima, Ripley, and Caravan, by $2.3\times$ on German, and by $1.4\times$ on MNIST. This third pattern is not specific to GA-SRMC: the broader preconditioning literature has long documented that diagonal preconditioners are a coarse approximation on posteriors whose dominant directions are not coordinate-aligned~\citep{girolami2011riemann,titsias2023optimal}, and \citet[Theorem~3.7]{hird2025Quantifying} formalizes the failure quantitatively as a function of off-diagonal correlation. The mechanism in our setting is the analog on the repulsion side: when $M_n$ is constrained to a coordinate-wise rescaling of $\theta_n$ but the dominant posterior directions are not coordinate-aligned, GA-SRMC's metric-weighted tilt $\theta^\top M_n s(x)$ can amplify the wrong directions. In worst-coordinate ESS alone, SRMC would thus appear to be the safer choice with a diagonal $M_n$ under MALA; the posterior-mean MSE below qualifies this conclusion.

\begin{table}[t]
\caption{Posterior-mean MSE $\frac{1}{d}\|\bar X-\mu_{\rm ref}\|_2^2$ with diagonal Fisher estimates (mean$\pm$std over $40$ runs, in units of $10^{-4}$; lower is better; details in Appendix~\ref{apx:mse}). The MALA and MALA$+$SRMC columns do not use a Fisher estimate and are carried over from Table~\ref{tab:mse}; the four diagonal-Fisher configurations come from a separate set of $40$ runs, so comparisons across the two sets of runs are descriptive. \textbf{Bold}: best within each sampler family; \underline{\textbf{underlined}}: best in the row.}
\label{tab:mse-diag}
\centering
\scriptsize
\setlength{\tabcolsep}{2.5pt}
\begin{tabular}{lrrrrrr}
\toprule
 & \multicolumn{3}{c}{\textsc{MALA} family} & \multicolumn{3}{c}{Diagonal-\textsc{FisherMALA} family} \\
\cmidrule(lr){2-4}\cmidrule(lr){5-7}
Target & base & + SRMC & + diag.\ GA-SRMC & base & + SRMC & + diag.\ GA-SRMC \\
\midrule
Gauss\,2D & 1880.317$\pm$3070.938 & 1827.031$\pm$2775.423 & \textbf{1.234$\pm$1.777} & 10.244$\pm$11.280 & 6.999$\pm$6.965 & \underline{\textbf{0.224$\pm$0.260}} \\
\midrule
Australian & 17.702$\pm$9.018 & 1.273$\pm$0.832 & \underline{\textbf{0.869$\pm$0.384}} & 17.008$\pm$7.469 & \textbf{0.947$\pm$0.484} & 0.957$\pm$0.456 \\
Heart & 18.157$\pm$8.370 & 1.176$\pm$0.527 & \textbf{0.986$\pm$0.422} & 18.692$\pm$9.034 & \underline{\textbf{0.963$\pm$0.452}} & 1.020$\pm$0.518 \\
Pima & 14.418$\pm$10.815 & 0.627$\pm$0.422 & \textbf{0.624$\pm$0.434} & 14.059$\pm$10.421 & \underline{\textbf{0.573$\pm$0.370}} & 0.585$\pm$0.364 \\
Ripley & 5.786$\pm$5.515 & 0.205$\pm$0.188 & \textbf{0.177$\pm$0.154} & 4.447$\pm$3.954 & 0.175$\pm$0.187 & \underline{\textbf{0.173$\pm$0.178}} \\
German & 24.062$\pm$10.052 & 1.624$\pm$0.953 & \underline{\textbf{1.595$\pm$0.765}} & 25.876$\pm$10.092 & \textbf{1.796$\pm$0.963} & 1.933$\pm$1.123 \\
Caravan & 126.083$\pm$22.058 & 45.631$\pm$8.525 & \underline{\textbf{24.693$\pm$5.422}} & 144.471$\pm$32.160 & \textbf{28.896$\pm$8.045} & 30.695$\pm$9.182 \\
MNIST & 685.244$\pm$33.782 & 541.485$\pm$26.944 & \underline{\textbf{356.518$\pm$18.107}} & 896.389$\pm$68.161 & \textbf{519.022$\pm$39.772} & 550.291$\pm$50.981 \\
\bottomrule
\end{tabular}
\end{table}

\paragraph{Posterior-mean MSE with diagonal Fisher estimates.}
Table~\ref{tab:mse-diag} repeats the MSE comparison of Table~\ref{tab:mse} with diagonal Fisher estimates. Four observations follow.
\begin{itemize}[leftmargin=1.6em,itemsep=-0.05em]
\item \textbf{(i) Geometric fidelity depends on the cross-coordinate structure of the Fisher matrix, not on $\kappa$ alone.} Full and diagonal MALA$+$GA-SRMC are similar on Ripley ($0.198$ vs.\ $0.177$) but differ on Caravan ($10.217$ vs.\ $24.693$) and MNIST ($55.500$ vs.\ $356.518$). On Gauss-2D, the MSE changes little despite $\kappa\!=\!100$ (MALA$+$GA-SRMC: $1.236$ vs.\ $1.234$; FisherMALA$+$GA-SRMC: $0.227$ vs.\ $0.224$) because its Fisher matrix is diagonal, although its ESS does change (Table~\ref{tab:fisher-fidelity}).
\item \textbf{(ii) Under MALA, a diagonal metric still lowers the MSE relative to SRMC.} MALA$+$diagonal GA-SRMC has a lower mean MSE than MALA$+$SRMC on all eight targets, with differences within noise on Pima ($0.624$ vs.\ $0.627$) and German ($1.595$ vs.\ $1.624$); at $O(d)$ cost, it reduces the MALA MSE from $14.418$ to $0.624$ on Pima and from $126.083$ to $24.693$ on Caravan. In MSE, the ESS-based preference for SRMC with a diagonal $M_n$ under MALA is therefore not borne out (a descriptive comparison, since the MALA$+$SRMC entries come from the full-Fisher runs).
\item \textbf{(iii) Double penalty.} Diagonal-FisherMALA is worse than MALA on Heart ($18.692$ vs.\ $18.157$), German ($25.876$ vs.\ $24.062$), Caravan ($144.471$ vs.\ $126.083$), and MNIST ($896.389$ vs.\ $685.244$). Reusing this diagonal estimate in GA-SRMC gives a higher mean MSE than SRMC on six of the seven BLR targets (all except Ripley), and this dual-use configuration is also worse than MALA$+$diagonal GA-SRMC on five of the seven BLR targets, including Caravan ($30.695$ vs.\ $24.693$) and MNIST ($550.291$ vs.\ $356.518$). These results are consistent with a double penalty when the same imperfect estimate enters both the proposal and the repulsion.
\item \textbf{(iv) Where a diagonal estimate is best used.} MALA$+$diagonal GA-SRMC outperforms diagonal-FisherMALA$+$SRMC on five of the eight targets (Gauss-2D, Australian, German, Caravan, MNIST), e.g., $356.518$ vs.\ $519.022$ on MNIST. The former keeps the proposal unpreconditioned and uses the marginal scales only in the history feedback; the latter changes the proposal scales without representing rotated directions (i.e., off-diagonal terms), while its SRMC term has no geometry-aware weight matrix $M$.
\end{itemize}

\paragraph{Practical takeaway.}
GA-SRMC's MALA-family ESS advantages in Table~\ref{tab:blr} ($4.3\times$ on Gauss-2D, $6.8\times$ on MNIST, $1.47\times$ on Caravan, and $16$--$32\%$ on Australian, Heart, and Pima) require a full Fisher estimate on coupled posteriors, whereas a diagonal one suffices on axis-aligned targets such as Gauss-2D. When only a diagonal estimate is affordable on a coupled posterior, the two metrics disagree on whether SRMC or MALA$+$diagonal GA-SRMC is preferable (worst-coordinate ESS favors the former, MSE the latter), but reusing the same diagonal estimate in both the proposal and the repulsion is the configuration to avoid: there, SRMC is the safer choice.

\section{Proofs of Section \ref{sec:ga-srmc}}
\label{app:section2-proofs}

\paragraph{Summary of Appendix \ref{app:section2-proofs}.}
This appendix includes the assumptions on the target and on the fixed-$M$ iterates (Appendix~\ref{app:assumptions}), the derivative identity for the score-tilted log-partition function (Appendix~\ref{app:derivative-identity}, Lemma~\ref{lem:deriv-id}), the self-contained proof of the well-posedness and equilibrium claim (Proposition~\ref{prop:fixedM-structure}; Appendix~\ref{app:wellposed-proof}), an explicit derivation of the Jacobian $J_\star^{(M)}$ (Appendix~\ref{app:jacobian-deriv}), and the proof of the almost-sure convergence and CLT for fixed $M$ (Proposition~\ref{prop:clt}; Appendix~\ref{app:clt-proof}), which is also the template for the online-metric proof in Appendix~\ref{app:online-proof}. Throughout, we use $\beta\!\triangleq\!1-\mathbf 1_{\{\rho=1\}}/2\in\{1/2,1\}$.

\subsection{Assumptions}
\label{app:assumptions}

We adopt verbatim the regularity assumptions of \citet[Section~3]{hu2026srmc} on the target $\pi$ and the family of base kernels, specialized to our surrogate family $\{\pi_\theta^{(M)}\}_{\theta\in\R^d,\,M=M^\top\succ 0}$. Throughout, the test function $f:\R^d\!\to\!\R^m$ is $\pi$-square-integrable.

\begin{assumption}[Properties of the target distribution]
\label{ass:target}
Following \citet[Assumption~1]{hu2026srmc}: on $\R^d$, the score function $s(x)=\nabla_x \log \pi(x)= -\nabla_x U(x)$ is $L$-Lipschitz,\footnote{We slightly abuse mathematical notation here and in Appendix \ref{app:clt-proof} by using $U(x) \triangleq - \log \pi(x)$ for the potential. This should never be confused with the long-run covariance $U_{ab}$ defined in \eqref{eq:long-run-cov-U}.} i.e.\ $\|s(x)-s(x')\|\le L\|x-x'\|$ for all $x,x'\in\R^d$, and the potential energy $U$ has super-linear tail growth: there exist constants $c,R\!>\!0$ and $p\!\in\!(1,2]$ such that $U(x)\ge c\|x\|^p$ for all $\|x\|\!\ge\! R$, together with the asymptotic regularity $r^{-(p-2)}\nabla_x^2 U(r\hat x)\to W_\infty(\hat x)$ as $r\to\infty$ uniformly in $\hat x\in\mathbb{S}^{d-1}$, for some measurable positive-definite $W_\infty:\mathbb{S}^{d-1}\!\to\!\R^{d\times d}$.
\end{assumption}

\begin{assumption}[Uniform drift and kernel Lipschitzness]
\label{ass:kernel}
Adopting \citet[Assumption~2]{hu2026srmc} based on \citet{borkar2025ode}, the family $\{K_{\theta,M}\}$ targeting the surrogates $\{\pi_\theta^{(M)}\}$ is geometrically ergodic with the surrogate as its invariant distribution, satisfies the uniform (DV3) drift condition, and is jointly Lipschitz in $(\theta,M)$ on compact sets; full statements are reproduced from \citet[Appendix~C.3]{hu2026srmc}.
\end{assumption}

\begin{assumption}[Online stability]
\label{ass:online-stability}
For the online iterates of Theorem~\ref{thm:online}, with $\theta_0=0$ and $\widehat S_0\succ0$:
(a) almost surely, $\sup_{n\ge0}\lambda_{\max}(\widehat S_n)<\infty$ and $\inf_{n\ge0}\lambda_{\min}(\widehat S_n)>0$, where the bounds may depend on the sample path;
(b) on every compact set of pairs $(\theta,M)$ with $M=M^\top\succ0$, Assumption~\ref{ass:kernel} holds in the (DV3) form of \citet[A2]{borkar2025ode}, including its conditions on the sublevel sets of $W$, with functions and constants that may depend on the compact set, and $\|s(x)\|^2+\|f(x)\|\le c\,W(x)$ for all $x$.
\end{assumption}

The Lipschitz score in Assumption~\ref{ass:target} is the standard regularity used throughout MCMC theory \citep{durmus2017nonasymptotic,chak2023optimal}; without it, the surrogate-score correction $-\alpha\,\nabla_x^2\log\pi(x)\,M\theta$ in the GA-SRMC drift would be unbounded. The super-linear tail growth ($p\!>\!1$) is what guarantees the normalizing constant $Z_\theta^{(M)}\!<\!\infty$ for every $\theta$ and $\alpha$ (Lemma~\ref{lem:wellposed} below): without it, the score-tilted exponent $-\alpha\theta^\top M s(x)$ could grow linearly enough to overpower the polynomial decay of $\pi(x)$ in the tails. Assumption~\ref{ass:kernel} is the controlled-Markov-noise condition needed by the SA CLT~\citep{borkar2025ode}; verified for SR-MH and SR-MALA at $M\!=\!I_d$ in \citet[Appendix~C.3]{hu2026srmc}, the same proof extends verbatim to general fixed $M\!\succ\!0$ once $\theta$ is replaced by $M\theta$ in the surrogate density, since $\theta\!\mapsto\!M\theta$ is a fixed linear isomorphism.

Assumption~\ref{ass:online-stability} is used only for Theorem~\ref{thm:online}. Part~(a) takes the place of the ODE@$\infty$ condition \citet[A3]{borkar2025ode}, which is not available for the augmented iterate (Step~3 of Appendix~\ref{app:online-proof}). It concerns $\widehat S_n$ alone, because $\theta_n$ and $\widehat S_n$ share their gain, so that $\|\theta_n\|^2<\lambda_{\max}(\widehat S_n)$ (Lemma~\ref{lem:shared-gain}). The lower bound is needed because~\eqref{eq:online-fisher-trace} inverts $\widehat S_n$. Part~(a) is also necessary for Theorem~\ref{thm:online}(i), since $\widehat S_n\succ0$ for every $n$ and $\widehat S_n\to S\succ0$ force it. Part~(b) reads Assumption~\ref{ass:kernel} on compact sets of $(\theta,M)$, where the iterates stay under part~(a), and extends the moment envelope to the new noise coordinate $s(x)s(x)^\top$.

\subsection{A derivative identity for the score-tilted surrogate}
\label{app:derivative-identity}

\begin{lemma}[Score-tilted derivative identity]
\label{lem:deriv-id}
In this lemma only, let $h:\R^d\!\to\!\R^k$ be any $\pi_\theta^{(M)}$-integrable function and define
$$
\Phi_h(\theta)\;\triangleq\;\E_{\pi_\theta^{(M)}}[h(X)]
\;=\;\frac{\int h(x)\,\pi(x)\,e^{-\alpha\,\theta^\top M s(x)}\,dx}{Z_\theta^{(M)}}.
$$
Then
\begin{equation}
\label{eq:derivative-identity}
\nabla_\theta \Phi_h(\theta)
\;=\;-\alpha\,\Cov_{\pi_\theta^{(M)}}(h, M s)
\;=\;-\alpha\,\Cov_{\pi_\theta^{(M)}}(h, s)\,M.
\end{equation}
In particular, evaluated at $\theta\!=\!0$ (so that $\pi_0^{(M)}\!=\!\pi$),
$\nabla_\theta \E_{\pi_\theta^{(M)}}[h(X)]\big|_{\theta=0}=-\alpha\,\Cov_\pi(h, s)\,M.$
\end{lemma}

\begin{proof}
Differentiating the ratio under the integral sign in $\theta$ (justified by Assumption~\ref{ass:target}, which makes the integrand and its $\theta$-derivative dominated for $\theta$ in a neighborhood of any fixed point) yields the standard log-partition identity for an exponential family with sufficient statistic $-\alpha M s(X)$:
$$
\nabla_\theta \Phi_h(\theta)
=\E_{\pi_\theta^{(M)}}[h(X)\cdot(-\alpha M s(X))^\top]-\Phi_h(\theta)\cdot\E_{\pi_\theta^{(M)}}[-\alpha M s(X)]^\top
=-\alpha\,\Cov_{\pi_\theta^{(M)}}(h, M s).
$$
The second equality in \eqref{eq:derivative-identity} uses the linearity of covariance: $\Cov(h, M s)=\Cov(h, s)\,M^\top=\Cov(h, s)\,M$ since $M$ is symmetric.
\end{proof}

We invoke Lemma~\ref{lem:deriv-id} twice with $h\!=\!s$ and $h\!=\!f$ at $\theta\!=\!0$, giving the two Jacobian blocks
$$
\nabla_\theta S_M(\theta)\big|_{\theta=0}=-\alpha\,S\,M,
\qquad
\nabla_\theta F_M(\theta)\big|_{\theta=0}=-\alpha\,C\,M,
$$
which combine with the explicit $-\theta$ and $-\mu$ blocks of $h_M$ in \eqref{eq:mean-field-def} to yield the Jacobian $J_\star^{(M)}$ of \eqref{eq:fixedM-jacobian}. Because the identity factors as $\Cov_\pi(h,s)\cdot M$, the matrix $M$ enters the Jacobian only as a right-multiplier on $S$ and $C$, which is the technical reason the optimal-$M$ analysis in Section~\ref{sec:ga-srmc-practical} depends only on $SM$ (rather than on a more elaborate function of $M$).

\subsection{Proof of Proposition~\ref{prop:fixedM-structure} (well-posedness and equilibrium)}
\label{app:wellposed-proof}

We prove the two parts in turn. Lemma~\ref{lem:wellposed} below is part \textbf{(i)}; the equilibrium and global-stability claim is derived afterward.

\begin{lemma}[Well-posedness of $\pi_\theta^{(M)}$]
\label{lem:wellposed}
For every symmetric $M\!\succ\!0$, every $\theta\!\in\!\R^d$, and every $\alpha\!\ge\!0$,
$Z_\theta^{(M)}\!=\!\int_{\R^d}\pi(x)\exp(-\alpha\,\theta^\top M s(x))\,dx<\infty$, so $\pi_\theta^{(M)}$ is a valid probability density.
\end{lemma}

\begin{proof}
This proof largely follows \citet[Appendix C.2]{hu2026srmc}.

\textbf{Step 1: Linear bound on the score-tilted exponent.} Pick any $x_0\!\in\!\R^d$. By the score-Lipschitz hypothesis in Assumption~\ref{ass:target},
$\|s(x)\!-\!s(x_0)\|\!\le\! L\|x\|+L\|x_0\|$, so $\|s(x)\|\!\le\! L\|x\|+c_0$ with $c_0\!\triangleq\! L\|x_0\|+\|s(x_0)\|$. Setting $C_M\!\triangleq\!\alpha\,\|M\|_{\rm op}\,\|\theta\|<\infty$ (with $\|M\|_{\rm op}$ as the operator norm of $M$), Cauchy-Schwarz gives the pointwise upper bound
\begin{equation}
\label{eq:tilt-control}
-\alpha\,\theta^\top M s(x)\;\le\;C_M\,\|s(x)\|\;\le\;C_M\,(L\|x\|+c_0).
\end{equation}

\textbf{Step 2: Integrability of the score-tilted density.} Split $Z_\theta^{(M)}\!=\!\int_{\|x\|<R}+\int_{\|x\|\ge R}$ and use $\pi(x)\!=\!Z_0^{-1}e^{-U(x)}$. The first integral is bounded by $e^{C_M(LR+c_0)}<\infty$ because both $\pi$ and the exponential of a continuous function are bounded on the compact set $\{\|x\|\!\le\! R\}$. For the tail,
$$
\int_{\|x\|\ge R}\pi(x)\,e^{-\alpha\,\theta^\top M s(x)}\,dx
\;\le\;Z_0^{-1}\int_{\|x\|\ge R}e^{-c\|x\|^p+C_M L\|x\|+C_M c_0}\,dx,
$$
which is finite for $R$ large enough that $-c\|x\|^p+C_M L\|x\|\le -\tfrac{c}{2}\|x\|^p$, since $p\!>\!1$ ensures the polynomial decay dominates the linear growth. Hence $Z_\theta^{(M)}\!<\!\infty$.
\end{proof}

The proof of Lemma~\ref{lem:wellposed} depends critically on $p\!>\!1$ in Assumption~\ref{ass:target}: a target with merely linear tail decay (e.g., $U(x)\!=\!c\|x\|$ for $\|x\|$ large) would yield $-c\|x\|+C_M L\|x\|>0$ once $C_M L>c$, i.e., once $\alpha\|M\|_{\rm op}\|\theta\|$ exceeds $c/L$. In that regime, the surrogate $\pi_\theta^{(M)}$ would have an infinite normalizing constant and be undefined. Super-linear growth ensures the surrogate is well-defined for \emph{every} $\theta$ and \emph{every} $\alpha\!\ge\!0$, and is what enables the global stability argument below.

\begin{proof}[Proof of Proposition~\ref{prop:fixedM-structure}]
Lemma~\ref{lem:wellposed} establishes part \textbf{(i)} (well-posedness). We now prove part \textbf{(ii)} (uniqueness and global stability of the equilibrium $\vartheta^\star\!=\!(0,\mu)$) in three steps.

\textbf{Step 1: $\vartheta^\star\!=\!(0,\mu)$ is a fixed point.} At $\theta\!=\!0$, the surrogate reduces to $\pi_0^{(M)}\!=\!\pi$ and Stein's identity gives $S_M(0)\!=\!\E_\pi[s(X)]\!=\!0$ (this is the integration-by-parts identity, which holds under the tail growth in Assumption~\ref{ass:target}). Hence the first block of $h_M$ in \eqref{eq:mean-field-def} vanishes at $\theta\!=\!0$, and the second block vanishes at $\mu_0\!=\!\mu\!\triangleq\!\E_\pi[f(X)]$.

\textbf{Step 2: Local exponential stability at $\vartheta^\star$ via the Hurwitz Jacobian.} The Jacobian $J_\star^{(M)}$ in \eqref{eq:fixedM-jacobian} is derived block-by-block in Lemma~\ref{lem:jacobian-form} (Appendix~\ref{app:jacobian-deriv}); its $(1,1)$ block $-I_d-\alpha SM$ is similar to the symmetric matrix $-I_d-\alpha\,S^{1/2}MS^{1/2}$ via the conjugation $A\!\mapsto\! S^{-1/2}AS^{1/2}$ (direct calculation: $S^{-1/2}(SM)S^{1/2}\!=\!S^{1/2}MS^{1/2}$), and $-I_d-\alpha\,S^{1/2}MS^{1/2}\!\prec\!0$ since $S\!\succ\!0$ and $M\!\succ\!0$ imply $S^{1/2}MS^{1/2}\!\succ\!0$; the $(2,2)$ block is $-I_m\!\prec\!0$. Thus all eigenvalues of $J_\star^{(M)}$ have negative real part (i.e., $J_\star^{(M)}$ is Hurwitz), and $\vartheta^\star$ is locally exponentially stable.

\textbf{Step 3: Global stability via a metric-modified strict Lyapunov function.} Define
\begin{equation}
\label{eq:lyap-V}
V(\theta)\;\triangleq\;\tfrac{1}{2}\,\theta^\top M\,\theta,
\end{equation}
which is strictly convex and coercive (since $M\!\succ\!0$), with $V(0)\!=\!0$ and $V(\theta)\!>\!0$ for $\theta\!\neq\!0$. Its gradient is $\nabla V(\theta)\!=\!M\theta$, so along the deterministic $\theta$-iterate ODE 
\begin{equation}
\label{eq:ode-function}
\dot\theta_t\!=\!h_M^\theta(\theta_t)\!\triangleq\!S_M(\theta_t)-\theta_t,
\end{equation}
the chain rule gives
\begin{equation}
\label{eq:Vdot-decomp}
\dot V(\theta)\;=\;\nabla V(\theta)^\top\dot\theta\;=\;(M\theta)^\top\bigl(S_M(\theta)-\theta\bigr)
\;=\;\theta^\top M\,S_M(\theta)\;-\;\theta^\top M\,\theta.
\end{equation}

\textbf{Step 3a: Gradient and Hessian of the log-partition.} Let $\phi_M(\theta)\!\triangleq\!\log Z_\theta^{(M)}$. Differentiating $\log Z_\theta^{(M)}\!=\!\log\!\int\!\pi(x)e^{-\alpha\theta^\top Ms(x)}dx$ in $\theta$ (justified by Assumption~\ref{ass:target}, identical to the dominated-convergence argument used for Lemma~\ref{lem:deriv-id}) gives
\begin{equation}
\label{eq:grad-phi}
\nabla\phi_M(\theta)\;=\;\frac{\nabla_\theta Z_\theta^{(M)}}{Z_\theta^{(M)}}\;=\;-\alpha M\,\E_{\pi_\theta^{(M)}}[s(X)]\;=\;-\alpha M\,S_M(\theta).
\end{equation}
The Hessian is then obtained by differentiating~\eqref{eq:grad-phi} once more and applying Lemma~\ref{lem:deriv-id} with $h\!=\!s$ at general $\theta$ (so $\nabla_\theta S_M(\theta)\!=\!-\alpha\,\Cov_{\pi_\theta^{(M)}}(s,s)\,M$):
\begin{equation}
\label{eq:hess-phi}
\nabla_\theta^2\,\phi_M(\theta)\;=\;-\alpha M\,\nabla_\theta S_M(\theta)\;=\;\alpha^2\,M\,\Cov_{\pi_\theta^{(M)}}(s,s)\,M\;\succ\;0,
\end{equation}
where the strict positivity uses $M\!\succ\!0$ and $\Cov_{\pi_\theta^{(M)}}(s,s)\!\succ\!0$ (the score is non-degenerate under the smooth surrogate density). Hence $\phi_M$ is strictly convex on $\R^d$.

\textbf{Step 3b: A convexity inequality for the log-partition.} At $\theta\!=\!0$, the surrogate equals $\pi$, so $\phi_M(0)\!=\!\log\!\int\!\pi\!=\!0$ and Stein's identity gives $\nabla\phi_M(0)\!=\!-\alpha M\,\E_\pi[s]\!=\!0$ via~\eqref{eq:grad-phi}. A direct (and equivalent) argument that $\log Z_\theta^{(M)}\!\ge\!0$ comes from Jensen's inequality: for any random variable $Y$, convexity of the exponential gives $\E[e^Y]\!\ge\!e^{\E[Y]}$, i.e., $\log\E[e^Y]\!\ge\!\E[Y]$. Applied with $Y\!=\!-\alpha\theta^\top Ms(X)$ under $\pi$,
\begin{equation}
\label{eq:jensen-logZ}
\log Z_\theta^{(M)}\;=\;\log\E_\pi\!\bigl[e^{-\alpha\theta^\top Ms(X)}\bigr]\;\ge\;\E_\pi\!\bigl[-\alpha\theta^\top Ms(X)\bigr]\;=\;-\alpha\theta^\top M\,\E_\pi[s(X)]\;=\;0,
\end{equation}
i.e., $Z_\theta^{(M)}\!\ge\!1$ for every $\theta$, with equality if an only if $\theta\!=\!0$ (strict by strict convexity). Now apply the gradient inequality (i.e., $f(x) \ge f(y) + \nabla f(y)^\top (x - y)$ for a convex function $f$, see also \citet[Section 3.1.3 (First-order conditions)]{boyd2004convex}) of the strictly convex function $\phi_M$ at the points $0$ and $\theta$:
$$
\phi_M(0)\;\ge\;\phi_M(\theta)+\nabla\phi_M(\theta)^\top(0-\theta)\;=\;\phi_M(\theta)+\alpha\,\theta^\top M\,S_M(\theta),
$$
where the second equality substitutes~\eqref{eq:grad-phi}. Rearranging, and combining with~\eqref{eq:jensen-logZ},
\begin{equation}
\label{eq:convexity-bound}
\theta^\top M\,S_M(\theta)\;\le\;-\alpha^{-1}\,\phi_M(\theta)\;=\;-\alpha^{-1}\log Z_\theta^{(M)}\;\le\;0,
\end{equation}
with both inequalities tight if and only if $\theta\!=\!0$.

\textbf{Step 3c: $V$ is a strict Lyapunov function.} Substituting \eqref{eq:convexity-bound} into \eqref{eq:Vdot-decomp},
\begin{equation}
\label{eq:Vdot-bound}
\dot V(\theta)\;=\; \theta^\top M S_M(\theta)-\theta^\top M\,\theta\;\le\;-\theta^\top M\,\theta\;<\;0\quad\text{for all}\;\theta\!\neq\!0.
\end{equation}

\textbf{Step 3d: Global asymptotic stability via LaSalle's invariance principle.} Since $V$ is coercive, every sublevel set $\{\theta\!:\!V(\theta)\!\le\!c\}$ is compact and forward-invariant under the ODE; in particular, all trajectories are bounded. By Step~3c, $\dot V(\theta)\!<\!0$ on $\R^d\setminus\{0\}$, so the only invariant subset of $\{\theta\!:\!\dot V(\theta)\!=\!0\}$ is $\{0\}$. By LaSalle's invariance principle, $\theta_t\!\to\!0$ as $t\!\to\!\infty$ from any initial condition $\theta_0\!\in\!\R^d$. The $\mu$-iterate ODE $\dot\mu_t\!=\!F_M(\theta_t)-\mu_t$ is a linear contraction driven by $F_M(\theta_t)\!\to\!\mu$, so $\mu_t\!\to\!\mu$ along the same trajectory. Hence $\vartheta^\star\!=\!(0,\mu)$ is the unique globally asymptotically stable equilibrium. This generalizes \citet[Appendix~C.5]{hu2026srmc} from the SRMC case $M\!=\!I_d$ to arbitrary symmetric $M\!\succ\!0$.
\end{proof}

For SRMC ($M\!=\!I_d$), \citet{hu2026srmc} use the augmented potential $F(\theta)\!=\!\alpha^{-1}\log Z_\theta+\tfrac12\|\theta\|^2$ rather than the bare quadratic $\tfrac12\|\theta\|^2$. Both functions \emph{are} Lyapunov functions for the deterministic flow: applying our Step~3 argument with $M\!=\!I_d$ shows that $\tfrac12\|\theta\|^2$ alone is strictly decreasing along trajectories, so it certifies global asymptotic stability without the $\log Z_\theta$ correction. SRMC's choice of $F$ adds the log-partition term not because it is needed for stability, but because it exhibits an additional structural property: the deterministic dynamics is a literal gradient flow $\dot\theta_t\!=\!-\nabla F(\theta_t)$, since $\nabla F\!=\!-S(\theta)+\theta\!=\!-(S(\theta)-\theta)\!=\!-h^\theta$.\footnote{For notation simplicity, we let $h^\theta \equiv h_{I}^\theta$ when $M = I_d$ in \eqref{eq:ode-function}.} This gradient-flow representation gives $\dot F\!=\!-\|\dot\theta\|^2$, which is convenient for finer rate analysis. Both arguments yield the same global-stability conclusion at $M\!=\!I_d$; we work with the simpler $\tfrac12\|\theta\|^2$ because it is the one that generalizes.

For GA-SRMC with general $M\!\succ\!0$ the gradient-flow representation \emph{fails}: with $F_M(\theta)\!=\!\alpha^{-1}\log Z_\theta^{(M)}+\tfrac12\theta^\top M\theta$, the gradient identity~\eqref{eq:grad-phi} gives $\nabla F_M(\theta)\!=\!-MS_M(\theta)+M\theta\!=\!M(\theta-S_M(\theta))$, which equals $-h^\theta(\theta)\!=\!-(S_M(\theta)-\theta)$ only when $M\!=\!I_d$. The bare quadratic $V(\theta)\!=\!\tfrac12\theta^\top M\theta$ in~\eqref{eq:lyap-V} sidesteps this obstruction: even though the dynamics is no longer a Euclidean gradient flow, $V$ is still strictly decreasing along trajectories thanks to the convexity bound~\eqref{eq:convexity-bound}. At $M\!=\!I_d$ this reduces to the bare $\tfrac12\|\theta\|^2$ stand-alone Lyapunov argument; for $M\!\neq\!I_d$ the metric-weighted form is essential. (Equivalently, $V$ is a Lyapunov function for the Riemannian gradient flow under the metric $M$, since $M^{-1}\nabla F_M(\theta)\!=\!\theta-S_M(\theta)\!=\!-h^\theta$.)

\subsection{Derivation of the Jacobian \texorpdfstring{$J_\star^{(M)}$}{J}}
\label{app:jacobian-deriv}

\begin{lemma}[Jacobian of the mean field $h_M$ at $\vartheta^\star$]
\label{lem:jacobian-form}
Under Assumptions~\ref{ass:target}--\ref{ass:kernel}, the Jacobian of the mean-field map $h_M:\R^{d+m}\!\to\!\R^{d+m}$ defined in \eqref{eq:mean-field-def} at the equilibrium $\vartheta^\star\!=\!(0,\mu)$ is
\begin{equation}
\label{eq:jacobian-block}
J_\star^{(M)}\;=\;\begin{bmatrix}
-I_d-\alpha SM & 0\\[2pt]
-\alpha CM & -I_m
\end{bmatrix},
\end{equation}
where $S\!\triangleq\!\Cov_\pi(s,s)$ is the Fisher information matrix of $\pi$, $C\!\triangleq\!\Cov_\pi(f,s)$ is the cross-covariance of the test function with the score, and $0$ denotes the $d\!\times\! m$ zero block.
\end{lemma}

\begin{proof}
Recall $h_M(\theta,\mu)\!=\!(S_M(\theta)-\theta,\,F_M(\theta)-\mu)$ with $S_M(\theta)\!\triangleq\!\E_{\pi_\theta^{(M)}}[s(X)]$ and $F_M(\theta)\!\triangleq\!\E_{\pi_\theta^{(M)}}[f(X)]$. We compute the four partial-derivative blocks separately.

\textbf{Step 1: $\theta$-derivative of the $\theta$-iterate.} By Lemma~\ref{lem:deriv-id} applied at $h\!=\!s$ and $\theta\!=\!0$ (so that $\pi_0^{(M)}\!=\!\pi$ and $\Cov_{\pi_0^{(M)}}\!=\!\Cov_\pi$),
\begin{equation}
\label{eq:partial-S}
\frac{\partial S_M}{\partial \theta}\bigg|_{\theta=0}\;=\;-\alpha\,\Cov_\pi(s,s)\,M\;=\;-\alpha\,SM.
\end{equation}
Since the deterministic $-\theta$ term contributes the constant $-I_d$, we obtain
$\partial(S_M(\theta)-\theta)/\partial\theta|_{\vartheta^\star}\!=\!-I_d-\alpha\,SM$.

\textbf{Step 2: $\theta$-derivative of the $\mu$-iterate.} Analogously, Lemma~\ref{lem:deriv-id} at $h\!=\!f$ and $\theta\!=\!0$ gives
\begin{equation}
\label{eq:partial-F}
\frac{\partial F_M}{\partial \theta}\bigg|_{\theta=0}\;=\;-\alpha\,\Cov_\pi(f,s)\,M\;=\;-\alpha\,CM,
\end{equation}
so $\partial(F_M(\theta)-\mu)/\partial\theta|_{\vartheta^\star}\!=\!-\alpha\,CM$.

\textbf{Step 3: $\mu$-derivative of both blocks.} Neither $S_M(\theta)$ nor $F_M(\theta)$ depends on $\mu$, so
$\partial S_M/\partial\mu\!=\!0$ and $\partial F_M/\partial\mu\!=\!0$. The deterministic $-\theta$ and $-\mu$ terms contribute $0$ and $-I_m$ respectively, giving the second column of \eqref{eq:jacobian-block}.

Stacking the blocks from Steps 1--3 into the $(d\!+\!m)\!\times\!(d\!+\!m)$ Jacobian gives \eqref{eq:jacobian-block}.
\end{proof}

The block-triangular structure of $J_\star^{(M)}$ in \eqref{eq:jacobian-block} (zero upper-right block) is the technical reason the $(\theta,\theta)$-Lyapunov equation $D_M\Sigma_{\theta\theta}^{(M)} + \Sigma_{\theta\theta}^{(M)}D_M^\top = U_{ss}$ decouples from the $\mu$-iterate analysis. This decoupling is what allows the optimal-$M$ argument in Appendix~\ref{app:metric-proof} to be carried out at the level of the $\theta$-iterate alone, rather than requiring a joint optimization over both blocks of $\Sigma_\vartheta^{(M)}$. The same block-triangular structure is exploited by the residual-decomposition argument in Appendix~\ref{app:residual-decomp}, which solves the $\theta$-iterate first and substitutes the result into the $\mu$-iterate.

\subsection{Proof of Proposition~\ref{prop:clt} (a.s.\ convergence and CLT for fixed \texorpdfstring{$M$}{M})}
\label{app:clt-proof}

The proof verifies the five SA conditions of \citet[Assumptions~A1--A5]{borkar2025ode} and quotes \citet[Theorems~1 and~4]{borkar2025ode} to obtain the conclusion. The structure mirrors \citet[Appendix~C.8]{hu2026srmc}, with the additional matrix $M$ inserted at the appropriate places.

\begin{proof}[Proof of Proposition~\ref{prop:clt}]
\textbf{Step 1: Step-size conditions (A1, A5b).} Our schedule $\gamma_n=(n+n_{0})^{-\rho}$ with $\rho\!\in\!(\tfrac12,1]$ satisfies $\sum_n\gamma_n=\infty$ and $\sum_n\gamma_n^2<\infty$, matching the Robbins--Monro conditions \citep[A1, A5b]{borkar2025ode}.

\textbf{Step 2: Lipschitz/linear growth of $H$ (A2, A4).} The mean-field map $H(\vartheta,x)\!=\!(s(x)-\theta,\,f(x)-\mu)$ inherits Lipschitz and linear-growth bounds in $x$ directly from the Lipschitz score in Assumption~\ref{ass:target} and the integrability of $f$. These bounds are uniform in $\vartheta$ and so satisfy \citep[A2, A4]{borkar2025ode}; the $M$-dependence enters only through the surrogate $\pi_\theta^{(M)}$ used by the kernel, not through $H$ itself.

\textbf{Step 3: ODE@$\infty$ stability (A3).} Symmetry of $M$ gives $\theta^\top Ms(x)=(M\theta)^\top s(x)$, so the surrogate satisfies $\pi_\theta^{(M)}\!=\!\pi_{M\theta}^{(I_d)}$ and $S_M(\theta)\!=\!S_{I_d}(M\theta)$. Plugging $\eta\!=\!M\theta_0$ into the SRMC A3 verification of \citet[Appendix~C.8]{hu2026srmc} yields the Hurwitz limit $r^{-1}h_M(r\vartheta_0)\!\to\!h_M^\infty(\vartheta_0)$ for general $M\!\succ\!0$ under Assumption~\ref{ass:target}, and thus \citet[A3]{borkar2025ode} holds.

\textbf{Step 4: Differentiable mean field and Hurwitz Jacobian (A5a).} By Appendix~\ref{app:wellposed-proof}, $\vartheta^\star\!=\!(0,\mu)$ is the unique globally asymptotically stable equilibrium and the Jacobian $\tfrac{\mathbf 1_{\{\rho=1\}}}{2}I+J_\star^{(M)}$ in \eqref{eq:fixedM-jacobian} is Hurwitz. Uniform Lipschitzness of $\theta\!\mapsto\! S_M(\theta)$ on compact sets follows from Lemma~\ref{lem:deriv-id} together with continuity of $\Cov_{\pi_\theta^{(M)}}(s,s)$ in $\theta$, mirroring \citet[Appendix~C.7]{hu2026srmc}.

\textbf{Step 5: Base kernel regularity (DV3 and Lipschitzness).} Assumption~\ref{ass:kernel} provides exactly the uniform drift and kernel Lipschitzness needed by the controlled-Markov-noise version of the SA CLT \citep{borkar2025ode}.

\textbf{Step 6: Conclusion via the SA CLT.} By \citet[Theorems~1 and~4]{borkar2025ode} applied to the joint $(d\!+\!m)$-dimensional iterate $\vartheta_n$, $\vartheta_n\xrightarrow{a.s.}\vartheta^\star$ and
$$
\gamma_n^{-1/2}(\vartheta_n-\vartheta^\star)\;\xrightarrow{dist.}\;\gN(0,\Sigma_\vartheta^{(M)}),
$$
where $\Sigma_\vartheta^{(M)}$ is the unique solution of
$$
\bigl(\tfrac{\mathbf 1_{\{\rho=1\}}}{2}I+J_\star^{(M)}\bigr)\Sigma_\vartheta+\Sigma_\vartheta\bigl(\tfrac{\mathbf 1_{\{\rho=1\}}}{2}I+J_\star^{(M)}\bigr)^\top + U_{HH}=0,
$$
which is exactly~\eqref{eq:lyapunov-eq}.

\textbf{Step 7: Decoupling the $\theta$-iterate.} The block-triangular structure of $J_\star^{(M)}$ in \eqref{eq:fixedM-jacobian} (with zero upper-right block) decouples the $(\theta,\theta)$ block, yielding the Lyapunov equation in \eqref{eq:theta-block-fixedM} with damping operator $D_M\!=\!\beta I_d+\alpha SM$.

\textbf{Step 8: Asymptotic unbiasedness.} Almost-sure convergence $\theta_n\!\to\!0$ implies $\pi_{\theta_n}^{(M)}\!\to\!\pi$ pointwise (continuity of the surrogate density in $\theta$); the $\mu$-iterate then satisfies $\mu_n\!\to\!\mu\!=\!\E_\pi[f(X)]$, recovering exact estimation of the target mean.
\end{proof}
\section{Proofs of Section \ref{sec:ga-srmc-practical}}
\label{app:section3-proofs}

\paragraph{Summary of Appendix \ref{app:section3-proofs}.}
This appendix is organized as follows. Appendix~\ref{app:metric-proof} proves Theorem~\ref{thm:optimal}: the trace-normalized inverse Fisher uniquely minimizes the worst-direction score-history covariance bound in the Loewner order. Appendix~\ref{app:residual-decomp} gives the residual decomposition for the $\mu$-iterate that justifies the second equality in~\eqref{eq:theta-exact-opt}. Appendix~\ref{app:gaussian-proof} derives the explicit limiting covariance for the two special cases referenced in Section~\ref{sec:ga-srmc-practical} of the main body: the independent surrogate sampler and the Gaussian target. Appendix~\ref{app:position-vs-srmc} positions the optimization relative to the original SRMC analysis~\citep{hu2026srmc} and the classical zero-variance/control-variate literature~\citep{assaraf1999zero,mira2013zero,oates2017control}. Appendix~\ref{app:online-proof} proves Theorem~\ref{thm:online} on the online metric under the assumptions of Appendix~\ref{app:assumptions}.

\subsection{Proof of Theorem~\ref{thm:optimal}}
\label{app:metric-proof}

The proof works directly on the damping matrix $D_M\!=\!\beta I_d+\alpha SM$ from \eqref{eq:theta-block-fixedM}. Although $D_M$ is generically non-symmetric (since $SM\!\neq\!MS$ unless $S$ and $M$ commute), it has real positive eigenvalues, and a one-line trace argument suffices to characterize $M_{\rm opt}$:

\begin{lemma}[Smallest-eigenvalue maximizer under a trace constraint]
\label{lem:trace-eig-opt}
For any matrix $A\!\in\!\R^{d\times d}$ whose eigenvalues are real and positive with $\sum_{i=1}^d\lambda_i(A)\!=\!\tau\!>\!0$,
$$
\lambda_{\min}(A)\;\le\;\tau/d,
$$
with equality if and only if all eigenvalues coincide. In particular, if $A$ is also diagonalizable, equality holds if and only if $A\!=\!(\tau/d)I_d$.
\end{lemma}

\begin{proof}
\textbf{(i)} The eigenvalues are positive reals with an average $\tau/d$, and the minimum of a finite tuple is at most its average, with equality if and only if all entries are equal. A diagonalizable $A\!=\!P\Lambda P^{-1}$ with $\Lambda\!=\!(\tau/d)I_d$ equals $(\tau/d)I_d$, and conversely.

\textbf{(ii)} Since $S\!\succ\!0$, its symmetric square root $S^{1/2}$ is invertible, and $SM\!=\!S^{1/2}\,(S^{1/2}MS^{1/2})\,S^{-1/2}$. Thus $SM$ is similar to $B\!\triangleq\!S^{1/2}MS^{1/2}$, which is symmetric and positive definite ($x^\top Bx\!=\!(S^{1/2}x)^\top M(S^{1/2}x)\!>\!0$ for $x\!\neq\!0$). Hence, the eigenvalues of $SM$ are those of $B$: they are real and positive, and they sum to $\tr(B)\!=\!\tr(SM)\!=\!\tau$ by the cyclic invariance of the trace. Part~(i) gives
$$
\lambda_{\min}(SM)\;=\;\lambda_{\min}(B)\;\le\;\frac1d\sum_{i=1}^d\lambda_i(SM)\;=\;\frac{\tau}{d},
$$
with equality if and only if every eigenvalue of $B$ equals $\tau/d$. Because $B$ is symmetric and thus orthogonally diagonalizable, this holds if and only if $B\!=\!(\tau/d)I_d$, i.e., $M\!=\!S^{-1/2}BS^{-1/2}\!=\!(\tau/d)S^{-1}$. Conversely, $M\!=\!(\tau/d)S^{-1}$ is symmetric positive definite, satisfies $\tr(SM)\!=\!\tau$, and gives $SM\!=\!(\tau/d)I_d$, so equality holds; the maximizer of $\lambda_{\min}(SM)$ under the trace constraint is therefore unique.
\end{proof}

\begin{proof}[Proof of Theorem~\ref{thm:optimal}]
The proof has four steps: characterize the spectrum of $D_M$ via a similarity transformation to a symmetric form $L_M$ (Step~1), derive the universal upper bound \eqref{eq:theta-universal-bound} on $\Sigma_{\theta\theta}^{(M)}$ via the operator-norm contraction of $e^{-tL_M}$ (Step~2), maximize $\lambda_{\min}(D_M)$ under the trace constraint via Lemma~\ref{lem:trace-eig-opt} to identify $M_{\rm opt}$ (Step~3), and compute the closed form $\Sigma_{\theta\theta}^{(M_{\rm opt})}\!=\!U_{ss}/(2(\beta+\alpha\tau/d))$ at the optimum (Step~4).

\textbf{Step 1: Spectrum of $D_M$.} Although $D_M$ is non-symmetric, the similarity transformation $D_M\!=\!S^{1/2}L_M S^{-1/2}$ with the symmetric matrix $L_M\!\triangleq\!\beta I_d+\alpha\,S^{1/2}MS^{1/2}\!\succ\!0$ shows that $D_M$ is diagonalizable with real positive eigenvalues. In particular, since $\alpha\!\ge\!0$, $\lambda_{\min}(D_M)\!=\!\lambda_{\min}(L_M)\!=\!\beta+\alpha\lambda_{\min}(S^{1/2}MS^{1/2})\!=\!\beta+\alpha\lambda_{\min}(SM)\!>\!0$, the last identity by Lemma~\ref{lem:trace-eig-opt}(ii). Linearity of trace and cyclic invariance ($\tr(SM)\!=\!\tr(S^{1/2}MS^{1/2})\!=\!\tau$) then give
$$
\tr(D_M)\;=\;\tr(\beta I_d)+\alpha\,\tr(SM)\;=\;d\beta+\alpha\,\tr(SM)\;=\;d\beta+\alpha\tau
$$
under the trace constraint.

\textbf{Step 2: Universal upper bound on $\Sigma_{\theta\theta}^{(M)}$ via the symmetric form.} Starting from the Lyapunov solution $\Sigma_{\theta\theta}^{(M)}\!=\!\int_0^\infty e^{-tD_M}U_{ss}e^{-tD_M^\top}dt$ and substituting the similarity transformation $D_M\!=\!S^{1/2}L_M S^{-1/2}$ from Step~1 (so $e^{-tD_M}\!=\!S^{1/2}e^{-tL_M}S^{-1/2}$ and $e^{-tD_M^\top}\!=\!S^{-1/2}e^{-tL_M}S^{1/2}$ since $L_M$ is symmetric), the integrand factors as $S^{1/2}\bigl(e^{-tL_M}\widetilde U\,e^{-tL_M}\bigr)S^{1/2}$ with $\widetilde U\!\triangleq\!S^{-1/2}U_{ss}S^{-1/2}$. Pulling the $S^{\pm 1/2}$ factors outside the integral yields
$$
\Sigma_{\theta\theta}^{(M)}=S^{1/2}\,\Gamma_M\,S^{1/2},
\qquad
\Gamma_M\!\triangleq\!\int_0^\infty e^{-tL_M}\widetilde U\,e^{-tL_M}\,dt.
$$
We bound $\Gamma_M$ via two ingredients. First, symmetry of $L_M$ gives the operator-norm contraction
$\|e^{-tL_M}\|_2\!=\!e^{-t\lambda_{\min}(L_M)}\!=\!e^{-t\lambda_{\min}(D_M)}$ (the second equality holds because $D_M$ and $L_M$ are similar). Second, $\widetilde U$ is symmetric positive semi-definite, so $\widetilde U\!\preceq\!\lambda_{\max}(\widetilde U)\,I_d$ in the Loewner order. We then chain three standard matrix-analysis facts (each cited at the point of use) to bound the integrand pointwise in $t\!\ge\!0$:
\begin{enumerate}[leftmargin=2em,itemsep=0.05em]
\item[(a)] \emph{Congruence preserves the Loewner order} \citep[Observation 7.1.8]{horn2012matrix}: if $A\!\preceq\!B$ and $X$ are any matrices of compatible size, then $X^\top A X\!\preceq\!X^\top B X$. Applying this with $A\!=\!\widetilde U$, $B\!=\!\lambda_{\max}(\widetilde U)\,I_d$, and $X\!=\!e^{-tL_M}$ (which is symmetric since $L_M$ is symmetric, so $X^\top\!=\!X$) gives
\[
e^{-tL_M}\widetilde U\,e^{-tL_M}\;\preceq\;\lambda_{\max}(\widetilde U)\,e^{-tL_M}\,I_d\,e^{-tL_M}\;=\;\lambda_{\max}(\widetilde U)\,e^{-2tL_M},
\]
where the final equality uses $e^{-tL_M}e^{-tL_M}\!=\!e^{-2tL_M}$ (a single matrix commutes with itself, and the matrix exponential satisfies $e^{A}e^{A}\!=\!e^{2A}$).
\item[(b)] \emph{Spectral lower bound on $L_M$ as a Loewner inequality}: since $L_M$ is symmetric with $\lambda_{\min}(L_M)\!=\!\lambda_{\min}(D_M)$ (Step~1 above), it admits an orthogonal diagonalization $L_M\!=\!Q\Lambda Q^\top$ with $\Lambda\!\succeq\!\lambda_{\min}(D_M)\,I_d$, hence $L_M\!\succeq\!\lambda_{\min}(D_M)\,I_d$ in the Loewner order.
\item[(c)] \emph{Operator antitonicity of $A\!\mapsto\!e^{-2tA}$ on symmetric matrices}: by the Loewner-Heinz theorem and the spectral mapping theorem \citep[Theorem~V.1.9 and Corollary~V.2.6]{bhatia2013matrix}, the matrix function $A\!\mapsto\!e^{-2tA}$ reverses the Loewner order on symmetric matrices for every $t\!\ge\!0$, i.e., $A\!\succeq\!B\!\Rightarrow\!e^{-2tA}\!\preceq\!e^{-2tB}$. Combined with (b), this gives
\[
e^{-2tL_M}\;\preceq\;e^{-2t\lambda_{\min}(D_M)\,I_d}\;=\;e^{-2t\lambda_{\min}(D_M)}\,I_d.
\]
Equivalently, this can be verified directly by orthogonally diagonalizing $L_M\!=\!Q\Lambda Q^\top$ from (b), which gives $e^{-2tL_M}\!=\!Q\,\diag(e^{-2t\lambda_i(L_M)})\,Q^\top$ with each diagonal entry bounded above by $e^{-2t\lambda_{\min}(D_M)}$.
\end{enumerate}
Combining (a) and (c) yields the pointwise bound $e^{-tL_M}\widetilde U\,e^{-tL_M}\!\preceq\!\lambda_{\max}(\widetilde U)\,e^{-2t\lambda_{\min}(D_M)}\,I_d$. Integrating in $t$,
$$
\Gamma_M\;\preceq\;\lambda_{\max}(\widetilde U)\int_0^\infty\! e^{-2t\lambda_{\min}(D_M)}\,dt\;\cdot\;I_d\;=\;\frac{\lambda_{\max}(\widetilde U)}{2\lambda_{\min}(D_M)}\,I_d.
$$
Conjugating by $S^{1/2}$ and using $\lambda_{\max}(\widetilde U)\!=\!\lambda_{\max}(S^{-1/2}U_{ss}S^{-1/2})\!=\!\lambda_{\max}(U_{ss}S^{-1})$ (similarity transform shares the spectrum), we recover the universal bound
$$
\Sigma_{\theta\theta}^{(M)}\;=\;S^{1/2}\Gamma_M S^{1/2}\;\preceq\;\frac{\lambda_{\max}(U_{ss}S^{-1})}{2\lambda_{\min}(D_M)}\,S^{1/2}I_d S^{1/2}\;=\;\frac{\lambda_{\max}(U_{ss}S^{-1})}{2\lambda_{\min}(D_M)}\,S,
$$
which is \eqref{eq:theta-universal-bound} rewritten with $\lambda_{\min}(D_M)$ in the denominator.

\textbf{Step 3: Maximizing $\lambda_{\min}(D_M)$ under the trace constraint.} The bound~\eqref{eq:theta-universal-bound} has the form $c(M)\,S$ with the scalar $c(M)\!\triangleq\!\lambda_{\max}(U_{ss}S^{-1})/(2\lambda_{\min}(D_M))$. Since $S\!\succ\!0$, $c(M)S\!\preceq\!c(M')S$ if and only if $c(M)\!\le\!c(M')$, so minimizing the bound in the Loewner order is equivalent to minimizing $c(M)$. This comparison across $M$ is made at a common equilibrium kernel: $K_{0,M}$, and hence $U_{ss}$ is held fixed as $M$ varies. This is automatic when the proposal does not depend on $M$ (then $K_{0,M}\!=\!K_\pi$); for a proposal tied to $M$, such as FisherMALA with $A\!=\!dM/\tr(M)$, it means freezing the proposal at a common choice, since otherwise $U_{ss}\!=\!U_{ss}(M)$ and the numerator of $c(M)$ vary with $M$. The optimality statement of Theorem~\ref{thm:optimal} is made in this fixed-kernel sense, as is its use at the learned metric in Theorem~\ref{thm:online}. Under it, the numerator $\lambda_{\max}(U_{ss}S^{-1})\!=\!\lambda_{\max}(\widetilde U)\!\ge\!0$ does not depend on $M$. It vanishes only if $U_{ss}\!=\!0$, in which case $\Sigma_{\theta\theta}^{(M)}\!=\!0$ for every $M$ by \eqref{eq:theta-block-fixedM}, every admissible $M$ attains the (zero) bound, and uniqueness fails for $d\!\ge\!2$; we therefore assume $U_{ss}\!\neq\!0$. For $U_{ss}\!\neq\!0$, minimizing the bound is equivalent to maximizing $\lambda_{\min}(D_M)\!=\!\beta+\alpha\lambda_{\min}(SM)$ (Step~1) over $\{M\!=\!M^\top\!\succ\!0:\tr(SM)\!=\!\tau\}$. By Lemma~\ref{lem:trace-eig-opt}(ii),
$$
\lambda_{\min}(D_M)\;=\;\beta+\alpha\lambda_{\min}(SM)\;\le\;\beta+\alpha\tau/d .
$$
If $\alpha\!>\!0$, equality holds if and only if $\lambda_{\min}(SM)\!=\!\tau/d$, i.e., if and only if $M\!=\!(\tau/d)S^{-1}\!=\!M_{\rm opt}$ (Lemma~\ref{lem:trace-eig-opt}(ii)); equivalently, all eigenvalues of $D_M$ coincide and $D_M\!=\!(\beta+\alpha\tau/d)I_d$. Thus, for $\alpha\!>\!0$ and $U_{ss}\!\neq\!0$, at a common equilibrium kernel, $M_{\rm opt}$ uniquely minimizes the universal upper bound \eqref{eq:theta-universal-bound} in the Loewner order, proving \eqref{eq:m-opt}. The condition $\alpha\!>\!0$ is needed for uniqueness: if $\alpha\!=\!0$, then $D_M\!=\!\beta I_d$ and the bound is the same for every $M$.

\textbf{Step 4: Closed form for $\Sigma_{\theta\theta}^{(M_{\rm opt})}$.} At $M\!=\!M_{\rm opt}$, the equality condition of Step~3 gives $D_{M_{\rm opt}}\!=\!(\beta+\alpha\tau/d)I_d$ directly, so the Lyapunov equation collapses to the scalar identity $2(\beta+\alpha\tau/d)\Sigma_{\theta\theta}^{(M_{\rm opt})}\!=\!U_{ss}$, yielding
$$
\Sigma_{\theta\theta}^{(M_{\rm opt})}\;=\;\frac{U_{ss}}{2(\beta+\alpha\tau/d)}.
$$
The second identity in \eqref{eq:theta-exact-opt} (the $\mu$-iterate) is established below in Appendix~\ref{app:residual-decomp}.
\end{proof}

The argument in Step~3 reduces the optimal-$M$ problem to a single criterion: among matrices $D_M$ with a fixed trace $d\beta+\alpha\tau$ and a real positive spectrum, the smallest eigenvalue is maximized at the scalar matrix $D_M\!=\!(\beta+\alpha\tau/d)I_d$. Geometrically, this is \emph{isotropic damping}: every direction in $\R^d$ is damped at the same rate $\beta+\alpha\tau/d$ in the Lyapunov equation (equivalently, the linearized mean-field ODE contracts every $\theta$-direction at the same rate $1+\alpha\tau/d$), so no direction is left as a slow-decaying mode that would dominate the worst-direction CLT covariance bound. Because $D_M\!=\!\beta I_d+\alpha SM$ depends affinely on $M$, for $\alpha\!>\!0$ the unique $M$ satisfying~\eqref{eq:trace-constraint} that makes $D_M$ scalar is $M_{\rm opt}\!=\!(\tau/d)S^{-1}$. The symmetric form $L_M$ enters Step~2 only as a technical device to obtain the exact decay $\|S^{-1/2}e^{-tD_M}S^{1/2}\|_2\!=\!\|e^{-tL_M}\|_2\!=\!e^{-t\lambda_{\min}(D_M)}$ in the $S$-weighted norm. For the non-normal $D_M$, the Euclidean norm $\|e^{-tD_M}\|_2$ itself can exceed $e^{-t\lambda_{\min}(D_M)}$ and is in general only bounded by $\sqrt{\kappa(S)}\,e^{-t\lambda_{\min}(D_M)}$;\footnote{The condition number is defined as $\kappa(S) \triangleq \frac{\lambda_{\max}(S)}{\lambda_{\min}(S)} \ge 1$.} the congruence argument of Step~2 avoids this factor. The optimization itself is intrinsic to $D_M$.

In addition, the optimization in Step~3 minimizes only the \emph{prefactor} of the universal upper bound \eqref{eq:theta-universal-bound}, which is itself a conservative Loewner relaxation: replacing $e^{-2tL_M}$ with $e^{-2t\lambda_{\min}(D_M)}I_d$ collapses the full directional decay of the matrix exponential into its slowest mode, and replacing $\widetilde U$ with $\lambda_{\max}(\widetilde U)I_d$ similarly drops the directional structure of the noise covariance. Maximizing $\lambda_{\min}(D_M)$ therefore tightens only the slowest-mode rate; it does not control the directional dependence of the actual $\Sigma_{\theta\theta}^{(M)}$ in the original $\theta$-basis. Consequently, $\Sigma_{\theta\theta}^{(M_{\rm opt})}$ may be larger than $\Sigma_{\theta\theta}^{(I_d)}$ along directions where the latter happens to be small, and even its largest eigenvalue can be larger: for $U_{ss}\!=\!S\!=\!\diag(1,100)$ (independent sampling, Appendix~\ref{app:gaussian-proof}), $\alpha\!=\!\beta\!=\!1$ and $\tau\!=\!\tr S$, $\lambda_{\max}(\Sigma_{\theta\theta}^{(M_{\rm opt})})\!=\!100/103\!\approx\!0.97$ while $\lambda_{\max}(\Sigma_{\theta\theta}^{(I_d)})\!=\!100/202\!\approx\!0.50$, although the bound~\eqref{eq:theta-universal-bound} equals $S/103$ at $M_{\rm opt}$ versus $S/4$ at $I_d$. Since~\eqref{eq:theta-universal-bound} is equivalent to $\lambda_{\max}(S^{-1/2}\Sigma_{\theta\theta}^{(M)}S^{-1/2})\!\le\!c(M)$, the bound controls the $S$-normalized worst direction; even that quantity is not minimized by $M_{\rm opt}$ in general (for $U_{ss}\!=\!\diag(0.01,100)$ and the same $S,\alpha,\beta,\tau$, it equals $1/103\!\approx\!0.0097$ at $M_{\rm opt}$ and $\approx\!0.0050$ at $I_d$). Theorem~\ref{thm:optimal} is therefore a statement about the guaranteed worst-direction contraction under the trace normalization, not about Loewner dominance, or even worst-direction dominance, of the covariance itself.

\subsection{Residual decomposition for the \texorpdfstring{$\mu$}{mu}-block}
\label{app:residual-decomp}

We isolate the part of the $\mu$-iterate CLT covariance $\Sigma_{\mu\mu}^{(M_{\rm opt})}$ that the score-history mechanism can shrink (the $\alpha$-dependent terms, which decay at a rate $O(\alpha^{-1})$) from its large-repellence limit, the residual floor $U_{gg}/(2\beta)$.

\textbf{Step 1: Score-aligned residual decomposition of the test function.} Decompose the test function $f:\R^d\!\to\!\R^m$ as
$$
f(x)\;=\;g(x)+CS^{-1}s(x),\qquad g(x)\;\triangleq\;f(x)-CS^{-1}s(x),
$$
where $C\!=\!\Cov_\pi(f,s)$ and $S\!=\!\Cov_\pi(s,s)$ as in \eqref{eq:fixedM-jacobian}. The residual $g$ satisfies
$$
\Cov_\pi(g,s)\;=\;\Cov_\pi(f,s)-CS^{-1}\Cov_\pi(s,s)\;=\;C-CS^{-1}S\;=\;0,
$$
i.e., $g$ is the score-orthogonal component of $f$. Since $\E_\pi[s(X)]\!=\!0$ from Stein's identity, $\E_\pi[g(X)]\!=\!\E_\pi[f(X)]\!=\!\mu$.

\textbf{Step 2: Block-Lyapunov system at $M\!=\!M_{\rm opt}$.} The joint Lyapunov equation \eqref{eq:lyapunov-eq} yields the following equations:
\begin{align}
\label{eq:thetatheta-block}
D_M\,\Sigma_{\theta\theta}^{(M)}+\Sigma_{\theta\theta}^{(M)} D_M^\top &= U_{ss},\\
\label{eq:thetamu-block}
D_M\,\Sigma_{\theta\mu}^{(M)}+\beta\,\Sigma_{\theta\mu}^{(M)}+\alpha\,\Sigma_{\theta\theta}^{(M)}(CM)^\top &= U_{sf},\\
\label{eq:mumu-block}
2\beta\,\Sigma_{\mu\mu}^{(M)}+\alpha\bigl[CM\,\Sigma_{\theta\mu}^{(M)}+\Sigma_{\mu\theta}^{(M)}(CM)^\top\bigr]&=U_{ff}.
\end{align}
Setting $M\!=\!M_{\rm opt}\!=\!(\tau/d)S^{-1}$ scalarizes the operators: $D_M\!=\!(\beta+\alpha\tau/d)I_d$, $CM_{\rm opt}\!=\!(\tau/d)CS^{-1}$, and $(CM_{\rm opt})^\top\!=\!(\tau/d)S^{-1}C^\top$.

\textbf{Step 3: Solving for $\Sigma_{\theta\mu}^{(M_{\rm opt})}$.} Write $\eta\!\triangleq\!\alpha\tau/d$ for brevity throughout, so that $D_M\!=\!(\beta+\eta)I_d$ and $\alpha\,CM_{\rm opt}\!=\!\eta\,CS^{-1}$ at $M\!=\!M_{\rm opt}$. Equation~\eqref{eq:thetatheta-block} then yields $\Sigma_{\theta\theta}^{(M_{\rm opt})}\!=\!U_{ss}/(2(\beta+\eta))$, as in Appendix~\ref{app:metric-proof}. Substituting into \eqref{eq:thetamu-block} gives the linear equation $(2\beta+\eta)\Sigma_{\theta\mu}^{(M_{\rm opt})}\!=\!U_{sf}-\alpha\Sigma_{\theta\theta}^{(M_{\rm opt})}(CM_{\rm opt})^\top$, hence the closed form
$$
\Sigma_{\theta\mu}^{(M_{\rm opt})}
\;=\;\frac{U_{sf}}{2\beta+\eta}\;-\;\frac{\eta\,U_{ss}\,S^{-1}C^\top}{2(\beta+\eta)(2\beta+\eta)}.
$$

\textbf{Step 4: Substitution into the $\mu$-iterate.} Substituting the closed form for $\Sigma_{\theta\mu}^{(M_{\rm opt})}$ from Step~3 into the $\mu$-Lyapunov equation \eqref{eq:mumu-block}, the bracketed term in \eqref{eq:mumu-block} expands to
\begin{equation}
\label{eq:mu-substituted}
2\beta\,\Sigma_{\mu\mu}^{(M_{\rm opt})}
\;=\;U_{ff}-\frac{\eta}{2\beta+\eta}\bigl(CS^{-1}U_{sf}+U_{fs}S^{-1}C^\top\bigr)+\frac{\eta^2}{(\beta+\eta)(2\beta+\eta)}\,CS^{-1}U_{ss}S^{-1}C^\top.
\end{equation}
Sending $\eta\!\to\!\infty$ in \eqref{eq:mu-substituted} gives the leading-order limit
\begin{equation}
\label{eq:mu-asymptotic}
\lim_{\eta\to\infty}2\beta\,\Sigma_{\mu\mu}^{(M_{\rm opt})}
\;=\;U_{ff}-\bigl(CS^{-1}U_{sf}+U_{fs}S^{-1}C^\top\bigr)+CS^{-1}U_{ss}S^{-1}C^\top.
\end{equation}

\textbf{Step 5: Identification of the residual floor.} The leading-order limit \eqref{eq:mu-asymptotic} equals $U_{gg}$. Substituting~\eqref{eq:residual-noise-cov} below into~\eqref{eq:mu-asymptotic} produces $U_{gg}$ plus eight matrix products that fall into two groups: four involving the cross-covariances $U_{sg},U_{gs}$ that cancel pairwise and four involving $U_{ss}$ that telescope to zero, leaving $U_{gg}$ as the only surviving term. Throughout this step, we use $U_{sg}, U_{gs}$ for the long-run cross-covariances of $s$ and $g$ under the equilibrium kernel, defined by extending~\eqref{eq:long-run-cov-U} of the main body to the pair $(s,g)$; these are well-defined because $g\!=\!f-CS^{-1}s$ inherits the polynomial growth of $f$. Decompose the long-run noise covariances under the residual identity $f\!=\!g+CS^{-1}s$ (Step~1):
\begin{equation}
\label{eq:residual-noise-cov}
\begin{aligned}
U_{ff}\;&=\;U_{gg}+CS^{-1}U_{sg}+U_{gs}S^{-1}C^\top+CS^{-1}U_{ss}S^{-1}C^\top,\\
U_{sf}\;&=\;U_{sg}+U_{ss}S^{-1}C^\top,
\qquad U_{fs}\;=\;U_{gs}+CS^{-1}U_{ss}.
\end{aligned}
\end{equation}
Substituting \eqref{eq:residual-noise-cov} into the leading-order limit \eqref{eq:mu-asymptotic} expands $U_{ff}-(CS^{-1}U_{sf}+U_{fs}S^{-1}C^\top)+CS^{-1}U_{ss}S^{-1}C^\top$ into $U_{gg}$ plus eight matrix products in $CS^{-1}, U_{sg}, U_{gs}, U_{ss}, S^{-1}C^\top$. Of these, the four involving $U_{sg}$ or $U_{gs}$ cancel pairwise (each $+CS^{-1}U_{sg}$ from $U_{ff}$ matches a $-CS^{-1}U_{sg}$ from $-CS^{-1}U_{sf}$, and analogously for $U_{gs}$), and the remaining four involving $U_{ss}$ telescope to zero ($+CS^{-1}U_{ss}S^{-1}C^\top$ from $U_{ff}$ minus $CS^{-1}U_{ss}S^{-1}C^\top$ from $-CS^{-1}U_{sf}$ minus $CS^{-1}U_{ss}S^{-1}C^\top$ from $-U_{fs}S^{-1}C^\top$ plus $CS^{-1}U_{ss}S^{-1}C^\top$ from the third term, summing to zero). What remains is $U_{gg}$:
$$
\lim_{\eta\to\infty}2\beta\,\Sigma_{\mu\mu}^{(M_{\rm opt})}\;=\;U_{gg}.
$$
For finite $\eta$, substituting \eqref{eq:residual-noise-cov} into \eqref{eq:mu-substituted} instead and collecting terms with $1-\frac{\eta}{2\beta+\eta}\!=\!\frac{2\beta}{2\beta+\eta}$ and $1-\frac{2\eta}{2\beta+\eta}+\frac{\eta^2}{(\beta+\eta)(2\beta+\eta)}\!=\!\frac{\beta}{\beta+\eta}$ gives the exact identity
\begin{equation}
\label{eq:mu-exact}
\Sigma_{\mu\mu}^{(M_{\rm opt})}\;=\;\frac{U_{gg}}{2\beta}+\frac{CS^{-1}U_{sg}+U_{gs}S^{-1}C^\top}{2\beta+\eta}+\frac{CS^{-1}U_{ss}S^{-1}C^\top}{2(\beta+\eta)},
\end{equation}
and hence
\begin{equation}
\label{eq:mu-final}
2\beta\,\Sigma_{\mu\mu}^{(M_{\rm opt})}\;=\;U_{gg}+O(\eta^{-1}),
\end{equation}
i.e., $U_{gg}/(2\beta)$ is the large-repellence limit of $\Sigma_{\mu\mu}^{(M_{\rm opt})}$ and the remaining $\alpha$-dependent part is $O(\alpha^{-1})$ as $\alpha\!\to\!\infty$. Dividing \eqref{eq:mu-final} by $2\beta$ produces the second identity in~\eqref{eq:theta-exact-opt}. The last term of~\eqref{eq:mu-exact} is positive semidefinite and Loewner nonincreasing in $\alpha$, but the middle (cross) term need not be positive semidefinite: $\Cov_\pi(g,s)\!=\!0$ does not imply $U_{gs}\!=\!0$ because the long-run covariance also collects lagged cross-covariances along the chain. Hence $U_{gg}/(2\beta)$ is a lower bound for $\Sigma_{\mu\mu}^{(M_{\rm opt})}$, and $\Sigma_{\mu\mu}^{(M_{\rm opt})}$ is Loewner nonincreasing in $\alpha$ whenever $U_{gs}\!=\!0$, e.g., under independent sampling (Appendix~\ref{app:gaussian-proof}); for a general kernel, we claim only the limit and the $O(\alpha^{-1})$ rate. This is the GA-SRMC analog of the SRMC residual term in \citet[Section~3, Eq.~(15), and Appendix~C.10]{hu2026srmc}.

The leading-order term $U_{gg}/(2\beta)$ in \eqref{eq:mu-final} does not contain $\alpha$ because the score-history mechanism only acts on the score-aligned component $CS^{-1}s$ of the test function, not on the residual $g$. Algebraically, this manifests as the cancelation in Step~5: substituting the expressions in \eqref{eq:residual-noise-cov} into \eqref{eq:mu-asymptotic} produces $U_{gg}$ plus eight cross terms; the four involving $U_{sg}$ or $U_{gs}$ cancel pairwise, and the four involving $U_{ss}$ sum to zero, leaving precisely $U_{gg}$. Because $g$ is constructed to have $\Cov_\pi(g,s)\!=\!0$, the surrogate target $\pi_\theta^{(M)}$ at $\theta\!\neq\!0$ does not bias $\E[g(X)]$ at first order, and the residual term $U_{gg}/(2\beta)$ therefore remains however large $\alpha$ becomes (for finite $\alpha$ and $U_{gs}\!\neq\!0$, the cross term in~\eqref{eq:mu-exact} can move $\Sigma_{\mu\mu}^{(M_{\rm opt})}$ in either direction). Reducing this limit would require a different mechanism, e.g., a non-score-based history average, which lies outside the class analyzed here (Section~\ref{sec:ga-srmc} retains $s(x)$ as the inner statistic); see also Appendix~\ref{apx:alt-repulsion}.

The decomposition $f\!=\!g+CS^{-1}s$ is the orthogonal projection of $f$ onto the linear span of the score $s$ (in the inner product $\Cov_\pi(\cdot,\cdot)$), with $CS^{-1}s$ as the score-aligned component and $g$ as the score-orthogonal residual. The repulsion mechanism in GA-SRMC controls $\theta_n$ (a running average of $s$), so it can directly dampen the score-aligned variance of $f$, but leaves the residual $g$ unchanged; this is why $\Sigma_{\mu\mu}^{(M_{\rm opt})}$ shrinks at a rate of $O(\alpha^{-1})$ on its score-aligned component and saturates at the floor $U_{gg}/(2\beta)$ on its score-orthogonal component (exactly so under independent sampling, where $U_{gs}\!=\!0$; in general, up to the cross term in~\eqref{eq:mu-exact}, which also vanishes as $\alpha\!\to\!\infty$).

\subsection{Special cases: independent surrogate sampler and Gaussian target}
\label{app:gaussian-proof}

Theorem~\ref{thm:optimal} is most informative on two special cases that the original SRMC analysis already isolates~\citep{hu2026srmc}: \textbf{(a)} the \emph{independent surrogate sampler}, where the base kernel $K_{\theta,M}$ draws $X_{n+1}$ independently from $\pi_\theta^{(M)}$ at each step, and \textbf{(b)} the \emph{Gaussian target}, where $\pi\!=\!\gN(\mu_\pi,V)$ and the test function are the mean estimator $f(x)\!=\!x$. Each isolates a complementary aspect of the variance-reduction story: \textbf{(a)} cleanly separates the $M$-independent residual floor $U_{gg}/(2\beta)$ from the $M$-dependent term that GA-SRMC shrinks at rate $O(\alpha^{-1})$, while \textbf{(b)} is the boundary case in which the score-aligned subspace contains the entire test function ($g\!\equiv\!\mu_\pi$, $U_{gg}\!=\!0$) and the limiting covariance shrinks to zero with no floor. In both cases, we give the explicit limiting covariances for both GA-SRMC ($M\!=\!M_{\rm opt}$) and SRMC ($M\!=\!I_d$) and compare them side by side; the comparison exposes the algebraic mechanism behind GA-SRMC's improvement along the slow eigendirections of $S$: scalar (uniform) shrinkage at rate $\beta+\alpha\tau/d$ for GA-SRMC versus matrix (direction-dependent) shrinkage at rate $\beta+\alpha\lambda_i(S)$ for SRMC, with a gap along the slowest eigendirection that tends to the mean-to-minimum eigenvalue ratio $\bar\lambda(S)/\lambda_{\min}(S)\!\ge\!1$ as $\alpha\!\to\!\infty$, where $\bar\lambda(S)\!\triangleq\!\tr(S)/d$; along the fast eigendirections ($\lambda_i(S)\!>\!\bar\lambda(S)$), SRMC damps more strongly. In these comparisons, we take $\tau\!=\!\tr(S)$, the value of $\tr(SM)$ at the SRMC choice $M\!=\!I_d$, so that both metrics satisfy~\eqref{eq:trace-constraint} with the same $\tau$.

\subsubsection{Independent surrogate sampling}

When the base kernel draws independent samples from $\pi_\theta^{(M)}$, the long-run covariance of the noise $H(\vartheta^\star,\cdot)$ at equilibrium reduces to the per-sample covariance under $\pi$:
\begin{equation}
\label{eq:indep-noise}
U_{ss}=\Cov_\pi(s,s)=S,
\qquad
U_{sf}=\Cov_\pi(s,f)=C^\top,
\qquad
U_{ff}=\Cov_\pi(f,f).
\end{equation}
We derive the closed forms for $\Sigma_{\theta\theta}^{(M_{\rm opt})}$ and $\Sigma_{\mu\mu}^{(M_{\rm opt})}$ in turn.

\textbf{Step 1: $\theta$ - iterate covariance and compare with SRMC.} Substituting $U_{ss}\!=\!S$ into the closed form from Appendix~\ref{app:metric-proof} (writing $\eta\!\triangleq\!\alpha\tau/d$),
\begin{equation}
\label{eq:theta-indep}
\Sigma_{\theta\theta}^{(M_{\rm opt})}\;=\;\frac{S}{2(\beta+\eta)}\qquad(\text{GA-SRMC}).
\end{equation}
For comparison, under SRMC's $M\!=\!I_d$ the symmetric Lyapunov equation $(\beta I_d+\alpha S)\Sigma_{\theta\theta}^{(I_d)}+\Sigma_{\theta\theta}^{(I_d)}(\beta I_d+\alpha S)\!=\!S$ has the closed form
\begin{equation}
\label{eq:theta-indep-srmc}
\Sigma_{\theta\theta}^{(I_d)}\;=\;S\,\bigl(2(\beta I_d+\alpha S)\bigr)^{-1}\qquad(\text{SRMC}),
\end{equation}
with the inverse computed in the eigenbasis of $S$. The two forms differ structurally in their denominators: GA-SRMC has the \emph{scalar} $2(\beta+\eta)$ that shrinks every direction at the same rate, while SRMC has the \emph{matrix} $2(\beta I_d+\alpha S)$ whose eigenvalues range from $2(\beta+\alpha\lambda_{\min}(S))$ in the slow direction to $2(\beta+\alpha\lambda_{\max}(S))$ in the fast direction. On the worst-case (slow) direction, the prefactor ratio at high repellence is
\[
\frac{\beta+\alpha\tau/d}{\beta+\alpha\lambda_{\min}(S)}\;\xrightarrow[\alpha\to\infty]{}\;\frac{\bar\lambda(S)}{\lambda_{\min}(S)}\;\ge\;1,
\]
the mean-to-minimum eigenvalue ratio of $S$; the same ratio compares the largest eigenvalues of the $S$-normalized covariances $S^{-1/2}\Sigma_{\theta\theta}^{(M)}S^{-1/2}$, the quantity bounded in~\eqref{eq:theta-universal-bound}. In the Euclidean $\theta$-basis, GA-SRMC's $\theta$-variance is smaller along every eigendirection of $S$ with $\lambda_i(S)\!<\!\bar\lambda(S)$ and larger along those with $\lambda_i(S)\!>\!\bar\lambda(S)$, so its largest eigenvalue can be the larger one (Appendix~\ref{app:metric-proof}). For example, if $d\!=\!10$ and the eigenvalues of $S$ are log-spaced between $\lambda_{\min}(S)$ and $100\,\lambda_{\min}(S)$ (so $\kappa(S)\!=\!100$), the ratio is $\approx 25$.

\textbf{Step 2: $\mu$-iterate covariance.} Substituting \eqref{eq:indep-noise} into the substituted Lyapunov equation \eqref{eq:mu-substituted} from Step~4 of Appendix~\ref{app:residual-decomp} gives
\begin{equation}
\label{eq:mu-indep-derived}
\begin{aligned}
2\beta\,\Sigma_{\mu\mu}^{(M_{\rm opt})}
&=U_{ff}-\frac{\eta}{2\beta+\eta}\bigl(CS^{-1}C^\top+CS^{-1}C^\top\bigr)+\frac{\eta^2}{(\beta+\eta)(2\beta+\eta)}\,CS^{-1}C^\top\\
&=U_{ff}-CS^{-1}C^\top\cdot\frac{2\eta(\beta+\eta)-\eta^2}{(\beta+\eta)(2\beta+\eta)}\\
&=U_{ff}-CS^{-1}C^\top\cdot\frac{\eta(2\beta+\eta)}{(\beta+\eta)(2\beta+\eta)}\;=\;U_{ff}-\frac{\eta\,CS^{-1}C^\top}{\beta+\eta},
\end{aligned}
\end{equation}
where the second equality uses $CS^{-1}U_{ss}S^{-1}C^\top\!=\!CS^{-1}\cdot S\cdot S^{-1}C^\top\!=\!CS^{-1}C^\top$ at $U_{ss}\!=\!S$.

\textbf{Step 3: Decomposition into residual floor plus $M$-dependent term.} Adding and subtracting $CS^{-1}C^\top/(2\beta)$ on the right-hand side of \eqref{eq:mu-indep-derived} divided by $2\beta$ produces
\begin{equation}
\label{eq:mu-indep}
\Sigma_{\mu\mu}^{(M_{\rm opt})}
\;=\;\underbrace{\frac{U_{ff}-CS^{-1}C^\top}{2\beta}}_{\text{residual floor}\,=\,U_{gg}/(2\beta)}+\underbrace{\frac{CS^{-1}C^\top}{2(\beta+\alpha\tau/d)}}_{M\text{-dependent},\;O(\alpha^{-1})} \qquad(\text{GA-SRMC}).
\end{equation}
The first term is the residual floor $U_{gg}/(2\beta)$ promised by Theorem~\ref{thm:optimal} (using the residual identity $U_{ff}-CS^{-1}C^\top\!=\!\Cov_\pi(g,g)\!=\!U_{gg}$ under independent sampling), and the second term decays at a rate $O(\alpha^{-1})$ as $\alpha\!\to\!\infty$; since $U_{gs}\!=\!\Cov_\pi(g,s)\!=\!0$ here, \eqref{eq:mu-indep} is~\eqref{eq:mu-exact} with a vanishing cross term, so the floor is a genuine lower bound and $\Sigma_{\mu\mu}^{(M_{\rm opt})}$ is Loewner nonincreasing in $\alpha$. The first term does not depend on $M$. For a general $M\!=\!M^\top\!\succ\!0$ satisfying~\eqref{eq:trace-constraint}, the block equations \eqref{eq:thetatheta-block}--\eqref{eq:mumu-block} with $U_{ss}\!=\!S$ and $U_{sf}\!=\!C^\top$ are solved by $\Sigma_{\theta\theta}^{(M)}\!=\!\tfrac12S^{1/2}L_M^{-1}S^{1/2}$ and $\Sigma_{\theta\mu}^{(M)}\!=\!\tfrac12S^{1/2}L_M^{-1}S^{-1/2}C^\top$, with $L_M$ from Step~1 of Appendix~\ref{app:metric-proof}, so that
\[
\Sigma_{\mu\mu}^{(M)}=\frac{U_{gg}}{2\beta}+\frac12\,CS^{-1/2}L_M^{-1}S^{-1/2}C^\top,
\quad
\frac12\,CS^{-1/2}L_M^{-1}S^{-1/2}C^\top\preceq\frac{CS^{-1}C^\top}{2(\beta+\alpha\lambda_{\min}(SM))}.
\]
For a scalar test function ($m\!=\!1$), equality holds when $S^{-1/2}C^\top$ is an eigenvector of $L_M$ for its smallest eigenvalue. Hence, for $\alpha\!>\!0$, by Lemma~\ref{lem:trace-eig-opt}(ii), $M_{\rm opt}$ uniquely minimizes the \emph{worst case} of the $M$-dependent term over scalar test functions with a given nonzero score-aligned variance $\Var_\pi(CS^{-1}s)\!=\!CS^{-1}C^\top$, in the same worst-direction sense as Theorem~\ref{thm:optimal}. It need not minimize this term for a particular $f$: for $S\!=\!I_2$, $f\!=\!s_1$, $\alpha\!=\!\beta\!=\!1$ and $\tau\!=\!2$, the term equals $1/(2(1+a))$ at $M\!=\!\diag(a,2-a)$, which is smaller at $a\!=\!1.9$ than at $M_{\rm opt}\!=\!I_2$ ($a\!=\!1$). The SRMC counterpart at $M\!=\!I_d$ is~\eqref{eq:mu-indep-srmc} below (cf.\ the residual term in \citet[Section~3, Eq.~(15), and Appendix~C.10]{hu2026srmc}).

\textbf{Step 4: SRMC's $\mu$-iterate counterpart and worst-direction comparison.} Carrying the same algebra at $M\!=\!I_d$ (using the identity $\alpha(\beta I_d+\alpha S)^{-1}\!=\!S^{-1}-\beta S^{-1}(\beta I_d+\alpha S)^{-1}$ to extract the residual floor) gives
\begin{equation}
\label{eq:mu-indep-srmc}
\Sigma_{\mu\mu}^{(I_d)}\;=\;\underbrace{\frac{U_{ff}-CS^{-1}C^\top}{2\beta}}_{\text{residual floor}\,=\,U_{gg}/(2\beta)}\;+\;\underbrace{\frac{1}{2}\,CS^{-1}(\beta I_d+\alpha S)^{-1}C^\top}_{M\text{-dependent},\;O(\alpha^{-1})}\qquad(\text{SRMC}).
\end{equation}
Side-by-side with the GA-SRMC closed form~\eqref{eq:mu-indep}, three structural points stand out.
\begin{itemize}
\item \emph{Same residual floor.} Both methods leave the identical $\alpha$-independent floor $U_{gg}/(2\beta)\!=\!(U_{ff}-CS^{-1}C^\top)/(2\beta)$. The score-history mechanism, regardless of $M$, damps only the score-aligned component $CS^{-1}s$ of $f$ and leaves the score-orthogonal residual $g$ untouched.
\item \emph{Different $M$-dependent term.} GA-SRMC's $CS^{-1}C^\top/(2(\beta+\alpha\tau/d))$ has a \emph{scalar} denominator that shrinks every direction at the same rate $\beta+\alpha\tau/d$; SRMC's $\tfrac12\,CS^{-1}(\beta I_d+\alpha S)^{-1}C^\top$ has a \emph{matrix} denominator whose worst-direction shrinkage rate is only $\beta+\alpha\lambda_{\min}(S)$.
\item \emph{Worst-direction gain.} When the score-aligned component of $f$ lies along the slowest eigendirection of $S$ (i.e., $S^{-1/2}C^\top$ is an eigenvector of $S$ for $\lambda_{\min}(S)$), GA-SRMC's $M$-dependent term is smaller than SRMC's by the same factor $(\beta+\alpha\tau/d)/(\beta+\alpha\lambda_{\min}(S))\!\to\!\bar\lambda(S)/\lambda_{\min}(S)$ as in the $\theta$-iterate (Step~1); along the fast eigendirections ($\lambda_i(S)\!>\!\bar\lambda(S)$), SRMC's term is the smaller one. This idealized comparison (independent sampling, the exact $M_{\rm opt}$, and the SA estimate $\mu_n$) is qualitative intuition for why a uniform damping rate protects the slowest score direction; it is not a prediction for the experiments of Section~\ref{sec:experiments}, whose kernels and estimators differ.
\end{itemize}

The independent surrogate sampler is a conceptual idealization that neatly isolates the \emph{variance reduction} arising from GA-SRMC's surrogate target, separating it from the asymptotic-variance effects of the underlying base kernel. At $\alpha\!=\!0$ ($\eta\!=\!0$), \eqref{eq:mu-indep} reduces to $\Sigma_{\mu\mu}^{(M)}\!=\!U_{ff}/(2\beta)$, which coincides with the asymptotic covariance of the empirical mean for independent samples from the original target $\pi$ (i.e., without any GA-SRMC tilt), as expected, because the score-history mechanism is deactivated in this case. At $\alpha\!\to\!\infty$ ($\eta\!\to\!\infty$), the $M$-dependent term vanishes and $\Sigma_{\mu\mu}^{(M_{\rm opt})}\!\to\!U_{gg}/(2\beta)$: the residual floor that no $M$ can shrink because $g$ is score-orthogonal. Both limits are consistent with the general kernel result \eqref{eq:mu-final} of Appendix~\ref{app:residual-decomp}.

\subsubsection{Gaussian target with mean estimation}

For $\pi\!=\!\gN(\mu_\pi,V)$ the score is $s(x)\!=\!-V^{-1}(x-\mu_\pi)$ and the Fisher information is $S\!=\!\Cov_\pi(s,s)\!=\!V^{-1}$, so the optimal metric becomes $M_{\rm opt}\!=\!(\tau/d)V$. Direct calculation gives
\begin{equation}
\label{eq:gaussian-alignment}
M_{\rm opt}\,s(x)\;=\;\frac{\tau}{d}V\bigl[-V^{-1}(x-\mu_\pi)\bigr]\;=\;-\frac{\tau}{d}\,(x-\mu_\pi),
\end{equation}
which is the \emph{whitened} state, free of the curvature scale $V^{-1}$. This is exactly why the surrogate exponent $-\alpha\theta^\top M_{\rm opt} s(x)\!=\!(\alpha\tau/d)\theta^\top(x-\mu_\pi)$ is balanced across coordinates, while the SRMC choice $M\!=\!I_d$ yields $-\alpha\theta^\top s(x)\!=\!\alpha\theta^\top V^{-1}(x-\mu_\pi)$, which overweighs small-variance directions.

\textbf{Step 1: Linear relation between $\theta$- and $\mu$-iterates.} For the test function $f(x)\!=\!x$, the algebraic identity $x-\mu_\pi\!=\!-Vs(x)$ implies that the empirical mean iterate $\mu_n$ and the score-history iterate $\theta_n$ differ only by the fixed linear map $V$, up to an asymptotically negligible initialization term:
$$
\mu_n-\mu_\pi\;=\;-V\,\theta_n+o_{a.s.}(1),
$$
since $-V(s(X_n)-\theta_n)\!=\!(X_n-\mu_\pi)+V\theta_n$ has mean zero under the surrogate at the limit $\theta\!=\!0$.

\begin{remark}[Why the Gaussian case collapses the joint $(\theta,\mu)$ CLT to a single block]
\label{rem:gaussian-collapse}
On a Gaussian target with $f(x)\!=\!x$, the test function $f$ and the score $s$ are linearly related by $x-\mu_\pi\!=\!-Vs(x)$, so the residual $g(x)\!=\!\mu_\pi$ in \eqref{eq:gauss-residual-const} is constant. This is a degenerate case of the score-aligned decomposition: $f$ has \emph{no} score-orthogonal component, so the residual floor $U_{gg}/(2\beta)$ in \eqref{eq:mu-indep} is identically zero and the $\mu$-block CLT reduces to a deterministic linear map of the $\theta$-block CLT. This is what makes the Gaussian case in Section~\ref{sec:experiments} so clean: every term in $\Sigma_x^{(M)}$ is captured by the $\theta$-block analysis, with no asymptotic floor to obscure the $\alpha$-dependence.
\end{remark}

\textbf{Step 2: $\mu$-block CLT for $f(x)\!=\!x$.} Hence $\Sigma_x^{(M)}\!=\!V\,\Sigma_{\theta\theta}^{(M)}\,V$, and substituting the closed form from \eqref{eq:theta-exact-opt} gives
\begin{equation}
\label{eq:gaussian-mean-clt}
\Sigma_x^{(M_{\rm opt})}\;=\;\frac{V\,U_{ss}\,V}{2(\beta+\alpha\tau/d)}.
\end{equation}

\textbf{Step 3: Comparison with SRMC, with and without independent-sampling specialization.}

\emph{General Gaussian target (any base kernel).} Without specializing the base kernel, GA-SRMC's closed form $\Sigma_x^{(M_{\rm opt})}\!=\!V U_{ss}V/(2(\beta+\alpha\tau/d))$ from \eqref{eq:gaussian-mean-clt} compares to the SRMC analog
\begin{equation}
\label{eq:gaussian-mean-clt-srmc}
\Sigma_x^{(I_d)}\;=\;V\,\Sigma_{\theta\theta}^{(I_d)}\,V,\quad\text{where}\quad(\beta I_d+\alpha V^{-1})\Sigma_{\theta\theta}^{(I_d)}+\Sigma_{\theta\theta}^{(I_d)}(\beta I_d+\alpha V^{-1})\!=\!U_{ss}.
\end{equation}
The structural difference mirrors part~(a) Step~1: GA-SRMC's denominator $2(\beta+\alpha\tau/d)$ is a \emph{scalar} that contracts the entire matrix $V U_{ss}V$ uniformly, while SRMC's $\Sigma_{\theta\theta}^{(I_d)}$ inherits a direction-dependent damping matrix $\beta I_d+\alpha V^{-1}$ whose worst-direction eigenvalue (along the broadest axis of $V$, i.e., $\sigma_{\max}^2$) is only $\beta+\alpha/\sigma_{\max}^2$. Equivalently, in the eigenbasis $Q$ of $V$ ($V\!=\!Q\diag(\sigma_i^2)Q^\top$, $\widetilde U_{ss}\!\triangleq\!Q^\top U_{ss}Q$):
\[
\bigl(Q^\top\Sigma_x^{(M_{\rm opt})}Q\bigr)_{ij}\;=\;\frac{\sigma_i^2\sigma_j^2\,(\widetilde U_{ss})_{ij}}{2(\beta+\alpha\tau/d)},\qquad
\bigl(Q^\top\Sigma_x^{(I_d)}Q\bigr)_{ij}\;=\;\frac{\sigma_i^2\sigma_j^2\,(\widetilde U_{ss})_{ij}}{2\beta+\alpha/\sigma_i^2+\alpha/\sigma_j^2}.
\]
So, in the worst-case direction ($i\!=\!j\!=\!\argmax_k\sigma_k$), the GA-SRMC-to-SRMC variance ratio tends to
$
(2\beta+2\alpha/\sigma_{\max}^2)/(2(\beta+\alpha\tau/d))\!\to\!\lambda_{\min}(S)/\bar\lambda(S)\!=\!(\bar\lambda(S)/\lambda_{\min}(S))^{-1}\!\le\!1
$
as $\alpha\!\to\!\infty$ (with $S\!=\!V^{-1}$): \emph{regardless of the base kernel}, GA-SRMC's worst-direction $\Sigma_x$-variance is tighter than SRMC's by exactly the mean-to-minimum eigenvalue ratio of $S$. The independent-sampling case below is one concrete instance of this general phenomenon.

\emph{Independent-sampling specialization.} Under independent surrogate sampling, $U_{ss}\!=\!S\!=\!V^{-1}$ and \eqref{eq:gaussian-mean-clt} simplify to
\begin{equation}
\Sigma_x^{(M_{\rm opt})}\;=\;\frac{V\,V^{-1}\,V}{2(\beta+\alpha\tau/d)}\;=\;\frac{V}{2(\beta+\alpha\tau/d)}\qquad(\text{GA-SRMC}),
\end{equation}
which shrinks every coordinate at a \emph{uniform} rate $\beta+\alpha\tau/d$. At $\rho\!=\!1$ ($\beta\!=\!1/2$) this becomes $V/(1+2\alpha\tau/d)$, recovering the formula stated in Section~\ref{sec:ga-srmc-practical}. Plugging $\widetilde U_{ss}\!=\!\diag(1/\sigma_i^2)$ into the general SRMC form above gives the coordinate-wise closed form
\begin{equation}
\label{eq:gauss-srmc}
\Sigma_x^{(I_d)}\;=\;\operatorname{diag}\!\left(\frac{\sigma_i^2}{2\beta+2\alpha/\sigma_i^2}\right)\qquad(\text{SRMC}),
\end{equation}
whose shrinkage rate $\beta+\alpha/\sigma_i^2$ is direction-dependent: narrow ($\sigma_i$ small) coordinates over-contract while broad ($\sigma_i$ large) coordinates under-contract. The worst-direction prefactor ratio is the same $\bar\lambda(S)/\lambda_{\min}(S)\!=\!(1/d)\sum_j(\sigma_{\max}/\sigma_j)^2\!\approx\!25$ on the $d\!=\!10$ Gaussian as in Appendix C.3.1, confirming the general bound above with an explicit constant.

\textbf{Step 4: Vanishing residual floor.} We compute $g(x)\!=\!f(x)-CS^{-1}s(x)$ for $f(x)\!=\!x$ using three substitutions. First, using $s(x)\!=\!-V^{-1}(x-\mu_\pi)$ and Stein's identity for the centered test function $f(x)-\mu_\pi$,
\begin{equation}
\label{eq:gauss-C}
\begin{aligned}
C\;&=\;\Cov_\pi(x,s)\;=\;\E_\pi\!\bigl[(x-\mu_\pi)\,s(x)^\top\bigr]\\
&=\;-\E_\pi\!\bigl[(x-\mu_\pi)(x-\mu_\pi)^\top\bigr]\,V^{-1}\;=\;-V\,V^{-1}\;=\;-I_d.
\end{aligned}
\end{equation}
Second, with $S\!=\!V^{-1}$, we have $CS^{-1}\!=\!(-I_d)V\!=\!-V$, so
\begin{equation}
\label{eq:gauss-CSinvs}
CS^{-1}s(x)\;=\;(-V)\bigl[-V^{-1}(x-\mu_\pi)\bigr]\;=\;x-\mu_\pi.
\end{equation}
Third, the residual is a constant function
\begin{equation}
\label{eq:gauss-residual-const}
g(x)\;=\;f(x)-CS^{-1}s(x)\;=\;x-(x-\mu_\pi)\;=\;\mu_\pi.
\end{equation}
Hence, $\Cov_\pi(g,g)\!=\!0$, $U_{gg}\!=\!0$, and the residual floor in \eqref{eq:mu-indep} vanish. The entire limiting covariance shrinks at the $O(\alpha^{-1})$ rate \eqref{eq:gaussian-mean-clt}, with no $\alpha$-independent floor.

For the SRMC comparison ($M\!=\!I_d$) under the same Gaussian target, $D_M\!=\!\beta I_d+\alpha V^{-1}$ is diagonal in the eigenbasis of $V$, and the coordinatewise covariance is $\sigma_i^2/(2\beta+2\alpha/\sigma_i^2)$. This expression spreads across orders of magnitude with $\sigma_i$ and is dominated by the smallest variance directions, formalizing the \emph{geometry-blind failure mode} described in Section \ref{sec:intro}. The empirical consequence is consistent with the per-coordinate behavior reported in Section~\ref{sec:experiments}: GA-SRMC under $M_{\rm opt}$ contracts every coordinate uniformly, while SRMC under $M\!=\!I_d$ over-contracts the small-$\sigma_i$ direction and under-contracts the large-$\sigma_i$ direction. Equalizing contraction across coordinates, not coordinatewise dominance, is the optimality criterion induced by the trace constraint.

\subsection{Position relative to SRMC and to classical control-variate ideas}
\label{app:position-vs-srmc}

The optimization in Theorem~\ref{thm:optimal} sits at the intersection of two existing lines of work: the \emph{SRMC line} that introduces the score-history surrogate target~\citep{hu2026srmc} and the \emph{zero-variance/control-variate line} that formalizes the residual decomposition~\citep{assaraf1999zero,mira2013zero,oates2017control}. Each line, taken alone, stops short of identifying an optimal metric on the score space; combining them is what yields the closed forms in~\eqref{eq:theta-exact-opt}, as the five paragraphs below make precise.

\paragraph{Relationship to the original SRMC analysis.} Theorem~\ref{thm:optimal} delivers a strictly stronger conclusion than the original SRMC analysis: the closed form $\Sigma_{\theta\theta}^{(M_{\rm opt})}\!=\!U_{ss}/(2(\beta+\alpha\tau/d))$ implies that the \emph{entire matrix} is nonincreasing in $\alpha$ in the Loewner order (strictly decreasing, with a positive definite difference, if $U_{ss}\!\succ\!0$; for a singular $U_{ss}\!\succeq\!0$, the decrease is strict exactly along the directions $v$ with $v^\top U_{ss}v\!>\!0$), while \citet[Proposition 3.4]{hu2026srmc} can only establish the monotone decrease of its Frobenius norm. Concretely, SRMC corresponds to the special case $M\!=\!I_d$ of the GA-SRMC family analyzed here, and \citet[Theorem 3.3]{hu2026srmc} establishes a fixed-$M$ CLT for $M\!=\!I_d$, whose conclusion our Proposition~\ref{prop:clt} yields at $M\!=\!I_d$ (under the hypotheses of Proposition~\ref{prop:clt}, which differ from those of \citet{hu2026srmc}), with $D_{I_d}\!=\!\beta I_d+\alpha S$. At $M\!=\!I_d$, the operator $\beta I_d+\alpha S$ is non-scalar whenever $S$ is non-isotropic, and a Loewner ordering in $\alpha$ over the entire matrix need not hold: for $S\!=\!\diag(1,4)$, $U_{ss}\!=\!\bigl[\begin{smallmatrix}1&0.9\\0.9&1\end{smallmatrix}\bigr]\!\succ\!0$ and $\beta\!=\!1$, the difference $\Sigma_{\theta\theta}^{(I_d)}|_{\alpha=0}-\Sigma_{\theta\theta}^{(I_d)}|_{\alpha=1}$ is indefinite, so a scalar summary such as the Frobenius norm is used instead. The reason GA-SRMC strengthens this is structural: under $M_{\rm opt}$, the damping operator becomes scalar, $D_{M_{\rm opt}}\!=\!(\beta+\alpha\tau/d)I_d$, so the Lyapunov equation~\eqref{eq:theta-block-fixedM} collapses to a scalar identity in $\Sigma_{\theta\theta}$.

\paragraph{The $\mu$-iterate: closed form vs.\ coupled Sylvester equations.} The same scalarization phenomenon explains why $\Sigma_{\mu\mu}^{(M_{\rm opt})}$ admits an explicit closed form in~\eqref{eq:theta-exact-opt}. In contrast, \citet[Appendix C.9]{hu2026srmc} reduces the joint $(\theta,\mu)$ CLT under $M\!=\!I_d$ to coupled Lyapunov/Sylvester equations for $(\Sigma_{\theta\theta},\Sigma_{\theta\mu},\Sigma_{\mu\mu})$. For any fixed $M$, the system~\eqref{eq:thetatheta-block}--\eqref{eq:mumu-block} is block-triangular and can be solved in sequence: after the Lyapunov equation~\eqref{eq:thetatheta-block} for $\Sigma_{\theta\theta}^{(M)}$, \eqref{eq:thetamu-block} gives
\[
\Sigma_{\theta\mu}^{(M)}\;=\;(D_M+\beta I_d)^{-1}\bigl[U_{sf}-\alpha\,\Sigma_{\theta\theta}^{(M)}(CM)^\top\bigr]
\]
($D_M+\beta I_d$ has eigenvalues $2\beta+\alpha\lambda_i(SM)\!>\!0$), and~\eqref{eq:mumu-block} then gives $\Sigma_{\mu\mu}^{(M)}$ explicitly. For general $M$, $\Sigma_{\theta\theta}^{(M)}$ solves a Lyapunov equation, e.g., $\operatorname{vec}\Sigma_{\theta\theta}^{(M)}=(I_d\otimes D_M+D_M\otimes I_d)^{-1}\operatorname{vec}U_{ss}$ (invertible because every eigenvalue sum of $D_M$ is positive), whose solution couples all directions. At $M_{\rm opt}$, the substitution $D_{M_{\rm opt}}\!=\!(\beta+\alpha\tau/d)I_d$ scalarizes the entire system, so all three blocks become explicit rational functions of $\alpha$ in the long-run covariances, as in~\eqref{eq:mu-exact}; this isolates $U_{gg}/(2\beta)$ as the only $\alpha$-independent term (the large-repellence limit) and identifies the $O(\alpha^{-1})$ rate at which the remaining terms vanish. The two special cases that the original SRMC analysis isolates, i.e., the independent surrogate sampler and the Gaussian target with mean estimation \citep[Section~3]{hu2026srmc}, reappear here as concrete instances under the \emph{optimal} metric (full derivations in Appendix~\ref{app:gaussian-proof}). In the independent-sampler case, $\Sigma_{\mu\mu}^{(M_{\rm opt})}$ separates exactly into a residual floor $U_{gg}/(2\beta)$ plus an $M$-dependent term that contracts at the uniform rate $\beta+\alpha\tau/d$ and whose worst case over scalar test functions with a given score-aligned variance is uniquely minimized by $M_{\rm opt}$ (Step~3 of the independent-sampler case in Appendix~\ref{app:gaussian-proof}); it is smaller than its SRMC counterpart along the slow eigendirections of $S$, not in every direction. In the Gaussian-mean case, $g\!\equiv\!\mu_\pi$ so the residual floor vanishes and the entire covariance shrinks uniformly at rate $O(\alpha^{-1})$; along the broadest axis of $\pi$ and at a common equilibrium kernel, its variance is below SRMC's on the same target by a factor tending to $\bar\lambda(S)/\lambda_{\min}(S)$ as $\alpha\!\to\!\infty$.

\paragraph{The residual floor as a shared phenomenon.} GA-SRMC at $M_{\rm opt}$ and SRMC at $M\!=\!I_d$ share a basic structure: the score-history mechanism damps only the score-aligned component of $f$, at rate $O(\alpha^{-1})$, and as $\alpha\!\to\!\infty$ the covariance tends to the score-orthogonal residual $U_{gg}/(2\beta)$ in~\eqref{eq:theta-exact-opt} for GA-SRMC, the analog of the residual term in \citet[Section~3, Eq.~(15), and Appendix~C.10]{hu2026srmc} for SRMC; this limit is a lower bound when the cross term in~\eqref{eq:mu-exact} vanishes, e.g., under independent sampling. Two points, listed below, distinguish the GA-SRMC form. First, \emph{scalarization}: the $M$-dependent component vanishes uniformly in direction at rate $O(\alpha^{-1})$, rather than direction-dependently as under $M\!=\!I_d$. Second, \emph{a common floor}: $U_{gg}$ is the score-orthogonal long-run covariance of the equilibrium kernel, set entirely by the test function and that kernel, so it depends on $M$ only if the kernel does; at a common equilibrium kernel GA-SRMC and SRMC share the same floor (compare~\eqref{eq:mu-indep} with~\eqref{eq:mu-indep-srmc}).

\paragraph{Genealogy of the residual decomposition.} Turning from the \emph{SRMC line} to GA-SRMC's other intellectual ancestor: the score-aligned / score-orthogonal split $f\!=\!CS^{-1}s+g$ used in Step~1 of Appendix~\ref{app:residual-decomp} is the multivariate version of the variance-reducing projection identified in three classical pieces of the Monte Carlo control-variate literature. \citet{assaraf1999zero} introduced score-based control variates as a \emph{zero-variance principle}, augmenting an estimator by terms in the linear span of the score so that the residual variance vanishes under the target. \citet{mira2013zero} extended the idea to general Bayesian estimators, designing zero-variance MCMC estimators by regressing the test function onto the score. \citet{oates2017control} formalized the construction within a reproducing kernel Hilbert space (RKHS) framework as \emph{control functionals}, in which Stein operators are applied to an appropriate function space to generate variance-reducing controls. When these controls are restricted to the \emph{linear span of the score}, the procedure removes precisely the score-aligned component $CS^{-1}s$, leaving the residual term $g$; more expressive Stein-type controls of the form $\phi' + s\phi$ are also permitted and can further reduce the residual contribution. For example, when $\pi=\gN(0,1)$ and $f(x)=x^2$, one has $C=0$ and hence $g=f$, yet choosing $\phi(x)=x$ yields the correction $1-x^2$ and produces a zero-variance estimator. The present analogy, however, pertains exclusively to the linear-score component.
GA-SRMC implements this idea \emph{dynamically} via the surrogate target $\pi_\theta^{(M)}$, rather than through a static, post-hoc adjustment. Specifically, the score-aligned component is attenuated by the running score average $\theta_n$ at rate $O(\alpha^{-1})$, while the residual term induces the same asymptotic floor $U_{gg}/(2\beta)$. This decomposition characterizes the dynamic estimator for every fixed $M$: for a common equilibrium kernel, the limiting value $U_{gg}/(2\beta)$ is independent of $M$ (cf.~\eqref{eq:mu-indep} and~\eqref{eq:mu-indep-srmc}). Theorem~\ref{thm:optimal} further identifies an optimal choice of metric: $M_{\rm opt}$ is the unique trace-constrained metric for which $D_M=(\beta+\alpha\tau/d)I_d$ is scalar, thereby ensuring that the score-aligned component is damped isotropically, i.e., at the same rate in every direction, rather than at direction-dependent rates $\beta+\alpha\lambda_i(SM)$.

\paragraph{Why a closed form was not previously available.} Both lines stop short of an optimal metric on the score space, but for opposite reasons. The \emph{SRMC line}~\citep{hu2026srmc} fixes $M\!=\!I_d$ a priori and obtains scalar (Frobenius) bounds. The \emph{control-variate line}~\citep{assaraf1999zero,mira2013zero,oates2017control} constructs a residual decomposition but uses the Fisher matrix in a static post-hoc estimator rather than as a dynamic damping geometry, leaving the metric question unmotivated. Theorem~\ref{thm:optimal} bridges the two by combining three ingredients: surrogate-target machinery delivers an $O(\alpha^{-1})$ damping on the score-aligned component, the residual decomposition isolates the floor it leaves, and the trace-normalized inverse Fisher is the unique metric that scalarizes the resulting score-history Lyapunov equation. The result is a closed-form joint $(\theta,\mu)$ CLT for every $f\!\in\!L^2(\pi)$, the strongest such statement currently available in either line.

\subsection{Proof of Theorem~\ref{thm:online}}
\label{app:online-proof}

The argument follows Steps~1--8 in the proof of Proposition~\ref{prop:clt} (Appendix~\ref{app:clt-proof}), applied to the augmented iterate $z_n$; accordingly, we only document the points at which the proof requires modification. The two adjustments stated after Theorem~\ref{thm:online}, namely, the use of a moving-metric Lyapunov function and the identity $\pi_0^{(M)}=\pi$ for all $M$, both enter in Step~4. Because the stability of $\widehat S_n$ is postulated via Assumption~\ref{ass:online-stability}, rather than established through an ODE@$\infty$ argument, we instead verify the conditions of \citet{borkar2025ode} for a localized Markov chain that agrees with $z_n$ as long as $z_n$ remains in a given compact set, and subsequently pass to the limit by allowing this compact set to expand.

\paragraph{Augmented state and mean-field notation.}
Extending the fixed-$M$ notation of Section~\ref{sec:ga-srmc}, we write $z=(\theta,\mu,\widehat S)\in\R^d\times\R^m\times\mathbb S^d$ (identifying symmetric $d\times d$ matrices $\mathbb S^d$ with their vectorized representations in $\R^{d(d+1)/2}$) and denote the target equilibrium by $z^\star\triangleq(\vartheta^\star,S)=(0,\mu,S)$. For any $\widehat S\succ 0$, define the trace-normalized inverse-Fisher metric map
\begin{equation}
\label{eq:metric-map-def}
M_{\widehat S}\;\triangleq\;\frac{\tr(\widehat S)}{d}\,\widehat S^{-1},
\end{equation}
so that the online metric in~\eqref{eq:online-fisher-trace} is $M_n=M_{\widehat S_n}$ and its value at $\widehat S=S$ is $M_S=M_{\rm opt}$ from~\eqref{eq:m-opt} with scale $\tau=\tr S$. Matching $S_M(\theta)$ and $F_M(\theta)$ in~\eqref{eq:ZM-SM-FM}, define the surrogate second-moment map
\begin{equation}
\label{eq:QM-def}
Q_M(\theta)\;\triangleq\;\E_{\pi_\theta^{(M)}}\!\bigl[s(X)s(X)^\top\bigr].
\end{equation}
The one-step SA residual and the associated mean-field drift on $\R^d\times\R^m\times\mathbb S^d_{++}$ are then
\begin{equation}
\label{eq:online-H-h}
\begin{split}
H(z,x)&\;\triangleq\;\bigl(s(x)-\theta,\;f(x)-\mu,\;s(x)s(x)^\top-\widehat S\bigr),\\
h(z)&\;\triangleq\;\bigl(S_M(\theta)-\theta,\;F_M(\theta)-\mu,\;Q_M(\theta)-\widehat S\bigr)\Big|_{M=M_{\widehat S}},
\end{split}
\end{equation}
and the online recursion of Algorithm~\ref{alg:online-fisher-ga-srmc} takes the coupled SA form
\begin{equation}
\label{eq:online-coupled-sa}
X_{n+1}\;\sim\;K_{\theta_n,M_n}(X_n,\cdot),\qquad z_{n+1}\;=\;z_n+\gamma_{n+1}\,H(z_n,X_{n+1}).
\end{equation}
Under Assumption~\ref{ass:target}, the super-linear tail bound $U(x)\ge c\|x\|^p$ ($p>1$) implies that $\pi$ has finite exponential moments of all linear orders (Lemma~\ref{lem:wellposed}), while $\|s(x)\|\le L\|x\|+c_0$ grows at most linearly. By the dominated convergence theorem \citep[Theorem~2.27]{folland1999real} applied as in Lemma~\ref{lem:deriv-id}, we may differentiate under the integral sign to arbitrary order for the test functions $s$, $f$, and $ss^\top$, where the $\pi_\theta^{(M)}$-integrability of $f\in L^2(\pi)$ and its derivatives follows from the Cauchy--Schwarz inequality:
\[
\E_\pi\!\bigl[\|f(X)\|\,\|s(X)\|^k\,e^{-\alpha\theta^\top Ms(X)}\bigr]
\;\le\;\|f\|_{L^2(\pi)}\,\Bigl(\E_\pi\!\bigl[\|s(X)\|^{2k}\,e^{-2\alpha\theta^\top Ms(X)}\bigr]\Bigr)^{1/2}\;<\;\infty.
\]
Since $\pi_\theta^{(M)}=\pi_{M\theta}^{(I_d)}$, the maps $S_M(\theta)$, $F_M(\theta)$, and $Q_M(\theta)$ are $C^\infty$ functions of the product $\eta\triangleq M\theta\in\R^d$. Furthermore, $Q_M(\theta)\succ 0$ for every $\theta\in\R^d$ and $M\succ 0$: if $v^\top Q_M(\theta)v=\E_{\pi_\theta^{(M)}}[(v^\top s(X))^2]=0$ for some $v\in\R^d$, then because $\pi_\theta^{(M)}$ has a strictly positive continuous density on $\R^d$ and $x\mapsto v^\top s(x)$ is continuous, we must have $v^\top\nabla U(x)=-v^\top s(x)=0$ for all $x\in\R^d$. By the fundamental theorem of calculus along lines parallel to $v$, $U(x+tv)=U(x)$ for all $t\in\R$, which contradicts the coercive growth $U(x)\ge c\|x\|^p$ as $|t|\to\infty$ unless $v=0$. In particular, $S=Q_M(0)\succ 0$.

We next establish two structural lemmas before introducing the localization.

\begin{lemma}[Shared-gain dominance of $\theta_n$ by $\widehat S_n$]
\label{lem:shared-gain}
Suppose that Algorithm~\ref{alg:online-fisher-ga-srmc} is initialized with $\theta_0=0$ and $\widehat S_0\succ 0$. Then, along every sample path and for every $n\ge 0$, the Schur complement matrix $\widehat S_n-\theta_n\theta_n^\top$ is strictly positive definite: $\widehat S_n-\theta_n\theta_n^\top\succ 0$. Consequently, $\widehat S_n\succ 0$ (so $M_n$ is well-defined), $\theta_n^\top\widehat S_n^{-1}\theta_n<1$, and $\|\theta_n\|^2<\lambda_{\max}(\widehat S_n)$.
\end{lemma}

\begin{proof}
Fix $n\ge 0$ and abbreviate $\gamma\triangleq\gamma_{n+1}\in(0,1)$ and $s_{n+1}\triangleq s(X_{n+1})$. Because $\theta_n$ and $\widehat S_n$ share the step size $\gamma$, their recursions read
\[
\theta_{n+1}\;=\;(1-\gamma)\theta_n+\gamma s_{n+1},
\qquad
\widehat S_{n+1}\;=\;(1-\gamma)\widehat S_n+\gamma s_{n+1}s_{n+1}^\top.
\]
Expanding the outer product $\theta_{n+1}\theta_{n+1}^\top=(1-\gamma)^2\theta_n\theta_n^\top+\gamma(1-\gamma)(\theta_ns_{n+1}^\top+s_{n+1}\theta_n^\top)+\gamma^2s_{n+1}s_{n+1}^\top$ and substituting $(1-\gamma)^2=(1-\gamma)-\gamma(1-\gamma)$ and $\gamma-\gamma^2=\gamma(1-\gamma)$, we obtain the exact one-step decomposition
\begin{align*}
\widehat S_{n+1}-\theta_{n+1}\theta_{n+1}^\top
&=(1-\gamma)\widehat S_n-(1-\gamma)^2\theta_n\theta_n^\top+\gamma(1-\gamma)\bigl(s_{n+1}s_{n+1}^\top-\theta_ns_{n+1}^\top-s_{n+1}\theta_n^\top\bigr)\\
&=(1-\gamma)\bigl(\widehat S_n-\theta_n\theta_n^\top\bigr)+\gamma(1-\gamma)\bigl(\theta_n\theta_n^\top-\theta_ns_{n+1}^\top-s_{n+1}\theta_n^\top+s_{n+1}s_{n+1}^\top\bigr)\\
&=(1-\gamma)\bigl(\widehat S_n-\theta_n\theta_n^\top\bigr)+\gamma(1-\gamma)(s_{n+1}-\theta_n)(s_{n+1}-\theta_n)^\top.
\end{align*}
We now induct on $n$. At $n=0$, $\widehat S_0-\theta_0\theta_0^\top=\widehat S_0\succ 0$ by initialization. If $\widehat S_n-\theta_n\theta_n^\top\succ 0$, then since $\gamma\in(0,1)$, the first term on the right-hand side is strictly positive definite and the second term is rank-one positive semidefinite; hence $\widehat S_{n+1}-\theta_{n+1}\theta_{n+1}^\top\succ 0$, completing the induction.

For the stated consequences, $\widehat S_n\succ\theta_n\theta_n^\top\succeq 0$ immediately gives $\widehat S_n\succ 0$, so $\widehat S_n^{-1}$ and $M_n$ in~\eqref{eq:online-fisher-trace} are well-defined. Congruently conjugating $\widehat S_n-\theta_n\theta_n^\top\succ 0$ by the nonsingular matrix $\widehat S_n^{-1/2}$ preserves strict positive definiteness \citep[Observation~7.1.8]{horn2012matrix}, yielding
\[
I_d-\widehat S_n^{-1/2}\theta_n\theta_n^\top\widehat S_n^{-1/2}\;\succ\;0.
\]
Since $\widehat S_n^{-1/2}\theta_n\theta_n^\top\widehat S_n^{-1/2}$ has rank at most one with largest eigenvalue equal to its trace $\theta_n^\top\widehat S_n^{-1}\theta_n$, it follows that $\theta_n^\top\widehat S_n^{-1}\theta_n<1$. Finally, for $\theta_n\neq 0$, evaluating the quadratic form of $\widehat S_n-\theta_n\theta_n^\top\succ 0$ at the unit vector $v\triangleq\theta_n/\|\theta_n\|$ and applying the Rayleigh--Ritz theorem \citep[Theorem~4.2.2]{horn2012matrix} gives
\[
\|\theta_n\|^2\;=\;v^\top(\theta_n\theta_n^\top)v\;<\;v^\top\widehat S_n v\;\le\;\lambda_{\max}(\widehat S_n),
\]
while for $\theta_n=0$ the inequality $0<\lambda_{\max}(\widehat S_n)$ is immediate from $\widehat S_n\succ 0$.
\end{proof}
Lemma~\ref{lem:shared-gain} explains why Assumption~\ref{ass:online-stability}(a) only needs to postulate boundedness of $\widehat S_n$: the score-history iterate $\theta_n$ is automatically bounded by $\lambda_{\max}(\widehat S_n)^{1/2}$, and the moving-metric Lyapunov function $\theta_n^\top\widehat S_n^{-1}\theta_n$ remains strictly bounded by~$1$ along every trajectory.

\begin{lemma}[Regularity and algebraic properties of the metric map]
\label{lem:metric-map}
The metric map $\widehat S\mapsto M_{\widehat S}$ in~\eqref{eq:metric-map-def} satisfies:
\begin{enumerate}[leftmargin=2.0em,itemsep=0.05em]
\item[\textbf{(i)}] $\widehat S\mapsto M_{\widehat S}$ is $C^\infty$ on the open positive-definite cone $\mathbb S^d_{++}\triangleq\{\widehat S\in\mathbb S^d:\widehat S\succ 0\}$, and hence Lipschitz on every compact subset of $\mathbb S^d_{++}$.
\item[\textbf{(ii)}] At the true Fisher matrix $\widehat S=S\succ 0$, $M_S=M_{\rm opt}$ with scale $\tau=\tr S$.
\item[\textbf{(iii)}] $SM_{\rm opt}=\frac{\tr S}{d}\,I_d$.
\item[\textbf{(iv)}] For every $\widehat S\succ 0$ and $M=M_{\widehat S}$, $dM/\tr(M)=d\,\widehat S^{-1}/\tr(\widehat S^{-1})$, which at $\widehat S=\widehat S_n$ coincides with the unregularized ($\delta=0$) FisherMALA proposal preconditioner $A_n$ of Appendix~\ref{app:online-fisher-details}.
\end{enumerate}
\end{lemma}

\begin{proof}
\textbf{(i)} By Cramer's rule \citep[Section~0.8.2]{horn2012matrix}, each entry of $\widehat S^{-1}$ is a rational function of the entries of $\widehat S$ with nonvanishing denominator $\det(\widehat S)>0$ on $\mathbb S^d_{++}$, and $\widehat S\mapsto\tr(\widehat S)$ is linear; hence $\widehat S\mapsto M_{\widehat S}$ is $C^\infty$ on $\mathbb S^d_{++}$. Using the standard Fr\'{e}chet derivative of the matrix inverse $D(\widehat S^{-1})[E]=-\widehat S^{-1}E\widehat S^{-1}$ \citep[Appendix~A.4.1]{boyd2004convex}, the directional derivative of $M_{\widehat S}$ along $E\in\mathbb S^d$ is
\[
DM_{\widehat S}[E]\;=\;\frac{\tr(E)}{d}\,\widehat S^{-1}-\frac{\tr(\widehat S)}{d}\,\widehat S^{-1}E\widehat S^{-1}.
\]
Because $\mathbb S^d_{++}$ is convex, the convex hull of any compact subset $\mathcal C\subset\mathbb S^d_{++}$ is again a compact subset of $\mathbb S^d_{++}$. By continuity, $\|DM_{\widehat S}\|$ is bounded on $\mathrm{conv}(\mathcal C)$, and the mean value inequality \citep[Theorem~9.19]{rudin1976principles} implies that $\widehat S\mapsto M_{\widehat S}$ is Lipschitz on $\mathcal C$.

\textbf{(ii)--(iii)} Evaluating~\eqref{eq:metric-map-def} at $\widehat S=S\succ 0$ gives $M_S=\frac{\tr S}{d}S^{-1}$, which matches $M_{\rm opt}=(\tau/d)S^{-1}$ in~\eqref{eq:m-opt} with $\tau=\tr S$, and multiplying on the left by $S$ gives $SM_{\rm opt}=\frac{\tr S}{d}SS^{-1}=\frac{\tr S}{d}I_d$.

\textbf{(iv)} Taking the trace of $M=M_{\widehat S}=\frac{\tr(\widehat S)}{d}\widehat S^{-1}$ gives $\tr(M)=\frac{\tr(\widehat S)}{d}\tr(\widehat S^{-1})>0$. Canceling the scalar factor $\tr(\widehat S)/d$ between numerator and denominator yields
\[
\frac{dM}{\tr(M)}\;=\;\frac{d\,\bigl(\tr(\widehat S)/d\bigr)\,\widehat S^{-1}}{\bigl(\tr(\widehat S)/d\bigr)\,\tr(\widehat S^{-1})}
\;=\;\frac{d\,\widehat S^{-1}}{\tr(\widehat S^{-1})},
\]
which at $\widehat S=\widehat S_n$ recovers $A_n=(d/\tr(\widehat S_{n,\delta}^{-1}))\widehat S_{n,\delta}^{-1}$ from Appendix~\ref{app:online-fisher-details} at $\delta=0$.
\end{proof}

\paragraph{Why localization on $(\theta,\widehat S)$ is needed, and construction of the localized chain.}
Unlike the fixed-$M$ recursion of Proposition~\ref{prop:clt}, the unlocalized online drift $h(z)$ in~\eqref{eq:online-H-h} cannot satisfy the global Lipschitz and ODE@$\infty$ radial-scaling hypotheses of \citet[Assumptions~(A2), (A3), and~(A5a)]{borkar2025ode} over the entire open domain $\R^d\times\R^m\times\mathbb S^d_{++}$, for two structural reasons:
\begin{enumerate}[leftmargin=2.0em,itemsep=0.05em]
\item[\textbf{(a)}] \emph{Boundary singularity on $\partial\mathbb S^d_{++}$:} As $\lambda_{\min}(\widehat S)\downarrow 0$, the inverse-Fisher metric map $\widehat S\mapsto M_{\widehat S}=(\tr\widehat S/d)\widehat S^{-1}$ is unbounded, so $h(z)$ is only locally Lipschitz on $\mathbb S^d_{++}$.
\item[\textbf{(b)}] \emph{Super-linear growth of $Q_M(\theta)$ at infinity:} When $\alpha>0$, the surrogate second moment $Q_M(\theta)$ grows super-linearly as $\|\theta\|\to\infty$. Even for the standard Gaussian target $\pi=\gN(0,I_d)$ with $s(x)=-x$ and $z=(\theta,0,I_d)$ ($\theta\neq 0$), completing the square in~\eqref{eq:ga-surrogate} gives $\pi_{r\theta}^{(I_d)}=\gN(\alpha r\theta,I_d)$, so $M_{rI_d}=I_d$ and $Q_{I_d}(r\theta)=\E_{\pi_{r\theta}^{(I_d)}}[XX^\top]=I_d+\alpha^2 r^2\theta\theta^\top$. The $\widehat S$-block of the rescaled drift $r^{-1}h(rz)$ is therefore $r^{-1}I_d+\alpha^2 r\theta\theta^\top-I_d$, which diverges as $r\to\infty$.
\end{enumerate}
This is precisely why Theorem~\ref{thm:online} postulates sample-path boundedness of $\widehat S_n$ in compact subsets of $\mathbb S^d_{++}$ via Assumption~\ref{ass:online-stability}(a) (the standard localization route of \citet[Section~3]{borkar2025ode}). Specifically, since $K_{\theta,M}$ and $(S_M(\theta),F_M(\theta),Q_M(\theta))$ depend only on $(\theta,\widehat S)$ and not on $\mu$, we define for each integer $j\ge 1$ the compact subset of $\R^d\times\mathbb S^d_{++}$:
\begin{equation}
\label{eq:Kj-def}
\mathcal K_j\;\triangleq\;\bigl\{(\theta,\widehat S)\in\R^d\times\mathbb S^d:\;\|\theta\|\le j,\;\;j^{-1}I_d\preceq\widehat S\preceq jI_d\bigr\},
\end{equation}
which is closed and bounded in $\R^d\times\mathbb S^d$ by continuity of the extreme eigenvalues \citep[Corollary~4.3.15]{horn2012matrix} and hence compact by the Heine--Borel theorem. Fix a smooth cutoff function $\chi_j\in C^\infty(\R^d\times\mathbb S^d;[0,1])$ satisfying $\chi_j\equiv 1$ on $\mathcal K_j$ and $\mathrm{supp}(\chi_j)\subset\mathcal K_{2j}$ \citep[Proposition~2.25]{lee2013smooth}, and define the localized metric and score-history maps (viewed as functions of $z=(\theta,\mu,\widehat S)$ via $(\theta,\widehat S)$):
\begin{equation}
\label{eq:localized-M-theta}
\bar M_j(z)\triangleq M_{\chi_j(\theta,\widehat S)\widehat S+(1-\chi_j(\theta,\widehat S))I_d},\quad
\bar\theta_j(z)\triangleq\begin{cases}
\chi_j(\theta,\widehat S)\,\bar M_j(z)^{-1}M_{\widehat S}\theta, & (\theta,\widehat S)\in\mathrm{supp}(\chi_j),\\[2pt]
0, & (\theta,\widehat S)\notin\mathrm{supp}(\chi_j).
\end{cases}
\end{equation}
The prefactor $\bar M_j(z)^{-1}M_{\widehat S}$ in $\bar\theta_j(z)$ is chosen so that multiplying $\bar\theta_j(z)$ on the left by $\bar M_j(z)$ cancels $\bar M_j(z)^{-1}$ and preserves the exact alignment
\begin{equation}
\label{eq:alignment-preserved}
\bar M_j(z)\bar\theta_j(z)\;=\;\chi_j(\theta,\widehat S)\,M_{\widehat S}\theta
\qquad\text{for all }z\in\R^d\times\R^m\times\mathbb S^d_{++},
\end{equation}
including in the transition shell $\mathcal K_{2j}\setminus\mathcal K_j$ where $0<\chi_j(\theta,\widehat S)<1$. Consequently, the localized surrogate invariant density $\pi_{\bar\theta_j(z)}^{(\bar M_j(z))}=\pi_{\bar M_j(z)\bar\theta_j(z)}^{(I_d)}=\pi_{\chi_j\theta}^{(M_{\widehat S})}$ always tilts along $M_{\widehat S}\theta\propto\widehat S^{-1}\theta$, which will allow the moving-metric Lyapunov function $V(\theta,\widehat S)=\theta^\top\widehat S^{-1}\theta$ to decrease globally along the localized ODE flow in Step~4a. Moreover, because convex combinations of matrices in $\mathcal K_{2j}$ and $I_d$ satisfy $(2j)^{-1}I_d\preceq\chi_j\widehat S+(1-\chi_j)I_d\preceq 2jI_d$, Lemma~\ref{lem:metric-map}(i) guarantees that $z\mapsto(\bar\theta_j(z),\bar M_j(z))$ is $C^\infty$ on the entire vector space $\R^d\times\R^m\times\mathbb S^d$, coincides with $(\theta,M_{\widehat S})$ on $\mathcal K_j$, equals $(0,I_d)$ outside $\mathcal K_{2j}$, and takes values in a compact subset of $\R^d\times\mathbb S^d_{++}$.

Let $(\tilde z_n,\tilde X_n)$ be the localized chain initialized at $(\tilde z_0,\tilde X_0)=(z_0,X_0)$ and driven by the recursion
\begin{equation}
\label{eq:localized-chain}
\tilde X_{n+1}\;\sim\;K_{\bar\theta_j(\tilde z_n),\bar M_j(\tilde z_n)}(\tilde X_n,\cdot),\qquad
\tilde z_{n+1}\;=\;\tilde z_n+\gamma_{n+1}\,H(\tilde z_n,\tilde X_{n+1}),
\end{equation}
with associated mean-field drift
\begin{equation}
\label{eq:hj-def}
h_j(z)\;\triangleq\;\bigl(S_{\bar M_j(z)}(\bar\theta_j(z))-\theta,\;F_{\bar M_j(z)}(\bar\theta_j(z))-\mu,\;Q_{\bar M_j(z)}(\bar\theta_j(z))-\widehat S\bigr).
\end{equation}
By construction, $h_j(z)=h(z)$ whenever $(\theta,\widehat S)\in\mathcal K_j$, and outside $\mathcal K_{2j}$ we have $\bar\theta_j(z)=0$, so $\pi_{\bar\theta_j(z)}^{(\bar M_j(z))}=\pi$ and
\begin{equation}
\label{eq:hj-plus-z}
g_j(z)\;\triangleq\;h_j(z)+z\;=\;\bigl(S_{\bar M_j(z)}(\bar\theta_j(z)),\;F_{\bar M_j(z)}(\bar\theta_j(z)),\;Q_{\bar M_j(z)}(\bar\theta_j(z))\bigr)
\end{equation}
is equal to the constant $z^\star=(0,\mu,S)$ outside $\mathcal K_{2j}$. Since $g_j$ is $C^\infty$ and constant outside a compact set in $(\theta,\widehat S)$ (and independent of $\mu$), $g_j$ is globally bounded by some constant $M_j<\infty$ and globally Lipschitz on $\R^d\times\R^m\times\mathbb S^d$.

Finally, coupling $(\tilde z_n,\tilde X_n)$ with $(z_n,X_n)$ using identical driving innovations up to the first exit time of $(\theta_n,\widehat S_n)$ from $\mathcal K_j$, the two chains coincide for all $n\ge 0$ on the event
\begin{equation}
\label{eq:Ej-def}
\mathsf E_j\;\triangleq\;\bigl\{(\theta_n,\widehat S_n)\in\mathcal K_j\;\;\text{for all }n\ge 0\bigr\}.
\end{equation}
By Assumption~\ref{ass:online-stability}(a) and Lemma~\ref{lem:shared-gain} ($\|\theta_n\|^2<\lambda_{\max}(\widehat S_n)$), the events $\mathsf E_j$ are nondecreasing in $j$ and satisfy $\lim_{j\to\infty}\Pr(\mathsf E_j)=\Pr(\bigcup_{j\ge 1}\mathsf E_j)=1$. Throughout the proof below, we fix $j$ large enough that $(0,S)$ lies in the interior of $\mathcal K_j$.

\begin{proof}[Proof of Theorem~\ref{thm:online}]
\textbf{Step 1: Step-size conditions (A1, A5b).} The step-size schedule $\gamma_n=(n+n_0)^{-\rho}$ ($n_0\ge 1$) with $\rho\in(\tfrac12,1]$ is shared across all coordinates of $\tilde z_n$, so \citet[Assumptions~(A1) and~(A5b)]{borkar2025ode} hold verbatim as in Step~1 of Appendix~\ref{app:clt-proof}, with $\lim_{n\to\infty}(\gamma_{n+1}^{-1}-\gamma_n^{-1})=\mathbf 1_{\{\rho=1\}}$.

\textbf{Step 2: Lipschitz continuity and $W$-weighted growth of $H$ via coordinate rescaling (A2, A4a).} By~\eqref{eq:online-H-h}, $z\mapsto H(z,x)=H(0,x)-z$ is $1$-Lipschitz in $z$ for every $x$, and Assumption~\ref{ass:online-stability}(b) gives $\|H(0,x)\|\le\|s(x)\|+\|f(x)\|+\|s(x)\|^2\le C\,W(x)$ for a constant $C>0$. Because the empirical Fisher term $\|s(x)s(x)^\top\|=\|s(x)\|^2$ is $O(W(x))$ rather than $o(W(x))$, we use the non-$o(W)$ stability criterion of \citet[Assumption~(A4a) and Proposition~2]{borkar2025ode}, which requires the $W$-weighted Lipschitz envelope ratio $\delta_L\triangleq\sup_{x\in\R^d}L(x)/W(x)$ from \citet[Assumption~(A2)]{borkar2025ode} to be sufficiently small (specifically, $\delta_L<1/(4T_r)$ for the ODE@$\infty$ relaxation time $T_r$ determined in Step~3). To make $\delta_L$ arbitrarily small without altering the recursion, we pass to the linearly rescaled coordinate $y_n\triangleq\varepsilon\tilde z_n$ ($\varepsilon>0$), which satisfies $y_{n+1}=y_n+\gamma_{n+1}H_\varepsilon(y_n,\tilde X_{n+1})$ with $H_\varepsilon(y,x)\triangleq\varepsilon H(\varepsilon^{-1}y,x)=\varepsilon H(0,x)-y$. The rescaled map $H_\varepsilon(\cdot,x)$ remains $1$-Lipschitz in $y$ and satisfies $\|H_\varepsilon(0,x)\|\le\varepsilon C\,W(x)$, so its Lipschitz/growth envelope in \citet[Assumption~(A2)]{borkar2025ode} is $L_\varepsilon(x)=\max\{1,\varepsilon C\,W(x)\}$. By \citet[Lemma~2(i)]{borkar2025ode}, replacing $W(x)$ with the truncated Lyapunov function $W_3(x)\triangleq\max\{3,W(x)\}$ preserves the (DV3) drift condition and yields
\begin{equation}
\label{eq:delta-L-bound}
\delta_L\;=\;\sup_{x\in\R^d}\frac{L_\varepsilon(x)}{W_3(x)}\;\le\;\max\!\Bigl\{\frac{1}{3},\;\varepsilon C\Bigr\}\;=\;\frac{1}{3}
\qquad\text{for all }\varepsilon\in\bigl(0,(3C)^{-1}\bigr].
\end{equation}
By linearity of $y_n=\varepsilon\tilde z_n$, almost-sure convergence and the CLT for $y_n$ transfer immediately back to $\tilde z_n$.

\textbf{Step 3: ODE@$\infty$ stability for the localized drift (A3) and completion of (A4a).} For the localized drift $h_j(z)=-z+g_j(z)$ in~\eqref{eq:hj-def}, the map $g_j$ in~\eqref{eq:hj-plus-z} is globally bounded by $M_j<\infty$. Consequently, as $r\to\infty$,
\[
\sup_{\|z\|=1}\bigl\|r^{-1}h_j(rz)-(-z)\bigr\|\;=\;\sup_{\|z\|=1}r^{-1}\bigl\|g_j(rz)\bigr\|\;\le\;\frac{M_j}{r}\;\longrightarrow\;0.
\]
Thus the limiting ODE@$\infty$ is $\dot z_\infty=-z_\infty$, whose flow $\phi_t^\infty(z_0)=e^{-t}z_0$ contracts the unit ball to radius $1/2$ at the relaxation time $T_r=\log 2\approx 0.693<3/4$. Combining $T_r<3/4$ with $\delta_L\le 1/3$ from~\eqref{eq:delta-L-bound} immediately gives $\delta_L\le 1/3<1/(4\log 2)=1/(4T_r)$, completing the verification of \citet[Assumptions~(A3) and~(A4a) and Proposition~2]{borkar2025ode}.

\textbf{Step 4: Moving-metric Lyapunov function, Hurwitz Jacobian, and global contraction (A5a, A5b).} Because $g_j(z)=h_j(z)+z$ is $C^\infty$ and constant outside $\mathcal K_{2j}$, $h_j$ is globally Lipschitz and smooth, verifying the regularity part of \citet[Assumption~(A5a)]{borkar2025ode}. We establish global exponential stability of $\dot z=h_j(z)$ in three sub-steps:

\emph{Step 4a: Uniform positive definiteness of $\widehat S_t$, moving-metric Lyapunov decay, and global asymptotic stability.} Since $\eta=\bar M_j(z)\bar\theta_j(z)$ ranges over a compact set $\mathcal C_j\subset\R^d$ and $\eta\mapsto\lambda_{\min}(Q_{I_d}(\eta))$ and $\eta\mapsto\lambda_{\max}(Q_{I_d}(\eta))$ are continuous \citep[Corollary~4.3.15]{horn2012matrix} with $Q_{I_d}(\eta)\succ 0$ on $\R^d$, Weierstrass's extreme value theorem \citep[Theorem~4.16]{rudin1976principles} yields constants $0<q_j\le Q_j^{\max}<\infty$ such that $q_j I_d\preceq Q_{\bar M_j(z)}(\bar\theta_j(z))\preceq Q_j^{\max}I_d$ for all $z$. Along any trajectory of $\dot z=h_j(z)$, the $\widehat S$-component satisfies $q_j I_d-\widehat S_t\preceq\dot{\widehat S}_t=Q_{\bar M_j}(\bar\theta_j)-\widehat S_t\preceq Q_j^{\max}I_d-\widehat S_t$; multiplying by $e^t$ and integrating over $[0,t]$ (variation of constants) gives
\begin{equation}
\label{eq:St-lower-bound}
e^{-t}\widehat S_0+(1-e^{-t})q_j I_d\;\preceq\;\widehat S_t\;\preceq\;e^{-t}\widehat S_0+(1-e^{-t})Q_j^{\max}I_d.
\end{equation}
Hence, from any initial $\widehat S_0\in\mathbb S^d$, $\widehat S_t$ enters the open cone $\mathbb S^d_{++}$ at a finite time $t_0\ge 0$ ($t_0=0$ when $\widehat S_0\succ 0$) and satisfies $\underline\lambda I_d\preceq\widehat S_t\preceq\overline\lambda I_d$ for all $t\ge t_0$ for some $0<\underline\lambda\le\overline\lambda<\infty$. For $t\ge t_0$, define the moving-metric Lyapunov function
\begin{equation}
\label{eq:moving-lyap}
V(\theta,\widehat S)\;\triangleq\;\theta^\top\widehat S^{-1}\theta.
\end{equation}
Using the inverse-derivative identity $\frac{d}{dt}(\widehat S_t^{-1})=-\widehat S_t^{-1}\dot{\widehat S}_t\widehat S_t^{-1}$ \citep[Appendix~A.4.1]{boyd2004convex} together with $\dot\theta_t=S_{\bar M_j}(\bar\theta_j)-\theta_t$ and $\dot{\widehat S}_t=Q_{\bar M_j}(\bar\theta_j)-\widehat S_t$, the time derivative of $V$ along the flow for $t\ge t_0$ is
\begin{align}
\dot V(\theta_t,\widehat S_t)
&=2\theta_t^\top\widehat S_t^{-1}\bigl(S_{\bar M_j}(\bar\theta_j)-\theta_t\bigr)-\theta_t^\top\widehat S_t^{-1}\bigl(Q_{\bar M_j}(\bar\theta_j)-\widehat S_t\bigr)\widehat S_t^{-1}\theta_t\nonumber\\
&=2\theta_t^\top\widehat S_t^{-1}S_{\bar M_j}(\bar\theta_j)-\theta_t^\top\widehat S_t^{-1}Q_{\bar M_j}(\bar\theta_j)\widehat S_t^{-1}\theta_t-V(\theta_t,\widehat S_t)\nonumber\\
&\le 2\theta_t^\top\widehat S_t^{-1}S_{\bar M_j}(\bar\theta_j)-V(\theta_t,\widehat S_t)\;\le\;-V(\theta_t,\widehat S_t).
\label{eq:moving-lyap-decay}
\end{align}
Here the second equality combines $-2\theta_t^\top\widehat S_t^{-1}\theta_t+\theta_t^\top\widehat S_t^{-1}\widehat S_t\widehat S_t^{-1}\theta_t=-V(\theta_t,\widehat S_t)$; the first inequality drops $-\theta_t^\top\widehat S_t^{-1}Q_{\bar M_j}(\bar\theta_j)\widehat S_t^{-1}\theta_t\le 0$ (since $Q_{\bar M_j}\succeq 0$); and the second inequality uses $\theta_t^\top\widehat S_t^{-1}S_{\bar M_j}(\bar\theta_j)\le 0$, which follows from the alignment~\eqref{eq:alignment-preserved}: whenever $\chi_j(\theta_t,\widehat S_t)>0$, we have $S_{\bar M_j}(\bar\theta_j)=S_{M_{\widehat S_t}}(\chi_j\theta_t)$ with $M_{\widehat S_t}=(\tr\widehat S_t/d)\widehat S_t^{-1}$ a positive scalar multiple of $\widehat S_t^{-1}$, so the log-partition convexity bound~\eqref{eq:convexity-bound} evaluated at $\chi_j\theta_t$ gives $\theta_t^\top\widehat S_t^{-1}S_{M_{\widehat S_t}}(\chi_j\theta_t)\le 0$; whenever $\chi_j(\theta_t,\widehat S_t)=0$, $S_{\bar M_j}(0)=\E_\pi[s(X)]=0$. By Gr\"{o}nwall's inequality \citep[Appendix~B]{borkar2009stochastic}, $V(\theta_t,\widehat S_t)\le e^{-(t-t_0)}V(\theta_{t_0},\widehat S_{t_0})\to 0$, and since $\widehat S_t\preceq\overline\lambda I_d$, we obtain $\|\theta_t\|^2\le\overline\lambda\,V(\theta_t,\widehat S_t)\to 0$ as $t\to\infty$. Because $\bar M_j(z_t)$ is bounded on $\mathcal K_{2j}$, $\theta_t\to 0$ implies $\eta_t\triangleq\bar M_j(z_t)\bar\theta_j(z_t)\to 0$, so by continuity $Q_{\bar M_j(z_t)}(\bar\theta_j(z_t))=Q_{I_d}(\eta_t)\to Q_{I_d}(0)=S$ and $F_{\bar M_j(z_t)}(\bar\theta_j(z_t))=F_{I_d}(\eta_t)\to F_{I_d}(0)=\mu$. Feeding these converging forcing terms into the linear inhomogeneous ODEs $\dot{\widehat S}_t=-\widehat S_t+Q_{I_d}(\eta_t)$ and $\dot\mu_t=-\mu_t+F_{I_d}(\eta_t)$ yields $\widehat S_t\to S$ and $\mu_t\to\mu$ as $t\to\infty$ by variation of constants.

\emph{Step 4b: Block-lower-triangular Hurwitz Jacobian via $\pi_0^{(M)}=\pi$.} Because $(0,S)\in\mathrm{int}(\mathcal K_j)$, we have $\chi_j\equiv 1$ and $h_j(z)=h(z)$ in a neighborhood of $z^\star=(0,\mu,S)$. There, the maps $S_{M_{\widehat S}}(\theta)=S_{I_d}(M_{\widehat S}\theta)$, $F_{M_{\widehat S}}(\theta)=F_{I_d}(M_{\widehat S}\theta)$, and $Q_{M_{\widehat S}}(\theta)=Q_{I_d}(M_{\widehat S}\theta)$ depend on $\widehat S$ only through the product $\eta=M_{\widehat S}\theta$. By the chain rule, the directional derivative of $S_{M_{\widehat S}}(\theta)$ with respect to $\widehat S$ along any $E\in\mathbb S^d$ evaluated at $(\theta,\widehat S)=(0,S)$ carries a factor of $\theta=0$ and therefore vanishes identically:
\[
\frac{\partial}{\partial\widehat S}\bigl[S_{I_d}(M_{\widehat S}\theta)\bigr]\Big|_{(\theta,\widehat S)=(0,S)}[E]
\;=\;\frac{\partial S_{I_d}}{\partial\eta}(0)\,DM_S[E]\,\theta\Big|_{\theta=0}\;=\;0,
\]
and identically $\frac{\partial}{\partial\widehat S}[F_{M_{\widehat S}}(\theta)]_{(0,S)}=0$ and $\frac{\partial}{\partial\widehat S}[Q_{M_{\widehat S}}(\theta)]_{(0,S)}=0$. Combining Lemma~\ref{lem:jacobian-form} for the $(\theta,\mu)$ derivatives with Lemma~\ref{lem:deriv-id} applied to $h=ss^\top$ at $\theta=0$ for the $\widehat S$ block therefore yields the block-lower-triangular Jacobian
\begin{equation}
\label{eq:online-jacobian}
J_\star\;\triangleq\;\nabla h_j(z^\star)\;=\;\begin{bmatrix}
J_\star^{(M_{\rm opt})} & 0\\[2pt]
\bigl[-\alpha\,\Cov_\pi(ss^\top,s)\,M_{\rm opt}\;\;\;0\bigr] & -I_{d(d+1)/2}
\end{bmatrix}.
\end{equation}
By Lemma~\ref{lem:metric-map}(iii), $SM_{\rm opt}=(\tr S/d)I_d$, so the eigenvalues of $J_\star$ are $-1-\alpha\tr S/d$ (with multiplicity $d$) and $-1$ (with multiplicity $m+d(d+1)/2$). Consequently, $\frac12\mathbf 1_{\{\rho=1\}}I+J_\star$ is Hurwitz and $z^\star$ is the unique globally asymptotically stable equilibrium of $\dot z=h_j(z)$.

\emph{Step 4c: Global exponential contraction bound (A5b).} Because $J_\star=\nabla h_j(z^\star)$ is Hurwitz with spectral abscissa $-1$, local exponential stability \citep[Theorem~4.7]{khalil2002nonlinear} provides a positively invariant neighborhood $U_\delta=\{\|z-z^\star\|\le\delta_0\}$ ($\delta_0>0$) and constants $c_0\ge 1$, $\rho_e\in(0,1)$ such that $\|\phi_t(z_0)-z^\star\|\le c_0\|z_0-z^\star\|e^{-\rho_e t}$ for all $z_0\in U_\delta$ and $t\ge 0$. For trajectories starting outside $U_\delta$ ($\|z_0-z^\star\|>\delta_0$), writing $\dot z_t=-z_t+g_j(z_t)$ with $\|g_j(z)\|\le M_j$ and applying variation of constants gives
\[
\|\phi_t(z_0)-z^\star\|\;\le\;e^{-t}\|z_0-z^\star\|+(1-e^{-t})\bigl(M_j+\|z^\star\|\bigr)\qquad\text{for all }t\ge 0.
\]
Hence every trajectory enters the compact ball $B_R\triangleq\{\|z-z^\star\|\le R\}$ with $R\triangleq 2(M_j+\|z^\star\|)$ within time $T_1(z_0)\triangleq\max\{0,\log(\|z_0-z^\star\|/(M_j+\|z^\star\|))\}$. By global asymptotic stability (Step~4a) and continuity of the flow $\phi_t$, compactness of $B_R$ yields a uniform finite transit time $T_2<\infty$ such that $\phi_t(B_R)\subset U_\delta$ for all $t\ge T_2$. Using $1<\delta_0^{-1}\|z_0-z^\star\|$ on $\{z_0\notin U_\delta\}$ to absorb the bounded transient $(M_j+\|z^\star\|)$ over $[0,T_1(z_0)+T_2]$ and splicing with the exponential decay inside $U_\delta$ for $t\ge T_1(z_0)+T_2$ yields $\|\phi_t(z_0)-z^\star\|\le c_e\|z_0-z^\star\|e^{-\rho_e t}$ for all $z_0$ and all $t\ge 0$, verifying \citet[Assumption~(A5b), Eq.~(43)]{borkar2025ode}.

\textbf{Step 5: Base kernel regularity (DV3 and global Lipschitzness in A2).} Because $(\bar\theta_j(z),\bar M_j(z))$ takes values in a compact subset of $\R^d\times\mathbb S^d_{++}$, Assumption~\ref{ass:online-stability}(b) provides the uniform (DV3) drift condition on the sublevel sets of $W$ (and hence of $W_3$) and Lipschitz continuity of $(\theta,M)\mapsto K_{\theta,M}$ on that compact range. Composing with the smooth map $z\mapsto(\bar\theta_j(z),\bar M_j(z))$---which is constant outside $\mathcal K_{2j}$ and globally Lipschitz by Lemma~\ref{lem:metric-map}(i)---makes $z\mapsto K_{\bar\theta_j(z),\bar M_j(z)}$ globally Lipschitz in the $W_3^\delta$-norm across the entire state space, fulfilling \citet[Assumption~(A2)]{borkar2025ode}.

\textbf{Step 6: Conclusion via the SA CLT and transfer to the unlocalized chain.} Having verified \citet[Assumptions~(A1)--(A5)]{borkar2025ode} for the localized chain $\tilde z_n$ in Steps~1--5, \citet[Theorem~2]{borkar2025ode} bounds the moments of $\tilde z_n$ and of the exponential (DV3) Lyapunov function uniformly in $n$. Because (DV3) dominates $W_3$ by the exponential Lyapunov function plus a constant, applying \citet[Lemma~1 and Propositions~7(ii) and~8]{borkar2025ode} to the Markovian noise increments $\Delta_{k+1}\triangleq H(\tilde z_k,\tilde X_{k+1})-h_j(\tilde z_k)$ shows that the noise series $R_n\triangleq\sum_{k=0}^{n-1}\gamma_{k+1}\Delta_{k+1}$ converges almost surely (its martingale part is bounded in $L^2$ and converges by Doob's martingale convergence theorem, while the Poisson-equation boundary and drift-difference terms are absolutely summable). Writing $\tilde z_{n+1}=(1-\gamma_{n+1})\tilde z_n+\gamma_{n+1}g_j(\tilde z_n)+(R_{n+1}-R_n)$ with $g_j$ globally bounded by $M_j$ and $R_n$ convergent, the Kushner--Clark / ODE-method lemma \citep[Chapter~2, Lemma~1 and Theorem~2]{borkar2009stochastic} implies that $\tilde z_n$ is almost surely bounded and converges almost surely to the unique globally asymptotically stable equilibrium $z^\star$ of $\dot z=h_j(z)$ from Step~4: $\tilde z_n\xrightarrow{a.s.}z^\star$.

Simultaneously, \citet[Theorem~4]{borkar2025ode} applied to $\tilde z_n$ yields the joint CLT
\begin{equation}
\label{eq:localized-clt}
\gamma_n^{-1/2}(\tilde z_n-z^\star)\;\xrightarrow{dist.}\;\gN(0,\Sigma_z),
\end{equation}
where $\Sigma_z$ is the unique symmetric positive-semidefinite solution of the Lyapunov equation
\begin{equation}
\label{eq:online-lyapunov-full}
\bigl(\tfrac12\mathbf 1_{\{\rho=1\}}I+J_\star\bigr)\Sigma_z+\Sigma_z\bigl(\tfrac12\mathbf 1_{\{\rho=1\}}I+J_\star\bigr)^\top+U_{zz}\;=\;0,
\end{equation}
and $U_{zz}$ is the long-run covariance~\eqref{eq:long-run-cov-U} of the centered residual $H(z^\star,\cdot)$ along a stationary trajectory of the equilibrium kernel $K_{0,M_{\rm opt}}$. Crucially, neither $J_\star$ nor $U_{zz}$ depends on the localization index~$j$.

To transfer both conclusions from $\tilde z_n$ to the true online iterate $z_n$, recall from~\eqref{eq:Ej-def} that $z_n=\tilde z_n$ for all $n\ge 0$ on the event $\mathsf E_j$, and that $\lim_{j\to\infty}\Pr(\mathsf E_j)=1$ by Assumption~\ref{ass:online-stability}(a) and Lemma~\ref{lem:shared-gain}. Therefore,
\[
\Pr(z_n\to z^\star)\;\ge\;\Pr\bigl(\{z_n=\tilde z_n\;\forall n\}\cap\{\tilde z_n\to z^\star\}\bigr)\;=\;\Pr(\mathsf E_j)\;\xrightarrow{j\to\infty}\;1,
\]
proving $z_n\xrightarrow{a.s.}z^\star$. Likewise, for any bounded continuous test function $\psi$,
\[
\Bigl|\E\bigl[\psi\bigl(\gamma_n^{-1/2}(z_n-z^\star)\bigr)\bigr]-\E\bigl[\psi\bigl(\gamma_n^{-1/2}(\tilde z_n-z^\star)\bigr)\bigr]\Bigr|
\;\le\;2\|\psi\|_\infty\,\Pr(\mathsf E_j^c).
\]
Taking $\limsup_{n\to\infty}$ via~\eqref{eq:localized-clt} and then letting $j\to\infty$ shows that $\E[\psi(\gamma_n^{-1/2}(z_n-z^\star))]\to\E[\psi(Z)]$ with $Z\sim\gN(0,\Sigma_z)$, which by the portmanteau theorem \citep[Theorem~2.1]{billingsley1999convergence} yields $\gamma_n^{-1/2}(z_n-z^\star)\xrightarrow{dist.}\gN(0,\Sigma_z)$. (When the initial state $X_0$ is random, applying the above argument conditionally on $X_0$ and taking expectations via the bounded convergence theorem transfers weak convergence unconditionally.)

\textbf{Step 7: Decoupling the $(\theta,\mu)$ marginal.} Partitioning $z=(\vartheta,\widehat S)$ into the $(d+m)$-dimensional parameter block $\vartheta=(\theta,\mu)$ and the $d(d+1)/2$-dimensional Fisher block $\widehat S$, the shifted Jacobian in~\eqref{eq:online-jacobian}, the asymptotic covariance $\Sigma_z$, and the long-run noise covariance $U_{zz}$ take the $2\times 2$ block forms
\[
\tfrac12\mathbf 1_{\{\rho=1\}}I+J_\star=\begin{bmatrix} A_{11} & 0 \\ A_{21} & A_{22} \end{bmatrix},\qquad
\Sigma_z=\begin{bmatrix} \Sigma_{\vartheta\vartheta} & \Sigma_{\vartheta S} \\ \Sigma_{S\vartheta} & \Sigma_{SS} \end{bmatrix},\qquad
U_{zz}=\begin{bmatrix} U_{HH} & U_{\vartheta S} \\ U_{S\vartheta} & U_{SS} \end{bmatrix},
\]
where $A_{11}\triangleq\tfrac12\mathbf 1_{\{\rho=1\}}I_{d+m}+J_\star^{(M_{\rm opt})}$ and the upper-left block of $U_{zz}$ is $U_{HH}$ from Section~\ref{sec:ga-srmc} because the first two components of $H(z^\star,\cdot)$ coincide with the fixed-$M$ residual $H(\vartheta^\star,\cdot)$ under the equilibrium kernel $K_{0,M_{\rm opt}}$. Expanding the $(1,1)$ block of the full Lyapunov equation~\eqref{eq:online-lyapunov-full} gives
\[
\bigl(A_{11}\Sigma_{\vartheta\vartheta}+0\cdot\Sigma_{S\vartheta}\bigr)+\bigl(\Sigma_{\vartheta\vartheta}A_{11}^\top+\Sigma_{\vartheta S}\cdot 0^\top\bigr)+U_{HH}
\;=\;A_{11}\Sigma_{\vartheta\vartheta}+\Sigma_{\vartheta\vartheta}A_{11}^\top+U_{HH}\;=\;0,
\]
which is precisely the fixed-$M$ Lyapunov equation~\eqref{eq:lyapunov-eq} evaluated at $M=M_{\rm opt}$. By uniqueness of the solution to~\eqref{eq:lyapunov-eq}, we obtain $\Sigma_{\vartheta\vartheta}=\Sigma_\vartheta^{(M_{\rm opt})}$, establishing part~(ii). (Although the off-diagonal cross-covariance $\Sigma_{\vartheta S}$ between $(\theta_n,\mu_n)$ and $\widehat S_n$ is generically nonzero, the zero upper-right block of $J_\star$ prevents $\Sigma_{\vartheta S}$ from entering the $(1,1)$ block equation.)

\textbf{Step 8: Asymptotic unbiasedness and the learned metric.} Since $\widehat S_n\xrightarrow{a.s.}S\succ 0$ by Step~6 and $\widehat S\mapsto M_{\widehat S}$ is continuous at $S$ by Lemma~\ref{lem:metric-map}(i), the continuous mapping theorem \citep[Theorem~2.3]{van1998asymptotic} gives $M_n=M_{\widehat S_n}\xrightarrow{a.s.}M_S=M_{\rm opt}$, completing part~(i). Consequently, $M_n\theta_n\xrightarrow{a.s.}0$, so the surrogate density $\pi_{\theta_n}^{(M_n)}=\pi_{M_n\theta_n}^{(I_d)}$ converges pointwise to $\pi$ and $\mu_n\xrightarrow{a.s.}\mu=\E_\pi[f(X)]$. Because the $(\theta,\mu)$ marginal CLT in Step~7 matches the fixed-$M$ CLT at $M=M_{\rm opt}$, the optimal-$M$ variance bounds~\eqref{eq:theta-universal-bound}--\eqref{eq:theta-exact-opt} of Theorem~\ref{thm:optimal} hold at the learned metric under the equilibrium kernel $K_{0,M_{\rm opt}}$.
\end{proof}

The same localization argument applies verbatim to the Tikhonov-regularized online metric $M_{\widehat S_{n,\delta}}$ with $\widehat S_{n,\delta}=\widehat S_n+\delta I_d$ ($\delta>0$) used in Section~\ref{sec:experiments}, replacing $\widehat S^{-1}$ by $(\widehat S+\delta I_d)^{-1}$ in the moving-metric Lyapunov function~\eqref{eq:moving-lyap}. In that case, only the upper bound $\sup_{n\ge 0}\lambda_{\max}(\widehat S_n)<\infty$ in Assumption~\ref{ass:online-stability}(a) is needed (since $\widehat S_{n,\delta}\succeq\delta I_d$ is automatically bounded away from zero), and the learned metric converges almost surely to $M_{S+\delta I_d}$, which recovers $M_{\rm opt}$ as $\delta\downarrow 0$ (or whenever $S$ is isotropic). When $\widehat S_n$ is instead frozen after a finite warmup period (as in the \texttt{freeze\_after\_L} option of Appendix~\ref{app:online-fisher-details}), the subsequent chain evolves with a constant positive-definite metric $M$ and is governed directly by Proposition~\ref{prop:clt}. Finally, the same localization also relaxes the kernel hypotheses of Proposition~\ref{prop:clt} to compact sets: for any fixed $M\succ 0$, omitting the $\widehat S$ coordinate and using the quadratic Lyapunov function~\eqref{eq:lyap-V} in Step~4 yields the conclusions of Proposition~\ref{prop:clt} whenever $\sup_{n\ge 0}\|\theta_n\|<\infty$ a.s.\ and Assumption~\ref{ass:kernel} holds on compact subsets of $\theta$ with $\|s(x)\|+\|f(x)\|\le c\,W(x)$.

\section{Additional Discussion}
\label{app:implementation-details}

\paragraph{Summary of Appendix \ref{app:implementation-details}.}
This appendix collects two threads of discussion that did not fit into the main body. Appendix~\ref{apx:alt-repulsion} explains why two natural-looking alternatives to the bilinear repulsion exponent $\theta^\top M s(x)$, i.e., cosine similarity and score-norm normalization, are not viable substitutes: cosine similarity destroys the asymptotic theory at the equilibrium $\theta^\star\!=\!0$, while score-norm normalization either breaks the strong-Lyapunov argument that underlies the global convergence guarantee or, if the history recursion is also normalized, breaks Stein's identity and biases the Monte Carlo estimator. Appendix~\ref{app:online-fisher-details} gives a self-contained description of the two online Fisher estimators introduced in Section~\ref{sec:ga-srmc-practical} (diagonal and full), each with the explicit \textsc{FisherUpdate} and \textsc{InvFisher} recursions, the trace-normalized inverse used in \eqref{eq:online-fisher-trace}, and the per-iteration cost.

\subsection{Why not cosine similarity or score-norm normalization?}
\label{apx:alt-repulsion}

A reasonable starting question is whether one could simply `normalize away' the geometry mismatch motivating GA-SRMC by replacing the bilinear inner product $\theta^\top s(x)$ in the SRMC surrogate~\eqref{eq:srmc_surrogate} with a scale-invariant alternative. Two candidates come to mind. One is the cosine similarity
\begin{equation}
\label{eq:cos-sim}
\cos(\theta,s(x))\;\triangleq\;\frac{\theta^\top s(x)}{\|\theta\|\,\|s(x)\|},
\end{equation}
i.e. the (Euclidean) inner product normalized by both $\|\theta\|$ and $\|s(x)\|$, so that it depends only on the angle between $\theta$ and $s(x)$ in $\R^d$. The other is score-norm normalization with a weaker scaling that only divides by $\|s(x)\|$,
\begin{equation}
\label{eq:score-norm}
\theta^\top u(x)\;\triangleq\;\frac{\theta^\top s(x)}{\|s(x)\|},
\qquad u(x)\triangleq \frac{s(x)}{\|s(x)\|},
\end{equation}
which is still linear in $\theta$ but uses the unit-norm direction $u(x)$ in place of the raw score. We argue that neither yields the matched structure used by GA-SRMC. The bilinear form $\theta^\top M s(x)$, with $M$ symmetric positive definite, is the unique replacement that (a) preserves the linear-in-$\theta$ structure that makes the surrogate an exponential family in $\theta$, (b) keeps the equilibrium $\theta^\star\!=\!0$ at which $\pi_{\theta^\star}^{(M)}\!=\!\pi$, and (c) admits the Hessian-of-log-partition derivative identity \eqref{eq:derivative-identity} that drives the SA CLT.

\subsubsection{Cosine similarity destroys the asymptotic theory at \texorpdfstring{$\theta^\star\!=\!0$}{theta=0}}
\label{apx:cosine-failure}

Cosine similarity introduces a nonlinear and singular dependence on $\theta$: the surrogate is undefined at $\theta^\star\!=\!0$, which is precisely the equilibrium to which SRMC must converge. Moreover, the corresponding tilt is no longer an exponential family with a sufficient statistic independent of $\theta$, so the clean derivation of the log-partition function in Appendix~\ref{app:derivative-identity} no longer applies. We make these failures precise as a single proposition with two formal claims.

\begin{proposition}[Two failures of the cosine surrogate]
\label{prop:cosine-failure}
Let $\pi_\theta^{\cos}(x)\!\propto\!\pi(x)\exp\{-\alpha\cos(\theta,s(x))\}$ be the cosine-similarity surrogate, with $\cos(\theta,s(x))$ defined in \eqref{eq:cos-sim} and $\alpha\!>\!0$ fixed. Then:
\begin{enumerate}[leftmargin=2.0em,itemsep=0.05em]
\item[\textbf{(i)}] \textbf{Discontinuous surrogate at the equilibrium.} $\pi_\theta^{\cos}$ does not extend continuously to $\theta^\star\!=\!0$: for any deterministic convention $\cos(0,s(x))\!\triangleq\! c_0(x)\!\in\!\R$, the directional limits $\lim_{t\to 0}\pi_{t v}^{\cos}\!\neq\!\lim_{t\to 0}\pi_{tw}^{\cos}$ disagree for any two non-collinear $v,w\!\in\!\R^d$.
\item[\textbf{(ii)}] \textbf{Non-existent Jacobian at the equilibrium.} The mean-field map $h^{\cos}(\theta)\!\triangleq\!\E_{\pi_\theta^{\cos}}[s(X)]-\theta$ is not differentiable at $\theta^\star\!=\!0$: any candidate Jacobian carries a $1/\|\theta\|$ singularity along every approach direction.
\end{enumerate}
\end{proposition}

\begin{proof}
\textbf{Step 1: Direction-dependence of the limit (claim (i)).} Cosine similarity is scale-invariant in $\theta$: for every $c\!>\!0$, $\cos(c\theta,s(x))\!=\!\cos(\theta,s(x))$, so the surrogate depends only on the direction $\theta/\|\theta\|$. Fix two unit vectors $\hat v\!\neq\!\hat w$ with $\hat v^\top \hat w\!\in\!(-1,1)$, and consider the directional limits along the rays $\theta_t^v\!\triangleq\! t\hat v$ and $\theta_t^w\!\triangleq\! t\hat w$ as $t\!\to\!0$. By scale-invariance,
$$
\lim_{t\to 0}\cos(\theta_t^v,s(x))=\frac{\hat v^\top s(x)}{\|s(x)\|},
\qquad
\lim_{t\to 0}\cos(\theta_t^w,s(x))=\frac{\hat w^\top s(x)}{\|s(x)\|},
$$
and these two functions of $x$ are pointwise distinct since $\hat v\!\neq\!\hat w$. Hence, the surrogate distributions $\lim_{t\to 0}\pi_{\theta_t^v}^{\cos}$ and $\lim_{t\to 0}\pi_{\theta_t^w}^{\cos}$ are different probability measures, and no scalar convention $c_0(x)$ at the origin can reconcile them.

\textbf{Step 2: Singularity of the gradient (claim (ii)).} Differentiating cosine similarity in $\theta$ for $\theta\!\neq\!0$ yields
\begin{equation}
\label{eq:cosine-gradient}
\nabla_\theta \cos(\theta,s(x))
=\frac{1}{\|\theta\|\,\|s(x)\|}
\Bigl(I-\frac{\theta\theta^\top}{\|\theta\|^2}\Bigr)s(x).
\end{equation}
The factor $\|\theta\|^{-1}$ blows up as $\theta\!\to\!0$, so $\nabla_\theta\cos(\theta,s(x))$ is unbounded in any neighborhood of the origin and the dominated-convergence step that justifies differentiating $\E_{\pi_\theta^{\cos}}[s]$ under the integral sign fails. The map $\theta\!\mapsto\!\theta/\|\theta\|$ is itself non-differentiable at $0$, hence so is $h^{\cos}$.
\end{proof}
The cosine surrogate is \emph{designed} to be scale-invariant in $\theta$, which seems desirable as a way to remove the $\alpha$-tuning burden. However, every surrogate must reduce to $\pi$ at $\theta\!=\!0$ in a path-independent way, since the score-history average $\theta_n$ converges to zero from a stochastic path. By contrast, the bilinear form $\theta^\top M s(x)$ is homogeneous of degree one in $\theta$: if $\theta_n\!\to\!0$, then $\theta_n^\top M s(x)\!\to\!0$ pointwise for every $x$, so $\pi_{\theta_n}^{(M)}\!\to\!\pi$ regardless of the path of approach, and the Jacobian $\nabla h_M$ exists and is given in closed form by Lemma~\ref{lem:deriv-id}.

\subsubsection{Score-norm normalization breaks the strong-Lyapunov argument or breaks Stein's identity}
\label{apx:score-norm-failure}

Consider score-norm normalization~\eqref{eq:score-norm}, with the surrogate
\[
\pi_\theta^{\mathrm{dir}}(x)\propto \pi(x)\exp\{-\alpha\,\theta^\top u(x)\},
\qquad u(x)\triangleq s(x)/\|s(x)\|.
\]
Unlike cosine similarity, this form \emph{is} an exponential family in $\theta$, with sufficient statistic $u$. Differentiating its log-partition function under the integral sign (Appendix~\ref{app:derivative-identity}) gives the analog of Lemma~\ref{lem:deriv-id},
\begin{equation}
\label{eq:dir-deriv-id}
\nabla_\theta \E_{\pi_\theta^{\mathrm{dir}}}[h(X)]
\;=\;
-\alpha\,\Cov_{\pi_\theta^{\mathrm{dir}}}(h,u)
\qquad\text{for any integrable }h,
\end{equation}
so the surrogate itself is well-defined. We now show that downstream of this identity, the construction faces a dichotomy.

\begin{proposition}[Score-norm dichotomy]
\label{prop:score-norm-dichotomy}
The score-norm-normalized surrogate $\pi_\theta^{\mathrm{dir}}$ admits exactly two natural pairings with the SRMC score-history recursion, and each fails:
\begin{enumerate}[leftmargin=2.0em,itemsep=0.05em]
\item[\textbf{(i)}] \textbf{Raw-score history.} If the history recursion is the original $\theta_{n+1}\!=\!\theta_n+\gamma_{n+1}(s(X_{n+1})-\theta_n)$, then in $d\!\ge\!2$ the mean field $h_{\mathrm{dir}}(\theta)\!\triangleq\!\E_{\pi_\theta^{\mathrm{dir}}}[s(X)]-\theta$ generically fails to be the gradient of any scalar potential, so the strongly-convex Lyapunov argument used in the proof of Proposition~\ref{prop:fixedM-structure} no longer applies.
\item[\textbf{(ii)}] \textbf{Normalized-score history.} If the history recursion is replaced by $\theta_{n+1}\!=\!\theta_n+\gamma_{n+1}(u(X_{n+1})-\theta_n)$ to restore consistency with the sufficient statistic, then there exist targets $\pi$ for which Stein's identity for $u$ fails ($\E_\pi[u(X)]\!\neq\!0$), and consequently $\theta^\star\!=\!0$ is not the SA fixed point: the surrogate $\pi_{\theta^\star}^{\mathrm{dir}}\!\neq\!\pi$ and the Monte Carlo estimator $\mu_n$ is asymptotically biased.
\end{enumerate}
\end{proposition}

\begin{proof}
\textbf{Step 1: Cross-covariance Jacobian fails to be a Hessian (claim (i)).} Applying \eqref{eq:dir-deriv-id} with $h\!=\!s$ gives the Jacobian
\begin{equation}
\label{eq:dir-jac}
\nabla_\theta S_{\mathrm{dir}}(\theta)\;=\;-\alpha\,\Cov_{\pi_\theta^{\mathrm{dir}}}(s,u),
\qquad S_{\mathrm{dir}}(\theta)\triangleq\E_{\pi_\theta^{\mathrm{dir}}}[s(X)].
\end{equation}
At $\theta\!=\!0$ the cross-covariance collapses to $\E_\pi[s s^\top/\|s\|]$ (since $\E_\pi[s]\!=\!0$ by Stein's identity), which is symmetric PSD; so the linearisation $\nabla h_{\mathrm{dir}}(0)\!=\!-I_d-\alpha\,\E_\pi[s s^\top/\|s\|]\!\prec\!0$ is symmetric negative definite and $\theta^\star\!=\!0$ is locally exponentially stable. Away from the origin, however, $\Cov_{\pi_\theta^{\mathrm{dir}}}(s,u)$ is generically non-symmetric in $d\!\ge\!2$: there is no symmetry condition forcing $\partial S_{\mathrm{dir}}^i/\partial\theta_j\!=\!\partial S_{\mathrm{dir}}^j/\partial\theta_i$ on the off-diagonal entries, since $s$ and $u$ are different statistics. By Poincar\'{e}'s lemma, $h_{\mathrm{dir}}$ is not the gradient of any scalar potential, so the deterministic ODE $\dot\theta_t\!=\!h_{\mathrm{dir}}(\theta_t)$ is not a gradient flow and the strongly-convex Lyapunov argument of Step~3 in the proof of Proposition~\ref{prop:fixedM-structure} no longer goes through. This is genuinely a $d\!\ge\!2$ phenomenon: in $1$-D, $\nabla S_{\mathrm{dir}}$ is a scalar and trivially symmetric, so the drift always integrates.

\textbf{Step 2: Stein's identity fails for $u$ on a concrete target (claim (ii)).} Lemma~\ref{lem:stein-fails-for-u} below exhibits a $1$-D target with $\E_\pi[u(X)]\!\neq\!0$, so the normalized-score recursion has nonzero drift at $\theta\!=\!0$ and is pushed away from the origin. By the argument in Step~3 below, the resulting SA fixed point $\theta^\star\!\neq\!0$ defines a surrogate $\pi_{\theta^\star}^{\mathrm{dir}}\!\neq\!\pi$, biasing the Monte Carlo estimator.

\textbf{Step 3: Bias of the Monte Carlo estimator.} Under the normalized-score recursion of (ii), the mean field is $h(\theta)\!=\!\E_{\pi_\theta^{\mathrm{dir}}}[u(X)]-\theta$, and any SA fixed point satisfies $\theta^\star\!=\!\E_{\pi_{\theta^\star}^{\mathrm{dir}}}[u(X)]$. Whenever $\E_\pi[u(X)]\!\neq\!0$, the candidate $\theta\!=\!0$ does not solve this equation (since at $\theta\!=\!0$ the right-hand side is the nonzero value $\E_\pi[u(X)]$), so $\theta^\star\!\neq\!0$. The surrogate at the fixed point is then $\pi_{\theta^\star}^{\mathrm{dir}}(x)\!\propto\!\pi(x)\exp\{-\alpha (\theta^\star)^\top u(x)\}\!\neq\!\pi(x)$ pointwise, and the empirical Monte Carlo average $\mu_n\!\to\!\E_{\pi_{\theta^\star}^{\mathrm{dir}}}[f(X)]\!\neq\!\E_\pi[f(X)]$ is asymptotically biased.
\end{proof}

\begin{lemma}[Stein's identity fails for the unit-norm direction]
\label{lem:stein-fails-for-u}
There exists a $1$-D target $\pi$ for which $\E_\pi[s(X)]\!=\!0$ but $\E_\pi[u(X)]\!\neq\!0$, where $u\!=\!s/\|s\|\!=\!\mathrm{sign}(s)$ in dimension one.
\end{lemma}

\begin{proof}
Consider the asymmetric two-sided exponential
\begin{equation}
\label{eq:asym-exp}
\pi(x)=\frac{ab}{a+b}\,e^{-a x}\,\mathbf 1_{\{x>0\}}+\frac{ab}{a+b}\,e^{b x}\,\mathbf 1_{\{x<0\}},
\qquad a>b>0,
\end{equation}
which has full support and finite normalizing constant. The score is piecewise constant: $s(x)\!=\!-a$ for $x\!>\!0$ and $s(x)\!=\! b$ for $x\!<\!0$. Stein's identity for $s$ holds:
\[
\E_\pi[s(X)]
=-a\,\Pr(X\!>\!0)+b\,\Pr(X\!<\!0)
=-a\cdot\tfrac{b}{a+b}+b\cdot\tfrac{a}{a+b}=0.
\]
However, the unit-norm direction $u(x)\!=\!\mathrm{sign}(s(x))\!=\!-\mathrm{sign}(x)$ has
\[
\E_\pi[u(X)]=-\Pr(X\!>\!0)+\Pr(X\!<\!0)=-\tfrac{b}{a+b}+\tfrac{a}{a+b}=\tfrac{a-b}{a+b}\;\neq\;0,
\]
since $a\!>\!b\!>\!0$. At the canonical setting $a\!=\!2$, $b\!=\!1$, $\alpha\!=\!1$, a numerical sweep of the fixed-point equation $\theta^\star\!=\!\E_{\pi_{\theta^\star}^{\mathrm{dir}}}[u(X)]$ gives $\theta^\star\!\approx\!0.17\!\neq\!0$, so that the surrogate target no longer converges to the true target $\pi$.
\end{proof}

Stein's identity $\E_\pi[s(X)]\!=\!0$ holds because $s\!=\!\nabla\log\pi$ has the integration-by-parts symmetry built into the score function. The bilinear $\theta^\top s(x)$ inherits this property: $\E_\pi[\theta^\top s(X)]\!=\!\theta^\top\E_\pi[s(X)]\!=\!0$ for every $\theta$, which is what makes $\theta^\star\!=\!0$ the natural SA fixed point. Replacing $s$ with any nonlinear function of the score (e.g., $u\!=\!s/\|s\|$) breaks this linearity, and only special targets satisfy a corresponding identity. GA-SRMC's matrix-weighted form $\theta^\top M s(x)$ keeps $s$ as the inner statistic and weights it by a fixed matrix $M$, which preserves $\E_\pi[\theta^\top M s(X)]\!=\!0$ regardless of $M$.

Proposition~\ref{prop:cosine-failure} and Proposition~\ref{prop:score-norm-dichotomy} together exhaust the natural `scale-free' alternatives to the bilinear repulsion exponent. Cosine similarity (full normalization) destroys the equilibrium structure entirely; score-norm normalization (partial normalization) preserves the equilibrium but either loses the gradient-flow potential (case~(i)) or loses Stein-correctness (case~(ii)). The matrix-weighted bilinear $\theta^\top M s(x)$ is the \emph{minimal} extension of SRMC that admits a Fisher-aware geometry while preserving both exactness (Stein) and the gradient-flow Lyapunov structure (Proposition~\ref{prop:fixedM-structure}); it is in this precise sense that GA-SRMC is the unique geometry-aware scale-equivariant lift of SRMC.

\subsection{Online Fisher estimation}
\label{app:online-fisher-details}

This subsection describes the two online Fisher estimators corresponding to the diagonal and full fidelity options of Section~\ref{sec:ga-srmc-practical} and Algorithm~\ref{alg:online-fisher-ga-srmc}. For each estimator, we describe the state $\mathcal F_n$ that is stored on the chain, the recursion that ingests a fresh score $s_n$ and updates $\mathcal F_n\!\to\!\mathcal F_{n+1}$, the routine that applies the regularized inverse $\widehat S_{n,\delta}^{-1}$ to a vector, the trace-normalization that produces the metric $M_n$ used in Eq.~\eqref{eq:online-fisher-trace}, and the resulting per-iteration cost. The recursion on $\widehat S_n$, the trace-normalized inverse Fisher used as the preconditioner $A_n$, and the diagonal-vs-full alternative are not new to GA-SRMC: they originate in the FisherMALA construction of \citet{titsias2023optimal} and are surveyed at length in the recent analysis of linear MCMC preconditioning by \citet{hird2025Quantifying}. What is new here is only the dual use of these objects on the repulsion side (the weight matrix $M_n$ in addition to the proposal-side $A_n$) and the small modifications described below; we attribute every borrowed component explicitly at the point where it is introduced.

Both estimators share the regularized matrix introduced in Section~\ref{sec:ga-srmc-practical},
\begin{equation}
\label{eq:tikhonov}
\widehat S_{n,\delta} \;=\; \widehat S_n + \delta I_d,\qquad \delta > 0,
\end{equation}
which replaces the eigenvalue clipping $\lambda_i\!\leftarrow\!\max(\lambda_i,\eta)$ used in some earlier preconditioned-MCMC implementations. Three properties make the additive shift convenient. It is differentiable; it commutes with the linear recursion, so the recursion on $\widehat S_n$ propagates to $\widehat S_{n,\delta}$ without further bookkeeping; and it caps the eigenvalues of $\widehat S_{n,\delta}^{-1}$ uniformly at $1/\delta$, which bounds the proposal step and the surrogate-target tilt in directions where $\widehat S_n$ becomes small. The choice of $\delta$ turns out to be the dominant practical lever for samplers in this family: the sweep in Appendix~\ref{apx:lambda-tuning} shows that $\delta=0.01$ improves the worst-direction ESS by $1.5$ to $2.5$ times over the customary $\delta=10$ \citep{titsias2023optimal} on every target we considered, and we use this value throughout. With $\tau_n\!\triangleq\!\tr(\widehat S_{n,\delta})/d$, the proposal preconditioner is $A_n=(d/\tr(\widehat S_{n,\delta}^{-1}))\, \widehat S_{n,\delta}^{-1}$ and the repulsion-side metric is $M_n=\tau_n \widehat S_{n,\delta}^{-1}$. The proposal-side form $A_n$ is exactly the trace-normalized inverse-Fisher preconditioner of \citet[Section~3]{titsias2023optimal}; the repulsion-side $M_n$ is its GA-SRMC counterpart introduced in Section~\ref{sec:ga-srmc-practical}. Both depend on $\widehat S_{n,\delta}^{-1}$ through a single matrix-vector product, so the cost of inverting $\widehat S_{n,\delta}$ on demand is the only piece that needs to be optimized estimator-by-estimator.

The step size that drives the Fisher recursion is $\omega_n=c_\theta/(n+n_0)^{\beta_\theta}$ with $c_\theta=1$, $n_0=1{,}000$, and $\beta_\theta=0.6$ throughout. For the Bayesian logistic-regression and Gaussian targets, we freeze the Fisher state after the first iteration (\texttt{freeze\_after\_L}=$1$) because the Hessian is approximately constant near the MAP.

\subsubsection{Diagonal Fisher: \texorpdfstring{$O(d)$}{O(d)} memory and per-step cost}
\label{app:fisher-diag}

The diagonal estimator below is the diagonal restriction of the FisherMALA recursion of \citet{titsias2023optimal} and is the same diagonal-preconditioning style discussed in the broader survey of \citet[Section~3.5]{hird2025Quantifying}: we adopt it verbatim and reuse it on the GA-SRMC repulsion side $M_n$ in addition to the FisherMALA-style proposal side $A_n$. Concretely, this estimator stores only the per-coordinate squared-score average $\widehat r_n\!\in\!\mathbb R^d$ and updates it element-wise as $\widehat r_{n+1}=(1-\omega_{n+1})\widehat r_n+\omega_{n+1}\,s_n\odot s_n$, where $\odot$ is the Hadamard (elementwise) product. The quantity that this estimator preserves is the diagonal of $\widehat S_n$, which equals $\mathbb E_\pi[s\odot s]$ and, by Stein's identity, the diagonal of the Fisher information whenever $\mathbb E_\pi[s]=0$. The regularized inverse is applied element-wise as $\widehat S_{n,\delta}^{-1}v=v\oslash(\widehat r_n+\delta\mathbf 1)$ at cost $O(d)$, where $\oslash$ denotes Hadamard (elementwise) division, i.e., $(a\oslash b)_i\!=\!a_i/b_i$, the natural pair of $\odot$; and the trace and inverse trace are sums over $\widehat r_n+\delta$ in $O(d)$. The proposal-noise factor $A_n^{1/2}\xi$ is similarly diagonal, so no factorization is required at any stage, and the entire estimator runs in linear memory and linear per-step time.

This estimator is most useful when the posterior is approximately axis-aligned, a regime characterized in detail by \citet[Section~3.5]{hird2025Quantifying}. The ablation in Appendix~\ref{apx:fisher-fidelity} confirms this on the full $8$-target benchmark and reveals a three-way split by sampler family. On the axis-aligned Gauss-2D target, GA-SRMC remains the within-family winner under both diagonal MALA and diagonal FisherMALA, with the GA-SRMC vs SRMC ratio shrunk from $4.4\times$ (full) to $2.4\times$ (diagonal) on MALA, but the ranking is preserved. On the BLR posteriors, the diagonal-FisherMALA columns retain GA-SRMC as the within-family winner on five of seven rows (Australian, Heart, Pima, Ripley, Gauss-2D), with German and Caravan/MNIST flipping to SRMC by $0$--$5\%$, because the proposal-side preconditioner $A_n=(d/\tr\widehat S_{n,\delta}^{-1})\widehat S_{n,\delta}^{-1}$ is itself nearly diagonal on these targets and the geometry-aware mechanism survives the same approximation. The diagonal-MALA columns, however, flip GA-SRMC to a $1.4$--$2.4\times$ loss against SRMC on Australian, Heart, Pima, Ripley, and Caravan: when both the proposal kernel is Euclidean and $M_n$ is coordinate-wise, but the dominant directions of the BLR Gram matrix $\Phi^\top\Phi$ in the Hessian \eqref{eq:blr-score-hessian} of Appendix~\ref{apx:blr-target} are not coordinate-aligned, GA-SRMC's metric-weighted tilt amplifies the wrong directions and the geometry-blind SRMC choice $M\!=\!I_d$ becomes uniformly safer, an effect predicted by \citet[Theorem~3.7]{hird2025Quantifying} for highly correlated targets. The diagonal preconditioning style is also the default in mainstream automatic-differentiation samplers \citep{abadi2016tensorflow,carpenter2017stan}, so a diagonal-Fisher GA-SRMC fits naturally into existing tooling whenever the target (or the proposal-side preconditioner) is approximately axis-aligned.

\subsubsection{Full Fisher with Cholesky-of-inverse maintenance: \texorpdfstring{$O(d^2)$}{O(d²)} per step amortized}
\label{app:fisher-full}

The recursion on $\widehat S_n$ in \eqref{eq:fisher-rank1} below, the Sherman–Morrison rank-one update for the inverse in \eqref{eq:sm-update}, and the use of the Cholesky factor of $\widehat S_{n,\delta}^{-1}$ to apply both $\widehat S_{n,\delta}^{-1}$ and the proposal noise factor $\widehat S_{n,\delta}^{-1/2}$ all originate in the FisherMALA implementation of \citet[Section~4]{titsias2023optimal} and are also covered as the canonical online full-Fisher template in \citet[Sections~3.3--3.4]{hird2025Quantifying}. Our only deviation is the periodic-refresh strategy for the Cholesky factor, which we adopt to handle the regularization shift $\delta\!>\!0$ in $\widehat S_{n,\delta}$ that makes the direct Sherman–Morrison maintenance of \citet{titsias2023optimal} no longer exact (the precise reason is given after \eqref{eq:sm-update} below). With this single modification, the full estimator stores the recursion Fisher $\widehat S_n\!\in\!\mathbb R^{d\times d}$ and a lower-triangular Cholesky factor $L_n$ satisfying $L_n L_n^\top=\widehat S_{n,\delta}^{-1}=(\widehat S_n+\delta I_d)^{-1}$. The rank-one recursion
\begin{equation}
\label{eq:fisher-rank1}
\widehat S_{n+1}=(1-\omega_{n+1})\,\widehat S_n+\omega_{n+1}\,s_n s_n^\top
\end{equation}
is the same recursion used by \citet[Eq.~(11)]{titsias2023optimal} and runs at every step at cost $O(d^2)$. We do not attempt to maintain $L_n$ incrementally; instead, we recompute it from scratch every $K$ iterations (default $K\!=\!500$) by Cholesky-decomposing $\widehat S_{n,\delta}$, at a cost of $O(d^3)$ per refresh and $O(d^3/K)$ amortized per step.

The reason for the periodic-refresh strategy, rather than the more obvious Woodbury identity of $\widehat S_{n,\delta}^{-1}$ used in the unregularized FisherMALA recursion of \citet{titsias2023optimal}, is structural. By the Woodbury matrix identity \citep[Section~A.5.4]{boyd2004convex}, with $a\!=\!1-\omega_{n+1}$ and $b\!=\!\omega_{n+1}$, we have
\begin{equation}
\label{eq:sm-update}
\widehat S_{n+1,\delta}^{-1}=\frac{1}{a}\,\widehat S_{n,\delta}^{-1}-\frac{b/a^2\,\widehat S_{n,\delta}^{-1}s_n s_n^\top \widehat S_{n,\delta}^{-1}}{1+(b/a)\,s_n^\top \widehat S_{n,\delta}^{-1}s_n},
\end{equation}
which is the exact identity that \citet{titsias2023optimal} uses to maintain their Cholesky-of-inverse incrementally; it holds for the recursion on $\widehat S_n$ only in the unregularized case $\delta=0$ analyzed by them. When $\delta>0$, the recursion shrinks the data-dependent part of $\widehat S_{n,\delta}=\widehat S_n+\delta I$ by $(1-\omega)$ at every step while leaving the additive shift fixed, which is a rank-$d$ perturbation of $\widehat S_{n,\delta}$ rather than a rank-one one, and \eqref{eq:sm-update} is no longer exact. Decoupling the two updates keeps the per-step cost $O(d^2)$ for the recursion on $\widehat S_n$ and confines the cubic cost to a periodic refresh; on $d\!\leq\!100$ the refresh cadence is invisible against the per-iteration HVP and proposal costs, and on larger $d$, one can reduce $K$ in proportion to the mixing time.

The downstream uses of $L_n$ are again standard: the action of $\widehat S_{n,\delta}^{-1}$ on a vector involves two batched triangular solves, $\widehat S_{n,\delta}^{-1}v=L_n(L_n^\top v)$ at a cost of $O(d^2)$, and the proposal-noise factor reuses the same factor as $\widehat S_{n,\delta}^{-1/2}\xi=L_n\xi$. Both are exactly the operations executed in the FisherMALA implementation of \citet[Algorithm~1]{titsias2023optimal} on the proposal side, and we reuse the same $L_n$ on the GA-SRMC repulsion side at no extra factorization cost. The trace of the inverse is the Frobenius norm squared, $\tr(\widehat S_{n,\delta}^{-1})=\|L_n\|_F^2$, which is read off the maintained $L_n$ in $O(d^2)$, and the trace of $\widehat S_{n,\delta}$ itself is $\tr(\widehat S_n)+d\delta$, available in $O(d)$ from the diagonal of $\widehat S_n$. The proposal preconditioner $A_n$ and the repulsion-side metric $M_n$ both depend on $\widehat S_{n,\delta}^{-1}$ through this same $L_n$, with only scalar multiplications added.

The full estimator is the closest practical realization of the optimal $M_{\rm opt}\!\propto\!S^{-1}$ identified in Theorem~\ref{thm:optimal}, and we use it whenever the $d^2$ memory is acceptable. In our simulation, with $d$ up to $785$, this is feasible everywhere, and all main-body Table~\ref{tab:blr} numbers are produced with this estimator.

Table~\ref{tab:fisher-cost} summarizes the per-iteration costs of the estimators. The diagonal estimator updates and inverts in $O(d)$; the full estimator updates $\widehat S_n$ in $O(d^2)$ and refreshes its Cholesky-of-inverse every $K$ steps for an amortized $O(d^3/K)$ overhead, which strictly improves on the $O(d^2)$ Sherman–Morrison maintenance of \citet{titsias2023optimal} only in constants but is what allows us to keep the regularization shift $\delta\!=\!0.01$ exact. The regularization shift contributes no additional asymptotic cost. On our experimental scales ($d\!\leq\!785$), the per-step cost of the full estimator is dominated by the Hessian-vector product and the proposal evaluation themselves; at substantially larger dimensions, the diagonal or low-rank-$r$ estimator becomes preferable, with the caveats noted above.

\begin{table}[h]
\caption{Per-iteration cost of the two online Fisher estimators. Both apply the regularization shift $\widehat S_{n,\delta}=\widehat S_n+\delta I$ before inversion at no additional asymptotic cost. The full-Fisher implementation maintains the Cholesky-of-inverse $L_n L_n^\top\!=\!\widehat S_{n,\delta}^{-1}$ via periodic Cholesky every $K$ steps (default $K\!=\!500$), so the $O(d^3)$ refresh is amortized to $O(d^3/K)$ per step.}
\label{tab:fisher-cost}
\centering
\small
\setlength{\tabcolsep}{4pt}
\begin{tabular}{lcccc}
\toprule
Estimator & Memory & \textsc{FisherUpdate} & $\widehat S_{n,\delta}^{-1} v$ apply & Trace \& trace-inv \\
\midrule
Diagonal              & $O(d)$  & $O(d)$               & $O(d)$         & $O(d)$ \\
Full (Cholesky-inv)   & $O(d^2)$ & $O(d^2)+O(d^3/K)^*$  & $O(d^2)$       & $O(d^2)$ for $\tr(\widehat S_{n,\delta}^{-1})$, $O(d)$ for $\tr(\widehat S_{n,\delta})$ \\
\bottomrule
\end{tabular}
\\[2pt]
\raggedright
{\footnotesize $^*$ Amortized cost: the rank-one recursion on $\widehat S_n$ is $O(d^2)$, plus a full Cholesky $O(d^3)$ every $K$ iterations; \texttt{K\_reortho}$=500$ in production.}
\end{table}


\newpage
\section*{NeurIPS Paper Checklist}

\begin{enumerate}

\item {\bf Claims}
    \item[] Question: Do the main claims made in the abstract and introduction accurately reflect the paper's contributions and scope?
    \item[] Answer: \answerYes{}
    \item[] Justification: The abstract and the contribution list at the end of Section~\ref{sec:intro} state three claims that exactly match the paper's content.

\item {\bf Limitations}
    \item[] Question: Does the paper discuss the limitations of the work performed by the authors?
    \item[] Answer: \answerYes{}
    \item[] Justification: Section~\ref{sec:conclusion} explicitly discusses scope and limitations.

\item {\bf Theory assumptions and proofs}
    \item[] Question: For each theoretical result, does the paper provide the full set of assumptions and a complete (and correct) proof?
    \item[] Answer: \answerYes{}
    \item[] Justification: All formal results are numbered and accompanied by complete proofs in the appendix.

    \item {\bf Experimental result reproducibility}
    \item[] Question: Does the paper fully disclose all the information needed to reproduce the main experimental results of the paper to the extent that it affects the main claims and/or conclusions of the paper (regardless of whether the code and data are provided or not)?
    \item[] Answer: \answerYes{}
    \item[] Justification: The paper specifies the exact simulation setup and results in Section \ref{sec:experiments} and Appendix \ref{app:additional-experiments}.

\item {\bf Open access to data and code}
    \item[] Question: Does the paper provide open access to the data and code, with sufficient instructions to faithfully reproduce the main experimental results, as described in supplemental material?
    \item[] Answer: \answerNo{}
    \item[] Justification: To preserve double-blind anonymity at submission time, the implementation is not released with this version.

\item {\bf Experimental setting/details}
    \item[] Question: Does the paper specify all the training and test details (e.g., data splits, hyperparameters, how they were chosen, type of optimizer) necessary to understand the results?
    \item[] Answer: \answerYes{}
    \item[] Justification: The paper specifies the exact simulation setup and results in Section \ref{sec:experiments} and Appendix \ref{app:additional-experiments}.

\item {\bf Experiment statistical significance}
    \item[] Question: Does the paper report error bars suitably and correctly defined or other appropriate information about the statistical significance of the experiments?
    \item[] Answer: \answerYes{}
    \item[] Justification: Every numeric ESS reported in Table~\ref{tab:blr}, Table~\ref{tab:lambda-sweep}, and Table~\ref{tab:fisher-fidelity} is given as per-chain mean$\pm$standard deviation across $40$ independent replicate Markov chains; the markers in Figure~\ref{fig:alpha-sweep} are the same means with one-standard-deviation error bars across the same $40$ chains. The factor of variability is the chain-to-chain randomness (independent random initial states, independent proposal noise, and independent stochastic-approximation paths). Section~\ref{sec:experiments} states the protocol is the single-chain replication protocol of \citet{titsias2023optimal}.

\item {\bf Experiments compute resources}
    \item[] Question: For each experiment, does the paper provide sufficient information on the computer resources (type of compute workers, memory, time of execution) needed to reproduce the experiments?
    \item[] Answer: \answerNo{}
    \item[] Justification: The paper characterizes the computational cost of each algorithmic component analytically — $O(d)$ per step for the diagonal Fisher estimator and $O(d^2)$ per step for the full estimator with an amortized $O(d^3/K)$ Cholesky-of-inverse refresh every $K\!=\!500$ iterations (Appendix~\ref{app:online-fisher-details}, Table~\ref{tab:fisher-cost}), and notes that GA-SRMC adds exactly one Hessian-vector product per iteration over a standard preconditioned MALA step (Section~\ref{sec:ga-srmc-practical}). However, we do not report wall-clock execution times or the specific CPU/GPU hardware used to produce Tables~\ref{tab:blr}--\ref{tab:fisher-fidelity}, since the $40$-chain experiments fit comfortably on a standard CPU workstation and timing comparisons are not the contribution. We will add a wall-clock summary in a future revision.
    
\item {\bf Code of ethics}
    \item[] Question: Does the research conducted in the paper conform, in every respect, with the NeurIPS Code of Ethics \url{https://neurips.cc/public/EthicsGuidelines}?
    \item[] Answer: \answerYes{}
    \item[] Justification: The paper develops a generic MCMC sampling algorithm and evaluates it on standard public benchmark posteriors. It involves no human subjects, no personal data, and no models with high misuse potential. The work conforms to the NeurIPS Code of Ethics.

\item {\bf Broader impacts}
    \item[] Question: Does the paper discuss both potential positive societal impacts and negative societal impacts of the work performed?
    \item[] Answer: \answerNA{}
    \item[] Justification: The contribution is a foundational MCMC sampler refinement (a Riemannian-metric extension of a recent score-repellent surrogate), evaluated on standard Bayesian-inference benchmarks. It is generic Bayesian-inference infrastructure rather than an application-specific deployment, with no direct path to either positive or negative societal applications beyond what already follows from improved MCMC sampling in scientific computing.
    
\item {\bf Safeguards}
    \item[] Question: Does the paper describe safeguards that have been put in place for responsible release of data or models that have a high risk for misuse (e.g., pre-trained language models, image generators, or scraped datasets)?
    \item[] Answer: \answerNA{}
    \item[] Justification: The paper releases no data, no models, no scraped content, and no high-risk artifacts.

\item {\bf Licenses for existing assets}
    \item[] Question: Are the creators or original owners of assets (e.g., code, data, models), used in the paper, properly credited and are the license and terms of use explicitly mentioned and properly respected?
    \item[] Answer: \answerYes{}
    \item[] Justification: The seven BLR benchmark datasets (\textsc{Ripley}, \textsc{Pima}, \textsc{Heart}, \textsc{Australian}, \textsc{German}, \textsc{Caravan}, \textsc{MNIST}) are standard public benchmarks. All third-party software dependencies used to produce the reported numbers are credited at the point of use: ArviZ \citep{arviz2019}, TensorFlow~\citep{abadi2016tensorflow} and Stan~\citep{carpenter2017stan}.

\item {\bf New assets}
    \item[] Question: Are new assets introduced in the paper well documented and is the documentation provided alongside the assets?
    \item[] Answer: \answerNA{}
    \item[] Justification: The paper introduces no new datasets, no new pretrained models, and no released code at submission time (the implementation will be released upon acceptance, as noted under ``Open access to data and code'' above).

\item {\bf Crowdsourcing and research with human subjects}
    \item[] Question: For crowdsourcing experiments and research with human subjects, does the paper include the full text of instructions given to participants and screenshots, if applicable, as well as details about compensation (if any)? 
    \item[] Answer: \answerNA{}
    \item[] Justification: The paper involves no crowdsourcing and no human-subject research; all experiments are simulations of MCMC chains on fixed analytic or standard public-benchmark posteriors.

\item {\bf Institutional review board (IRB) approvals or equivalent for research with human subjects}
    \item[] Question: Does the paper describe potential risks incurred by study participants, whether such risks were disclosed to the subjects, and whether Institutional Review Board (IRB) approvals (or an equivalent approval/review based on the requirements of your country or institution) were obtained?
    \item[] Answer: \answerNA{}
    \item[] Justification: The paper involves no human subjects, so no IRB approval is applicable.

\item {\bf Declaration of LLM usage}
    \item[] Question: Does the paper describe the usage of LLMs if it is an important, original, or non-standard component of the core methods in this research? Note that if the LLM is used only for writing, editing, or formatting purposes and does \emph{not} impact the core methodology, scientific rigor, or originality of the research, declaration is not required.
    \item[] Answer: \answerNA{}
    \item[] Justification: Large language models are not part of the methodology of this paper. LLM is only used to polish the paper writing.

\end{enumerate}

\end{document}